\documentclass{article}

\usepackage{preprint}

\usepackage{placeins} % USED THIS TO FORCE REFERENCES TO BE ON LAST PAGE (BUT STILL BEFORE APPENDIX)

\usepackage[utf8]{inputenc} % allow utf-8 input
\usepackage[T1]{fontenc}    % use 8-bit T1 fonts
\usepackage{url}            % simple URL typesetting
\usepackage{booktabs}       % professional-quality tables
\usepackage{amsfonts}       % blackboard math symbols
\usepackage{nicefrac}       % compact symbols for 1/2, etc.
\usepackage{microtype}      % microtypography
\usepackage{lipsum}
\usepackage{wrapfig}
\usepackage{subcaption}
\usepackage{cancel}

\usepackage{booktabs} % For formal tables
\usepackage{amsmath,amssymb}
\usepackage{mathtools}
\usepackage[toc,page]{appendix}
\usepackage{enumitem}
\usepackage{graphics}
\usepackage{array,multirow,graphicx}
\usepackage[dvipsnames]{xcolor}
\definecolor{darkblue}{HTML}{00008b}
\usepackage{algorithm}
\usepackage{algorithmicx}
\usepackage[noend]{algpseudocode}
\algnewcommand{\LineComment}[1]{\State \(//\) #1}

\makeatletter
\def\mathcolor#1#{\@mathcolor{#1}}
\def\@mathcolor#1#2#3{%
  \protect\leavevmode
  \begingroup
    \color#1{#2}#3%
  \endgroup
}
\makeatother

\newcommand{\mean}[1]{\overline{\mathbf{#1}}}
\newcommand{\latent}[1]{\mathbf{#1}}

\usepackage[left]{lineno}
\usepackage{etoolbox} %% <- for \cspreto, \csappto

\newcommand*\linenomathpatch[1]{%
  \cspreto{#1}{\linenomath}%
  \cspreto{#1*}{\linenomath}%
  \csappto{end#1}{\endlinenomath}%
  \csappto{end#1*}{\endlinenomath}%
}

\linenomathpatch{equation}
\linenomathpatch{gather}
\linenomathpatch{multline}
\linenomathpatch{align}
\linenomathpatch{alignat}
\linenomathpatch{flalign}

\usepackage[font=small]{caption}

\title{The Neural Coding Framework for Learning\\ Generative Models}
\author{
  Alexander Ororbia* \\%$^\dagger$ \\
  Department of Computer Science \\
  Rochester Institute of Technology \\
  Rochester, NY 14623 \\
  \texttt{ago@cs.rit.edu} \\
  \And 
  Daniel Kifer \\%$^\dagger$ \\
  Department of Computer Science \& Engineering \\
  The Pennsylvania State University \\
  State College, PA 16801 \\
  \texttt{duk17@psu.edu} \\
}

\begin{document}
\maketitle
\vspace{-0.5cm}
\begin{abstract}
Neural generative models can be used to learn complex probability distributions from data, to sample from them, and to produce probability density estimates. We propose \textcolor{black}{a computational framework for developing neural generative models} inspired by the theory of predictive processing in the brain. According to predictive processing theory, the neurons in the brain form a hierarchy in which neurons in one level form expectations about sensory inputs from another level. These neurons update their local models based on differences between their expectations and the observed signals. In a similar way, artificial neurons in our generative \textcolor{black}{models} predict what neighboring neurons will do, and adjust their parameters based on how well the predictions matched reality. In this work, we show that \textcolor{black}{the neural generative models learned within our framework perform} well in practice across several benchmark datasets and metrics and either \textcolor{black}{remain} competitive with or significantly \textcolor{black}{outperform} other generative models with similar functionality (such as the variational auto-encoder).
\end{abstract}


\section{Introduction}
\label{sec:intro}

% The brain is a generative model (this is hypothesis that the NGC commits itself to...) -- it should go in this direction of "brain does this, backprop doesn't, but we can change backprop so that this also happens in deep learning and it works")
One way to understand how the brain adapts to its environment is to view it as a type of generative pattern-creation model \cite{friston2006free}, one that is engaged in a never-ending process of self-correction, often without external teaching signals (or labels) \cite{ororbia2020continual}. Under this perspective, the brain is continuously making predictions about elements of its environment, a process that allows it to infer useful representations of the sensory data \textcolor{black}{that} it receives \cite{parr2018anatomy} as well as to synthesize novel patterns, which could serve as the potential basis for long-term planning and imagination itself \cite{clark2015surfing}.
%In other words, one way the brain could adapt to its environment is by learning to act like a form of ``inverse computer graphics'' \cite{hinton2007recognize}, or, in other words, a hierarchical model where layers farther away from pixel space are more abstract and function like coordinate vectors (with matrices in between preserving transformations), acquiring internal representations of its environment that serve as the building blocks to long-term planning and imagination itself \cite{clark2015surfing}. Learning could then be carried out through the use of error, i.e., mismatch signals between expected and actual perceptions. 
From the theoretical viewpoint of predictive processing \cite{clark2015surfing}, the brain could be likened to a hierarchical model \cite{friston2008hierarchical} whose levels are implemented by neurons (or clusters of neurons). If levels are likened to regions of the brain, the neurons at one level (region) attempt to predict the state of neurons at another level (region) and adjust/correct their local model synaptic parameters based on how different their predictions were from the observed signal. Furthermore, these neurons utilize various mechanisms to laterally stimulate/suppress each other \cite{liang2017interactions} to facilitate contextual processing (such as grouping/segmenting visual components of objects in a scene).
As we will demonstrate in this article, this viewpoint can be turned into a powerful framework for learning generative models.

% Describe the essence of machine learning-based generative modeling
In machine learning, one central goal is to construct agents that learn distributed representations that extract the underlying structure of data, without the use of explicit supervisory signals such as human-crafted labels, i.e., unsupervised learning. 
Generative models, or models capable of synthesizing instances of data that resemble a database of collected patterns, that are based on deep artificial neural networks (ANNs), e.g., variational autoencoders \cite{kingma2013auto} or generative adversarial networks \cite{goodfellow2014generative}, have been shown to be one way of acquiring these representations.
Once trained, an ANN model is used to ``fantasize'' patterns by injecting it with noise and propagating this noise through the system until the output nodes are reached.

% What is wrong with generative models? backprop!
However, despite the success in deploying ANNs as generative models across a variety of applications, the way that ANNs operate and learn is a far-cry from the neuro-mechanistic story we described earlier \cite{crick1989recent,zador2019critique}. Specifically, ANN generative models are trained with the popular workhorse algorithm known as back-propagation of errors (backprop) \cite{rumelhart_learning_1986}, which is an elegant mathematical solution to the credit assignment problem in deep networks -- synaptic weights are adjusted through the use of teaching signals that are created by propagating an error, which exists exclusively at the output of the ANN, backwards along a feedback pathway \cite{ororbia2019biologically} (a path created by re-using the same weights that transmitted signals forward \cite{grossberg1987competitive}) .
By virtue of this formulation, backprop imposes the constraint that the ANN take the form of a directed feedforward structure (and does not permit the use non-differentiable activation functions and makes integrating other mechanisms such as lateral connectivity difficult). While the neurons in an ANN are usually arranged hierarchically, they do not make local predictions and they do not laterally affect each other's activity. Furthermore, synaptic adjustment in backprop-based models is done non-locally, while in neurobiological networks this adjustment is often argued to be done locally \cite{hebb1949organization,magee1997synaptically,bi2001synaptic,isomura2018error} (there are far more local connections than long-range connections \cite{zhang2000universal} with the neocortex adhering a local connectivity pattern \cite{zador2019critique}) -- that is, neurons make use of the information immediately available to them (in both time and space) and do not wait on distant regions in order to adjust their synapses (with global information provided through neurotransmitters such as dopamine).
In response to the above problems, the statistical learning community has developed a plethora of mechanisms that either modify backprop through a heuristic or additional constraint \cite{pascanu2013difficulty,ioffe2015batch,MishkinM15,ba2016layer} or, recently, worked on developing learning procedures that embody elements of biological neuronal function and computation while enabling backprop-level learning \cite{hinton1988learning,lee2015difference,baldi2016learning,lillicrap2016random,scellier2017equilibrium,ororbia2019biologically} (see Supplementary Note 2 for a more detailed review). However, while insights from each development have proven valuable in bridging backprop with brain-like computation, many of these ideas only address one or a few of the issues described earlier and tend to focus on simple problems in classification.
While the question as to how credit assignment is exactly implemented in the brain is an open one, it would prove useful to machine learning, (computational) neuroscience, and cognitive science to have a framework that demonstrates how a neural system can learn something as complex as a generative model without backprop, using mechanisms and rules that are brain-inspired. 

In this work, inspired by predictive processing theory \cite{von1855ueber,bastos2012canonical,swanson2016predictive,clark2015surfing} and building on free energy minimization \cite{friston2008hierarchical,friston2009free}, which crucially casts predictive processing formally as variational Bayesian inference, we propose the neural generative coding (NGC) computational framework as a powerful way to learn generative ANNs, resolving several of the key backprop-centric issues described above. \textcolor{black}{Furthermore, we show that certain settings of NGC recover the work proposed in \cite{rao1999predictive} and \cite{friston2008hierarchical}.}
%Our framework resolves key backprop-centric issues described above and offers a powerful starting point for crafting agents that learn as well as backprop but are not subject to backprop's constraints, creating a stronger connection between neuro-cognition and statistical learning.
% certain settings in the framework recover some work proposed by [15] and [47]
We find that NGC models\textcolor{black}{, including \cite{rao1999predictive} and \cite{friston2008hierarchical},} not only remain competitive with backprop-based generative ANNs across several datasets in terms of pattern creation, such as the variational autoencoder \cite{kingma2013auto}, but that they also outperform these models on tasks that they are not directly trained for, such as classification and pattern completion. 
%their work is good (we also see it is better than backprop), and with this unification, we can add some improvements like more sparsity (don't use "better')
\textcolor{black}{
Our results further demonstrate that besides unifying predictive processing moddels, NGC allows for integration of improvements such as learnable recurrent error synapses and laterally-driven sparsity.}
As a result, our work presents promising evidence that brain-inspired alterations to traditional deep learning techniques can be a viable source of performance gains. 
%we can learn powerful neural systems that are not subject backprop's constraints and have the potential to generalize in ways current modern-day ANNs cannot.
%Furthermore, NGC shares strong grounding in predictive processing theory, which could prove useful for future research-driven extensions.

\begin{figure}[t]
\begin{center}
\begin{subfigure}{0.5\textwidth}
  \centering
  % include first image
  \includegraphics[width=7.8cm]{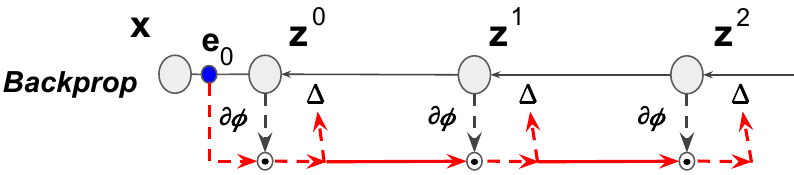}  
  \caption{}
  \label{fig:backprop_path}
\end{subfigure}%
\begin{subfigure}{0.5\textwidth}
  \centering
  % include second image
  \includegraphics[width=7.1cm]{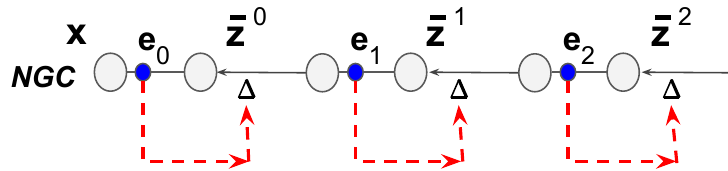}  
  \caption{}
  \label{fig:ncg_path}
\end{subfigure}
\end{center}
\vspace{-0.2cm}
\caption{
\textbf{Backprop in contrast with neural coding:}
(a) Credit assignment in backprop requires a strict, global feedback pathway, which requires the completion of the forward pass that carries information upstream (right to left), that carries the error message $\mathbf{e}^0$ at the output layer back along (left to right) the same synapses used in the forward pass to update downstream neurons  $\mathbf{\bar{z}}^1$ and $\mathbf{\bar{z}}^2$.
(b) Our proposed neural generative coding (NGC) model sidesteps this neurobiologically implausible requirement by learning with short, local error transmission pathways made possible through recurrent error synapses and stateful neural activities.
Credit assignment under NGC operates with local mismatch signals, $\mathbf{e}^1$ and $\mathbf{e}^2$, that readily communicate this information to their respective layers, $\mathbf{\bar{z}}^1$ and $\mathbf{\bar{z}}^2$. Black arrows indicate forward propagation while red arrows indicate backwards transmission. Solid lines indicate that a signal is transformed along a synapse(s) while dashed arrows indicate direct copying of information (in other words, it represents the identity function, or that no transformation is applied to the incoming information). $\partial \phi$ shows communication of the neuron's first derivative, $\Delta$ represents the computed change to the synapse (of the forward pass) that will use the (nearby) error signal, and $\odot$ indicates multiplication of the incoming signals.
}
\label{fig:bp_v_ngc}
\vspace{-0.4cm}
\end{figure}

\section{Results}
\label{sec:results}

\subsection{Generative Neural Coding Learns Viable Auto-Associative Generative Models}
\label{sec:results_gen_models}
%%%%%%%%%%%%%%%%%%%%%%%%%%%%%%%%%%%%%%%%%%%%%%%%%%%%%%%%%%%%%%%%%%%%%%%%%%%%%%%%%%%%%%%%%%%%%%

\textbf{Notation:} 
In this paper, $\odot$ indicates a Hadamard product, $\oslash$ indicates element-wise division, and $\cdot$ indicates a matrix/vector multiplication (or dot product if the two objects it is applied to are vectors of the exact same shape) and $(\mathbf{v})^T$ denotes the transpose.
We denote $\mathbf{v}_i$ (bold font indicates vector/matrix) means that we retrieve the $i$th element $v_i$ (italic indicates single scalar element) in the vector $\mathbf{v}$ and $\mathbf{W}_{ij}$ means that we retrieve the element $W_{ij}$ in the $i$th row and $j$th column of matrix $\mathbf{W}$.

%%%%%%%%%%%%%%%%%%%%%%%%%%%%%%%%%%%%%%%%%%%%%%%%%%%%%%%%%%%%%%%%%%%%%%%%%%%%%%%%%%%%%%%%%%%%%%
% Problem setting/description
\textbf{Problem Setting:} We start with a description of the problem setting -- an agent must learn to approximate a probability distribution  from a dataset $\mathbf{X}$ of samples. For notational reasons, this dataset is presented in column-major order, so that each column $\mathbf{x}$ represents a record (also known as an example or item).  $\mathbf{X}$ has $D$ rows and $S$ total columns. The items are assumed independent, so that $p(\mathbf{X}) = \prod_{\mathbf{x} \in \mathbf{X}} p(\mathbf{x})$ and $\log p(\mathbf{X}) = \sum_{\mathbf{x} \in \mathbf{X}} \log p(\mathbf{x})$. We are interested in  directed generative models that are capable of producing \emph{explicit} density estimates of the data distribution, i.e., models that estimate a probability density function (PDF) over a sample space, and we will leave the examination of most \emph{implicit} density estimators (i.e., models that do not produce explicit density estimates of the PDF but yield a function that produces samples from the estimated distribution), based on generative adversarial networks \cite{goodfellow2014generative} for future work. 

% What does deep learning practice do
%In modern-day, deep learning practice, an auto-associative network, or an ANN that learns to predict the input $\mathbf{x}$ given the input, such as a simple auto-encoder, would be constructed as one way of modeling the input distribution.
\textbf{The Typical Deep Learning Approach:} In modern-day deep learning practice, a feedforward ANN, also called a \emph{decoder}, would be constructed to model the desired input distribution. The decoder ($\mbox{NN}$) takes as input a noise vector or a sampled latent variable $\mathbf{z}$ and maps it to the parameters of a probability distribution, such a mean and covariance of a multivariate Gaussian, or, as in this paper, the mean of a multivariate Bernoulli distribution, i.e., $\mathbf{z}^0=\mbox{NN}(\mathbf{z})$ where $\mathbf{x}_i \sim B(n=1,\mathbf{z}^0_i)$. This artificial neural network would typically be made up of $L+1$ layers of neurons ($L$ layers of hidden neurons and one output layer of neurons), where the state in layer $\ell$ is represented by a vector $\mathbf{z}^\ell$. Each layer $\ell$ is interpreted as a transformation of the layer before it. In essence, $\mathbf{z}^{\ell-1}=\phi^{\ell-1}(\mathbf{W}^\ell \cdot \mathbf{z}^\ell)$ where $\mathbf{z}^L$ of layer $L$ to be the same as the input noise vector $\mathbf{z}$.
%That is, setting the state $\mathbf{z}^L$ of layer $L$ to be the same as the input noise vector $\mathbf{z}$, we get $\mathbf{z}^{L-1}=\phi^{L-1}(\mathbf{W}^L \cdot \mathbf{z}^L)$, where $\cdot$ indicates matrix multiplication, $\phi^{L-1}$ is an activation function and $\mathbf{W}^L$ are tunable weights. In general,  ${\mathbf{z}^\ell= \phi^\ell(\mathbf{W}^{\ell+1} \cdot \mathbf{z}^{\ell+1}) }$.
The output $\mathbf{z}^0$ of this decoder (Figure \ref{fig:bp_v_ngc} a) would be the parameters of a probability distribution, such as the mean of a Bernoulli distribution, or mean and covariance of a Gaussian (see Supplementary Note 7 for descriptions of the backprop-based networks used in this study). One can sample from this distribution to get a sample point $\mathbf{x}$ (or use the mean of the distribution directly).
To stabilize and speed up the model's learning process, an encoder is typically introduced which also takes in as input the sensory input $\mathbf{x}$ to be predicted. The encoder is designed to drive the parameters of a distribution, normally multivariate Gaussian, that shape and control the form of the latent variable $\mathbf{z}$, i.e., $\mu_z, \sigma^2_z = \mbox{NN}_e(\mathbf{x})$ where $\sigma^2_z = \mathbf{\Sigma}_z \odot \mathbf{I}$ (or diagonal covariance).

% How is deep learning generator trained in practice
To fit this model to the data, one would choose the weight parameters $\mathbf{W}^\ell$ to minimize a loss function $\psi$ such as the negative log-likelihood, typically using some variant of stochastic gradient descent. Often the backprop algorithm is used to compute the partial derivatives of  $\frac{\partial \psi}{\partial \mathbf{W}^\ell}$ needed for this optimization. 
%(applied to the outputs of the decoder), such as the Bernoulli log likelihood, with respect to each weight matrix $\mathbf{W}^\ell$, or $\frac{\partial \psi}{\partial \mathbf{W}^\ell}$ ($\Delta$ in Figure \ref{fig:bp_v_ngc} Left).
%This loss, which measures the network's performance on the given task, is applied to the outputs of the network. 
Computing the necessary derivatives according to backprop entails first computing an error signal at the output (downstream) layer, or $\mathbf{e}^0 = \frac{\partial \psi}{\partial \mathbf{z}^0}$. This error signal is then transmitted  to internal (upstream) neurons by carrying this signal back along the forward synapses that were originally used to  transform $\mathbf{z}^L$ (in short, this is done by multiplying the signal with the transpose of the forward weight matrices). Furthermore, knowledge of the derivative of each activation function $\phi^\ell$ is required during these computations (as shown in Figure \ref{fig:bp_v_ngc}, a).

\begin{figure}[t]
\begin{center}
\begin{subfigure}[t]{0.315\textwidth}
  \centering
  % include first image
  \includegraphics[width=4.7cm]{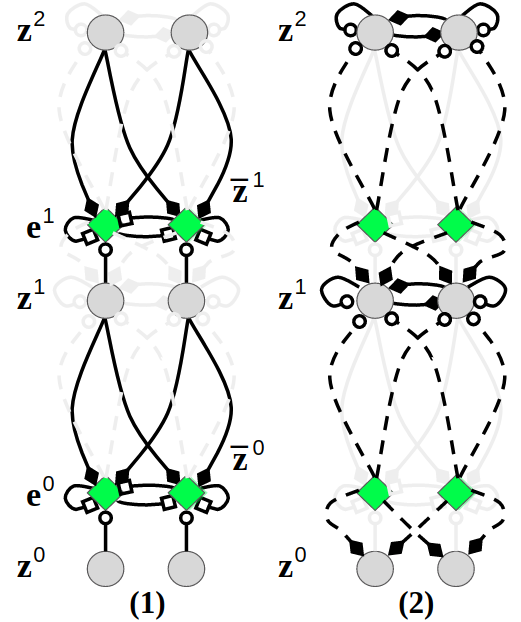}  
  %\caption{The three key computational steps taken by the GNCN at any step in time.}
  \caption{}
  \label{fig:inhibit}
\end{subfigure}%
\begin{subfigure}[t]{0.225\textwidth}
  \centering
  % include second image
  \includegraphics[width=3.75cm]{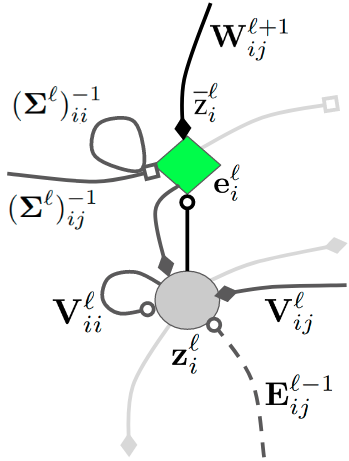}  
  %\caption{A single neural circuit consisting of a prediction neuron and an error neuron.}
  \caption{}
  \label{fig:predict}
\end{subfigure}
\begin{subfigure}[t]{0.3315\textwidth}
  \centering
  % include third image
  \includegraphics[width=6.85cm]{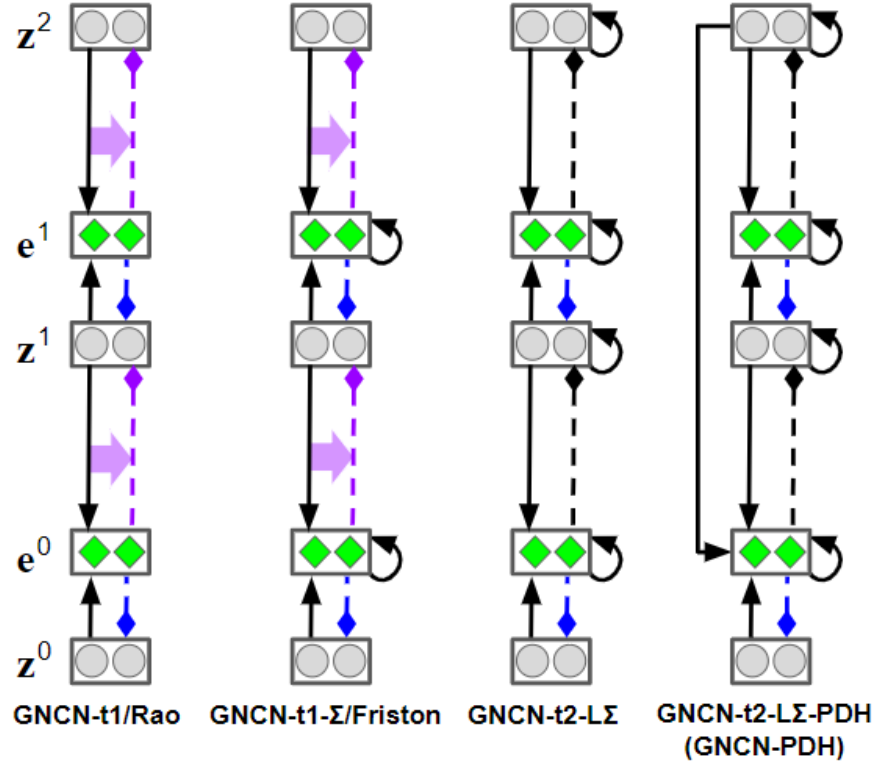}  
  %\caption{A single neural circuit consisting of a prediction neuron and an error neuron.}
  \caption{}
  \label{fig:ngc_arch}
\end{subfigure}
\end{center}
\vspace{-0.25cm}
\caption{
\textbf{\textcolor{black}{Neural generative coding} computation \& circuitry:}
(a) The two key computation steps taken by an entire NGC network (\textcolor{black}{a GNCN-t2-L$\Sigma$}) when processing an input ($\mathbf{z}^0 = \mathbf{x}$): (1) prediction and laterally-weighted error computation, (2) error-correction of neural states. In this diagram, we depict a toy network with $3$ layers of $2$ state neurons (grey circles), i.e., $\mathbf{z}^2 = ([z^2_0, z^2_1 ])^T$, $\mathbf{z}^1 = ([z^1_0, z^1_1 ])^T$, $\mathbf{z}^0 = ([z^0_0, z^0_1 ])^T$, that are updated iteratively over $T$ time steps. Two of these layers are linked to error neurons (green diamonds), i.e., $\mathbf{e}^1 = ([e^1_0, e^1_1 ])^T$, $\mathbf{e}^0 = ([e^0_0, e^0_1 ])^T$, which compute mismatch messages that are propagated throughout the system. The bottom layer $\mathbf{z}^0$ receives sensory input, i.e., an image. %, such as the pixels of an image.
(b) The basic neural computational unit that \textcolor{black}{an NGC model} is composed of, consisting of a single state neuron $z^\ell_i$ and an error neuron $e^\ell_i$ at layer $\ell$. In the circuit, observe that a state neuron not only receives messages from error neurons (carried by $E^{\ell-1}$ synapses) but also a self-excitation signal and inhibition signals from laterally connected neurons (via $V^\ell$ synapses). The error neuron receives a gain signal (via $(\Sigma^\ell)^{-1}$ synapses) from laterally connected error neurons.  Note that filled diamonds indicate inhibitory signals, non-filled circles indicate excitatory signals, and empty squares indicate multiplicative signals.
\textcolor{black}{
(c) The different GNCN architectures (under the NGC framework) experimented with in this study. Solid black arrow represents generative weights, dashed black arrow represents error weights, and solid pink arrow represents error weights that are function of a specific set of generative weights (horizontal pink arrow illustrates which generative weights map to which error weights).
}
} 
% these different signals connect to concept of compartmental neurons too -- different portions represent different signals (consult O'Reilly as well...)
\label{fig:ngc_process}
\vspace{-0.4cm}
\end{figure}

% But the brain doesn't do X, Y, Z...
\textbf{Backprop-Learning versus Brain-Like Learning:} While the backprop algorithm described above has proven to be popular and effective in training ANNs (including generative models \cite{lecun2015deep}), it has certain mechanisms that differ from the current understanding of brain-like learning. For example, in backprop:
\begin{enumerate}[nosep]
    \item Synapses that make up the forward information pathway need to directly be used in reverse to communicate teaching signals (the weight transport problem \cite{grossberg1987competitive}), 
    \item Neurons need to be able to know about and communicate their own activation function's first derivative,
    \item Neurons must wait for the neurons ahead of them to percolate their error signals way back so they know when and how to adjust their own synapses (the update-locking problem \cite{jaderberg2017decoupled}),
    \item There is a distinct form of information propagation for error feedback, one that starts from the system's output and works its way back to the input layer (see Figure \ref{fig:bp_v_ngc} a), which does not influence neural activity (the global feedback pathway problem \cite{ororbia2019biologically}), i.e., backprop creates signals that only affect weights but do not (at least directly) affect/improve the network's representations of the environment,
    \item The error signals have a  one-to-one correspondence with  neurons.
\end{enumerate}
The goal of this paper is to present a \textcolor{black}{modeling and learning framework} that uses fewer mechanisms that are incompatible with current understanding of brain-like learning. Specifically, we aim to address the first 4 items.
\textbf{The Neural Generative Coding Approach:} 
In contrast to the backprop-based way of designing and training ANN generative models, our proposed framework, neural generative coding (NGC), provides one way to emulate the several neuro-biological principles and properties described above by proposing \textcolor{black}{a family of models and their} corresponding training procedure.
In this framework, \textcolor{black}{any single model is referred to as a} generative neural coding network (GNCN, \textcolor{black}{see Supplementary Note 1 for naming convention details).
A} GNCN model has $L+1$ layers of neurons (also called state variables) $\mathfrak{N}^0,\mathfrak{N}^1,\dots, \mathfrak{N}^L$, where $\mathfrak{N}^0$ is the output layer. 
Each layer $\mathfrak{N}^\ell$ has $J^\ell$ neurons and each neuron has a latent state value represented by a single number. The combined latent state of all neurons in layer $\mathfrak{N}^\ell$ is represented by the  vector $\latent{z}^\ell \in \mathcal{R}^{J_\ell \times 1}$ (initially $\latent{z}^\ell = \mathbf{0}$ when a new data point is encountered), with $\latent{z}^0$ being the same as the data $\mathbf{x}$ (it is typically clamped to the input, i.e., $\latent{z}^0 = \mathbf{x}$). 
The network is interpreted as a specification of the probability $P(\latent{z}^0=\mathbf{x},\latent{z}^1,\cdots,\latent{z}^L) = P(\latent{z}^0~|~\latent{z}^1)\cdots P(\latent{z}^{L-1}~|~\latent{z}^L)P(\latent{z}^L)$, similar to a Bayesian network, and we use the notation  $\mathbf{Z} = \{ \mathbf{z}^1,\cdots,\mathbf{z}^L \}$ to refer to the state of all of the \emph{intermediate} neurons (i.e., excluding the output). Thus, layer $\mathfrak{N}^\ell$ represents the conditional probability $P(\latent{z}^{\ell}~|~\latent{z}^{\ell+1})$, with the last layer $\mathfrak{N}^L$ representing $P(\latent{z}^L)$.
For image data, the distribution $P(\latent{z}^0~|~\latent{z}^1)$ associated with the output layer is multivariate Bernoulli with mean vector $\mean{z}^0$ (which depends on $\latent{z}^1$). All of the other distributions $P(\latent{z}^{\ell}~|~\latent{z}^{\ell+1})$ are multivariate Gaussians with mean vector $\mean{z}^\ell$ (which depends on $\latent{z}^{\ell+1}$) and covariance matrix $\mathbf{\Sigma}^{\ell}$.
The mean vector $\mean{z}^\ell$ for layer $\mathfrak{N}^\ell$ is obtained in a feed-forward manner from the latent state of the neighboring layer (biases/offset terms have been omitted for clarity)\textcolor{black}{. Specifically, we model this generative process with the following equation:}
\begin{align}
    \mean{z}^\ell &\gets g^\ell \Big( 
    \overbrace{ 
    \mathbf{W}^{\ell+1} \cdot  \phi^{\ell+1}(\latent{z}^{\ell+1})
    }^\text{local, top-down prediction}
    + \alpha_m 
    \overbrace{
    \big(\mathbf{M}^{\ell+2} \cdot  \phi^{\ell+2}(\latent{z}^{\ell+2}) \big)
    }^\text{local, auxiliary prediction}
    \Big),\label{eqn:ff}
\end{align}
% Another interesting property of the GNCN-PDH is that it does not contain a complementary error matrix for $\mathbf{M}^{\ell+2}$, meaning that, by design, its forward generative pathway is different from its error message transmission pathway (yielding an asymmetric network model when one considers the error correction synapses). 
% }
where $\mathbf{W}^\ell$ is a forward/generative weight matrix, \textcolor{black}{$\alpha_m = 0$,} $g^\ell$ and $\phi^{\ell+1}$ are activation functions, and $\cdot$ indicates matrix multiplication. Thus each layer $\mathfrak{N}^\ell$ is specified by two functions $g^\ell$ and $\phi^\ell$, a trainable weight matrix $\mathbf{W}^\ell$ and a covariance matrix $\mathbf{\Sigma}^\ell$ (the last layer $\mathfrak{N}^L$ is specified just by the activation function $\phi^L$). % and covariance matrix $\Sigma^L$).
Note in Equation \ref{eqn:ff}, the mean vector $\mean{z}^\ell$ depends on the sampled realization  $\latent{z}^{\ell+1}$ from the previous layer, making this a hierarchical, Gaussian latent variable model. \textcolor{black}{Notice that, optionally, if the binary coefficent $\alpha_m = 1$, each layer also incorporates a learnable auxiliary generative matrix
$\mathbf{M}^{\ell+2}$, which conveys and injects state value information from the layer $\mathbf{N}^{\ell+2}$ into the prediction of layer $\mathbf{N}^{\ell}$ through a linear combination. % before application of the prediction nonlinearity $g^\ell$. 
We append the suffix ``-PDH'' (partially decomposable hierarchy) to a GNCN model name when  $\alpha_m = 1$ (see Figure \ref{fig:ngc_arch} for a visual depiction).}

The goal of \textcolor{black}{any} GNCN model is to learn a joint distribution over its $L+1$ neural states, i.e., $p(\mathbf{z}^0,\mathbf{z}^1,\cdots,\mathbf{z}^L)$, from which the marginal distribution of the data may be obtained via $p(\mathbf{x}) = \int_\mathbf{Z} p(\mathbf{x},\mathbf{z}^1,\cdots,\mathbf{z}^L) d \mathbf{Z}=\int_\mathbf{Z} p(\mathbf{z}^0,\mathbf{z}^1,\cdots,\mathbf{z}^L) d \mathbf{Z}$. Given that minimizing $-\log p(\mathbf{x})$ directly is intractable in general, our approach for training is to approximately minimize the log-likelihood based on the ideas behind the Expectation-Maximization (EM) algorithm. Specifically, we work with the analogue of the  \emph{complete-data} likelihood, which adds in the latent variables of the network (it does not marginalize over them) and sets up a 2 step process that adjusts the latent variables (like an E-step) and then updates the network parameters (M step).

\noindent
\textbf{Training the Model:} 
The complete data log-likelihood $\psi$ (also referred to as total discrepancy \cite{ororbia2020continual}) of the observed data $\mathbf{x}$ and the latent variables $\latent{z}^1, \dots, \latent{z}^L$ is defined formally as follows:
\textcolor{black}{
\begin{align}
\psi &=    \sum_j \Big(\mathbf{x}_j\log \mean{z}^0_j + (1-\mathbf{x}_j)\log (1-\mean{z}^0_j)\Big) + \sum_{\ell=1}^L \left(-\frac{1}{2}\log|(\mathbf{\Sigma}^\ell)^{-1}| - \frac{1}{2}(\latent{z}^\ell - \mean{z}^\ell)^T \cdot (\mathbf{\Sigma}^\ell)^{-1} \cdot (\latent{z}^\ell - \mean{z}^\ell)\right) \mbox{.} \label{eqn:data_loglike}
\end{align}
}
Importantly, the above complete data log likelihood connects our NGC models (and total discrepancy) to the principle of free energy \cite{friston2006free,friston2008hierarchical}, given that the sum of (weighted) prediction errors defined in Equation \ref{eqn:data_loglike} can be shown to be a form of variational free energy that is minimized through variational inference. Explicitly characterizing a neural system engaged with optimizing the above objective as a generative model, as we do in this paper, grounds NGC as performing a form of approximate Bayesian inference (much as \cite{friston2008hierarchical} did for the early predictive coding model of \cite{rao1999predictive}) and connects it to perception as (unconscious) inference \cite{helmholtz1866concerning} and the larger theoretical framework of the Bayesian brain \cite{deneve2005bayesian,knill2004bayesian}.

Since all of the latent variables are continuous, the updates below follow the form of the exact gradient, i.e., backprop (allowing for gradient descent), 
%or alternatives like feedback alignment \cite{lillicrap2016random} or LRA \cite{ororbia2018conducting}, 
to optimize the latent variables and the parameters.
The log likelihood has the following partial derivatives:
\begin{align}
    \frac{\partial \psi}{\partial (\mathbf{\Sigma}^\ell)^{-1}} &= \frac{1}{2}\mathbf{\Sigma}^\ell - \frac{1}{2}(\latent{z}^\ell - \mean{z}^\ell) \cdot (\latent{z}^\ell - \mean{z}^\ell)^T \label{eqn:precision_update} \\
    \frac{\partial \psi}{\partial \mathbf{\Sigma}^\ell} &= -\frac{1}{2}\mathbf{\Sigma}^{-1} + \frac{1}{2}\mathbf{\Sigma}^{-1} \cdot (\latent{z}^\ell - \mean{z}^\ell) \cdot (\latent{z}^\ell - \mean{z}^\ell)^T \cdot \mathbf{\Sigma}^{-1} \label{covariance_update} \\
    \frac{\partial \psi}{\partial \mathbf{W}^0} &= \left(\frac{\partial g^0}{\partial \mathbf{h}^0} \odot(\mathbf{x} \oslash \mean{z}^0 - (\mathbf{1}-\mathbf{x})\oslash (\mathbf{1}-\mean{z}^0))\right) \cdot  \left(\phi^{1}(\latent{z}^{1})\right)^T  \label{eqn:W_0_update}\\
    &\phantom{=}\text{where }\oslash \text{ is element-wise division}, 
     \odot \text{ is element-wise product, and } \mathbf{h}^\ell =\mathbf{W}^\ell \cdot \phi^{\ell+1}(\latent{z}^{\ell+1})\nonumber\\
    \frac{\partial \psi}{\partial \mathbf{W}^\ell} &=\frac{\partial \psi}{\partial \mathbf{h}^\ell} \cdot  \left(\phi^{\ell+1}(\latent{z}^{\ell+1})\right)^T \label{eqn:W_l_update}\\
    \mathcolor{black}{
    \frac{\partial \psi}{\partial \latent{z}^1} }  &=  \mathcolor{black}{\frac{\partial}{\partial \latent{z}^1}\left( \sum_j \Big(\mathbf{x}_j\log \mean{z}^0_j + (1-\mathbf{x}_j)\log (1-\mean{z}^0_j)\Big)\right) }
                    - (\mathbf{\Sigma}^{\ell})^{-1} \cdot (\latent{z}^{1}-\mean{z}^{1})  \label{eqn:z_0_deriv}  \\
    \frac{\partial \psi}{\partial \latent{z}^\ell} &=  
                    \left( \frac{\partial \mean{z}^{\ell-1}}
                    {\partial \latent{z}^\ell} \cdot \Big( 
                    (\mathbf{\Sigma}^{\ell-1})^{-1} \cdot (\latent{z}^{\ell-1}-\mean{z}^{\ell-1}) \Big) \right)-
                    (\mathbf{\Sigma}^{\ell})^{-1} \cdot (\latent{z}^{\ell}-\mean{z}^{\ell}) \label{eqn:z_l_deriv}
\end{align}
In this work, we incorporate two key concepts from local representation alignment (LRA) \cite{ororbia2018conducting,ororbia2019biologically}: 1) the use of error synapses to directly resolve the weight-transport problem, and 2) the omission of derivatives of activation functions which yield synapse rules that function like error-Hebbian updates. To incorporate these modifications, we introduce a vector of error neurons $\mathbf{e}^\ell$ (depicted in Figure \ref{fig:bp_v_ngc} b) that are tasked with computing how far off the mean vector is from the relevant nearby state. Note that the error neurons themselves are derived from the likelihood function:
\begin{align}
    \mathbf{e}^0 &= \frac{\partial \psi }{\partial \mean{z}^0 } = (\mathbf{x} \oslash \mean{z}^0 - (\mathbf{1}-\mathbf{x})\oslash (\mathbf{1}-\mean{z}^0))  \label{eqn:e_0} \\
    \mathbf{e}^\ell &= \frac{\partial \psi }{\partial \mean{z}^\ell } =  \overbrace{ (\mathbf{\Sigma}^{\ell-1})^{-1} }^\text{lateral modulation} \cdot \overbrace{ (\latent{z}^{\ell-1}-\mean{z}^{\ell-1}) }^\text{mismatch signal} \label{eqn:e_l}
\end{align}
and are implemented as separate sets of activities (like the state neurons). In Figure \ref{fig:ngc_process} (b), we depict the circuit for a single pair neurons at a layer $\mathfrak{N}^\ell$, i.e., a state neuron $\mathbf{z}^\ell_i$ and an error neuron $\mathbf{e}^\ell_i$.
Notice that, in Equation \ref{eqn:e_l} above, in order to compute the state of the error neuron $e^\ell_i$, the covariance matrix $\mathbf{\Sigma}^\ell$ acts as a lateral modulation matrix, which is inspired by the neuro-mechanistic concept of precision weighting in predictive processing theory \cite{moran2013free}. It allows error neuron $e^\ell_i$ to dynamically amplify/reduce the learning signal (i.e., $z^\ell_i-\mean{z}^\ell_i$) of its corresponding state neuron $z^\ell_i$, based on the learning signals of the other state neurons. %serving as an implicit estimate of the layer's inverse variance (confidence).
Empirically, we found these modulatory synapses to improve the crispness of model samples. Note that Equation \ref{eqn:e_l} would be applied to all $J_\ell$ error neurons for each layer $\ell = 1, \cdots, L-1$.

\textbf{Updating the States and Synapses:} To carry the activities/messages of the error neurons in order to calculate the update to $\latent{z}^\ell$, we replace the term $\frac{\partial \latent{z}^{\ell-1}}{\partial \latent{z}^\ell}$ with a learnable error matrix $\mathbf{E}^\ell$. This substitution allows us to rewrite Equations \ref{eqn:z_0_deriv} and \ref{eqn:z_l_deriv} as follows:
\begin{align}
   \frac{\partial \psi}{\partial \latent{z}^1}  &\approx \Delta \latent{z}^1 = \mathbf{E}^1 \cdot \Big( \frac{g^0}{\mathbf{h}^0} \odot \mathbf{e}^0 \Big) - \mathbf{e}^1 = \mathbf{E}^1 \cdot \Big( \latent{z}^\ell - \mean{z}^\ell  \Big) - \mathbf{e}^1 \label{eqn:z_1_delta} \\
    \frac{\partial \psi}{\partial \latent{z}^\ell} &\approx \Delta \latent{z}^\ell =  \mathbf{E}^\ell \cdot \mathbf{e}^{\ell-1} - \mathbf{e}^\ell \mbox{.} \label{eqn:z_l_delta}
\end{align}
Once the update for any layer $\mathfrak{N}^\ell$ has been calculated, the corresponding state neurons will proceed to update their state values. By analogy with latent variable models, where the state neurons $\mathbf{z}^\ell$ correspond to latent variables, this act can be viewed as an attempt to modify the states in a way that improves the complete data log-likelihood (i.e. modifying the $\mathbf{z}^\ell$ to cause  $p(\mathbf{x},\mathbf{z}^1,\cdots,\mathbf{z}^L) $ to increase).
One possible neuroscience-inspired way to perform this update is shown in Equation \ref{eqn:naive_state} (here $\beta$ is a hyperparameter akin to the machine learning concept of a learning rate). Specifically, this update is:
\begin{align}
    \mathbf{z}^\ell_i \leftarrow \mathbf{z}^\ell_i + \beta \Big( - \gamma \mathbf{z}^\ell_i + \overbrace{ \sum_{j \in J_{\ell-1}} ( \mathbf{E}^{\ell}_{ij} \mathbf{e}^{\ell-1}_j ) - \mathbf{e}^\ell_i }^\text{$\Delta \mathbf{z}_i$} \Big) \mbox{.} \label{eqn:naive_state}
\end{align}
Here $z^\ell_i$ is modified through three terms. The first is a decaying pressure caused by the leak term $-z^\ell_i$, controlled by the strength factor $\gamma$. The second term, $-e^\ell_i$, can be interpreted as top-down pressure where $e^\ell_i$ is a measure of how much the neuron's state differs from the predicted state $\mean{z}^\ell$ that is computed by the layer above. Finally, the third term 
%We update $i^\text{th}$ state neuron's activity by subtracting out $e^\ell_i$'s influence (this is known as a top-down pressure, which communicates to $z^\ell_i$ how far off its own value is from the value guessed by the neurons above) and 
adds in the error message from each error neuron $e^{\ell-1}_j$ in the layer below, communicated by special error synapses $\mathbf{E}^{\ell-1}$, acting as a form of bottom-up pressure.
It is important to mention that, in this paper, we investigated another particular form of Equation \ref{eqn:naive_state} as follows:
\begin{align}
    \mathbf{z}^\ell_i \leftarrow \mathbf{z}^\ell_i + \beta \Big( - \gamma \mathbf{z}^\ell_i + \frac{\partial \phi(\mathbf{z}^\ell_i)}{\partial \mathbf{z}^\ell_i} \Big[ \sum_{j \in J_{\ell-1}} \big( \mathbf{U}^{\ell}_{ij} \mathbf{e}^{\ell-1}_j \big) - \mathbf{e}^\ell_i \Big] - \lambda \text{sign}(\mathbf{z}_i)  \Big) \mbox{.} \label{eqn:classical_state}
\end{align} % CHECK IF EQUATIONS ARE CORRECT
where the last term $ - \lambda \text{sign}(\mathbf{z}_i)$ is a kurtotic prior that can be imposed to encourage most activity values to be closer to zero in a given neural state $\latent{z}^\ell$.
\textcolor{black}{Under the NGC framework, models that use separate, learnable error synapses $\mathbf{E}^\ell$ will be referred to as ``Type 2'' (t2) and those that use (non-learnable) error synapses that are a function of the forward weights $\mathbf{W}^\ell$ will be labeled as ``Type 1'' (see Figure \ref{fig:ngc_arch} for visual depictions of these models and Supplementary Note 1 for details on the NGC framework model naming convention).}
We can then manipulate certain variable values to recover different classical predictive coding models: 1) if $\gamma = 0$, $\mathbf{U} = (\mathbf{W})^T$,  $\mathbf{\Sigma}^\ell = \sigma^2_z \mathbf{I}$, and $\phi^\ell(v) = \text{tanh}(v)$, then we recover the model proposed in \cite{rao1999predictive} \textcolor{black}{-- we will refer to this as GNCN-t1/Rao} (using $\gamma > 0$ yields the state equation of \cite{whittington2017approximation}), and 2) if $\gamma = 0$, $\mathbf{U} = -(\mathbf{W})^T$, and $\phi^\ell(v) = \text{max}(0,v)$, then we recover the neural implementation of the graphical model proposed in \cite{friston2008hierarchical} \textcolor{black}{-- we will refer to this as GNCN-t1-$\Sigma$/Friston (see Figure \ref{fig:ngc_arch})}.

\begin{figure}[!t]
\begin{center}
\includegraphics[width=6.25cm]{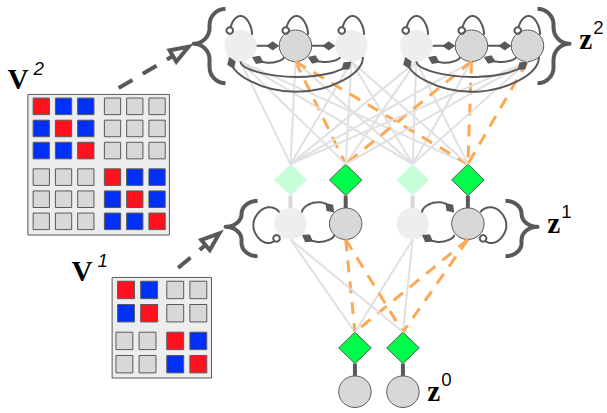}  
\end{center}
\vspace{-0.25cm}
\caption{
\textbf{Lateral dynamics of a generative neural coding network:}
Depending on how the lateral connectivity matrices, $\mathbf{V}^1$ \& $\mathbf{V}^2$, are designed, different competition patterns emerge among neurons in network. In the system with lateral matrices as shown above, each neuron is driven by its own self-excitation (red-colored blocks) and laterally inhibits (blue-colored blocks) other neurons that inhabit the same group (of $2$ or $3$ units). In the figure, one possible outcome (or state of the agent) of such a competition is shown (after $T$ steps). Self-excitation and lateral inhibition strengths are controlled by coefficients set a priori. Dashed orange edges show actively used synapses, filled diamonds indicate inhibitory signals, and non-filled circles indicate excitatory signals.} 
\label{fig:lateral_design}
\vspace{-0.35cm}
\end{figure}

However, instead of directly using this update rule, we may further incorporate another property of the brain -- activation sparsity (through lateral inhibition/excitation). Sparsity in real neuronal and artificial systems is often argued to be useful for learning compact representations (most activity values will be at/near zero), allowing for efficient storage and vastly improved energy efficiency.
%, but also allows the simultaneous representation of distinct patterns with little interference \cite{ahmad2016neurons} while still maintaining a large representational capacity. 
To emulate this type of sparsity, we integrate an explicit mechanism to force neurons to compete for activation (in contrast to the kurtotic prior used in Equation \ref{eqn:classical_state}), where we take inspiration from the known occurrence of lateral synapses in cortical regions of the brain (which are thought to facilitate contextual processing 
%and determine the properties of neural receptive fields
\cite{spratling2001dendritic}).
To do this, we introduce two terms to Equation \ref{eqn:naive_state} that use excitatory/inhibitory synapses stored in a matrix $\mathbf{V}^\ell$. This means that state neurons are updated as follows:
\begin{align}
    \mathbf{z}^\ell_i &\leftarrow \mathbf{z}^\ell_i + \beta \Big( \overbrace{-\gamma \mathbf{z}^\ell_i}^\text{leak} + \overbrace{ \Big( \sum_{j \in J_{\ell-1}} \mathbf{E}^{\ell}_{ij} \mathbf{e}^{\ell-1}_j \Big) - \mathbf{e}^\ell_i }^\text{bottom-up + top-down pressures} - \overbrace{ \Big( \sum_{j \in J_\ell, j \neq i} \mathbf{V}^\ell_{ij} \phi^\ell(\mathbf{z}^\ell_j) \Big) }^\text{lateral inhibition} + \overbrace{\mathbf{V}^\ell_{ii} \phi^\ell(\mathbf{z}^\ell_i)}^\text{self-excitation} \Big)  \label{eqn:state_unit}
\end{align}
Depending on the values set in $\mathbf{V}^\ell$, different types of sparsity patterns emerge, creating a flexible means for testing the benefits/drawbacks of different kinds of lateral competition patterns in an interpretable manner.
Figure \ref{fig:lateral_design} provides a graphical example of the type of interaction pattern we found worked well for the GNCN in this study, i.e., we forced $J_\ell/K$ groups (or columns) of $K$ neurons to compete with each other (see Supplementary Note 6). 
\textcolor{black}{The model employing Equation \ref{eqn:state_unit} and $\alpha_m = 0$ (in its Equation \ref{eqn:ff}) will be referred to as GNCN-t2-L$\Sigma$ and when $\alpha_m = 1$ (in its Equation \ref{eqn:ff}) the model will be referred to as the GNCN-PDH.}

The synaptic matrix updates in Equations \ref{eqn:W_0_update} and \ref{eqn:W_l_update} can be written in terms of the error neurons:
\begin{align}
    \frac{\partial \psi }{\partial \mathbf{W}^0} &\propto \Delta \mathbf{W}^0 = \mathbf{e}^0 \cdot (\phi^1( \mathbf{z}^1) )^T \label{eqn:W_0_delta} \\
     \frac{\partial \psi }{\partial \mathbf{W}^\ell} &\propto \Delta \mathbf{W}^\ell = \mathbf{e}^\ell \cdot (\phi^{\ell + 1}( \mathbf{z}^{\ell+1}) )^T \label{eqn:W_l_delta}
\end{align}
and, following in line with LRA, the error synapses can be updated as:
\begin{align}
    \Delta \mathbf{E}^\ell = \lambda \Big( \phi^{\ell+1}(\latent{z}^{\ell+1}) \cdot  (\mathbf{e}^\ell)^T \Big) \label{eqn:E_l_update}
\end{align}
where $\lambda \in [0,1]$ controls the strength of the error synaptic adjustment (usually set to values $< 1$).
%%%%%%%%%%%%
Note that no partial derivatives of the activation function $\phi^\ell$ is required \textcolor{black}{(in Type 2 GNCN models)} -- we are using an error-Hebbian style update rule, which has been shown to be effective empirically and crucially removes the need for state neurons to be aware of their own point-wise activation derivative. Note that this means $\phi^\ell$ could easily be a discrete function in this case, such as the signum or Heaviside.
%%%%%%%%%%%%%%%%

Putting all of the components above together, given a data point(s) $\mathbf{x}$, we can characterize the neural dynamics of the NGC system (graphically depicted in Figure \ref{fig:ngc_process} a) and train \textcolor{black}{any} GNCN following an online alternating maximization approach as follows:
\begin{enumerate}[noitemsep,nolistsep]
    \item[] // \textbf{Initialization:}
    \item Set $\latent{z}^0 = \mathbf{x}$ (clamp data to output neurons), and set $\latent{z}^\ell = \mathbf{0}, \; \forall \ell \ge 1$
    \item Compute mean estimates $\mean{z}^\ell, \; \forall \ell \ge 0$ (Equation \ref{eqn:ff}) \& use Equation \ref{eqn:e_0} and Equation \ref{eqn:e_l} to initialize $\mathbf{e}^\ell, \; \forall \ell \ge 0$
    \item[] // \textbf{Latent Update Step:} Search for most probable value of $\latent{z}^\ell, \forall \ell$
    \textcolor{black}{
    \item Use Equation \ref{eqn:z_1_delta} to get $\Delta \latent{z}^1$ \& Equation \ref{eqn:z_l_delta} to get $\Delta \latent{z}^\ell, \; \forall \ell > 1$. States are modified via (vectorized form of Equation \ref{eqn:state_unit}):}
    \begin{align*}
        \latent{z}^1 &= \latent{z}^1 + \beta \Big( \Delta \latent{z}^1 - \mathbf{V}^1 \cdot \latent{z}^1 - \gamma \latent{z}^1 \Big) \\
        \latent{z}^\ell &= \latent{z}^\ell + \beta \Big( \Delta \latent{z}^\ell - \mathbf{V}^\ell \cdot \latent{z}^\ell - \gamma \latent{z}^\ell \Big), \; \forall \ell > 1
    \end{align*}
    \item Run Equation  \ref{eqn:e_0} and  \ref{eqn:e_l} to obtain updated error neurons, i.e., $\mathbf{e}^\ell, \; \forall \ell \ge 0$
    \item Repeat Steps 3 and 4 for $T$ iterations
    \item[] // \textbf{Parameter Update Step:} Update parameters given estimated latent states
    \item Update synaptic matrices using Equations \ref{eqn:W_0_delta}, \ref{eqn:W_l_delta}, \ref{eqn:precision_update}, and \ref{eqn:E_l_update} (and normalize current matrices) \mbox{.}
\end{enumerate}
At test time, to reconstruct $\mathbf{x}$, one should only use Steps 1-5 of the recipe above. $\gamma$ controls the strength of the decay/leak applied to $\mathbf{z}^\ell$ and corresponds to placing an additional $\mathcal{N}(\mu=0,\Sigma=\lambda \mathbf{I})$ prior over the latent states.
In essence, the above complete process depicts that the GNCN adjusts it synaptic weight matrices once activity values for all $\mathbf{z}^\ell$ and $\mathbf{e}^\ell$ have been found after a $T$-step episode. The synaptic matrix updates are simply the products between relevant state activities and error neurons.
Finally, after each weight update \textcolor{black}{has been} made, a GNCN's weight matrices are normalized such that the Euclidean norms of their columns are $1.0$ (this step, which requires non-local information to perform, could possibly be induced biologically through the use of external neuromodulatory signals).
In essence, the steps described above and shown in Figure \ref{fig:ngc_process} (a) illustrate that the NGC framework embodies the idea that neural state and synaptic weight adjustment are the result of a process of generate-then-correct, or \emph{continual error correction}, in response to samples of the agent's environment.

% \textcolor{black}{
% \textbf{The Prediction Equation for the GNCN-PDH: }
% In the GNCN-PDH, the mean vector $\mean{z}^\ell$ for any layer $\mathfrak{N}^\ell$ is obtained in a feed-forward manner from the latent state of the neighboring two layers (biases/offset terms have been omitted for clarity):
% \begin{align}
%     \mean{z}^\ell \gets g^\ell(\mathbf{W}^{\ell+1} \cdot  \phi^{\ell+1}(\latent{z}^{\ell+1}) + \mathbf{M}^{\ell+2} \cdot  \phi^{\ell+2}(\latent{z}^{\ell+2})) ,\label{eqn:ff_pdh}
% \end{align}
% where we observed that an additional, learnable auxiliary generative matrix $\mathbf{M}^{\ell+2}$ conveys and injects state value information from the layer $\mathbf{N}^{\ell+2}$ into the prediction of layer $\mathbf{N}^{\ell}$ through a simple linear combination before application of the prediction nonlinearity $g^\ell$. Another interesting property of the GNCN-PDH is that it does not contain a complementary error matrix for $\mathbf{M}^{\ell+2}$, meaning that, by design, its forward generative pathway is different from its error message transmission pathway (yielding an asymmetric network model when one considers the error correction synapses). 
% }

\begin{table}[!t]
\begin{center}
\caption{
\textbf{Reconstruction \& Likelihood Measurements:}
Generative modeling results across datasets. The binary cross entropy (BCE) of the model reconstructions and the marginal log likelihood $\log p(\mathbf{x})$ (in nats) on the test set are reported (lower BCE is better and $\log p(\mathbf{x})$ closer to $0.0$ is better). We \textbf{bold} the best two scores of the models with respect to BCE (first column) as well as with respect to $\log p(\mathbf{x})$ (second column).
(Metrics averaged over $10$ trials - we report their mean and standard deviation.)
%The single horizontal separates the backprop-based models (above the line) from the NGC/predictive processing models (below the line).
}
\label{results:gen_results}
 \begin{tabular}{l | l c c | l c c} 
 \hline
  % & \multicolumn{3}{c}{\textbf{MNIST}} & \multicolumn{3}{c}{\textbf{KMNIST}} \\
 \textbf{Model} & & \multicolumn{1}{c}{\textbf{BCE}} & \multicolumn{1}{c}{$\log p(\mathbf{x})$}  & & \multicolumn{1}{c}{\textbf{BCE}} & \multicolumn{1}{c}{$\log p(\mathbf{x})$} \\ %[0.5ex] 
 \hline\hline
  GMM & \parbox[t]{0.15mm}{\multirow{7}{*}{\rotatebox[origin=c]{90}{MNIST}}} & -- & $-185.37 \pm 0.47$ &  \parbox[t]{0.15mm}{\multirow{7}{*}{\rotatebox[origin=c]{90}{KMNIST}}} & -- & $-426.73 \pm 1.03$ \\
 RAE & & $60.946 \pm 0.58$ & $-117.009 \pm 0.181$ &  & $175.597 \pm 0.506$ & $-234.411 \pm 1.173$ \\
 GVAE-CV & & $69.904 \pm 0.304$ & $-106.413 \pm 0.057$ & & $180.812 \pm 1.054$ & $-229.124 \pm 0.946$ \\
 GVAE & & $75.521 \pm 2.194$ & $-100.097 \pm 2.257$ & & $199.374 \pm 0.913$ & $\mathbf{-218.67 \pm 0.43}$ \\
 GAN-AE & & $73.54 \pm 3.492$ & $\mathbf{-94.418 \pm 0.069}$ & & $199.677 \pm 3.014$ & $\mathbf{-216.763 \pm 1.156}$ \\
 %\hline
 \textcolor{black}{GNCN-t1/Rao} & & $66.466 \pm 0.065$ & $-102.222 \pm 0.016$ & & $136.492 \pm 0.149$ & $-226.371 \pm 0.187$ \\
 \textcolor{black}{GNCN-t1-$\Sigma$/Friston} & & $\bf 47.246 \pm 0.191$ & $-101.921 \pm 0.242$ & & $\bf 114.629 \pm 0.353$ & $-229.037 \pm 0.517$ \\
 \textcolor{black}{GNCN-t2-L$\Sigma$} & & $59.658 \pm 0.021$ & $-98.271 \pm 0.247$ & & $135.864 \pm 0.165$ & $-220.41 \pm 0.223$ \\
 GNCN-PDH & & $\bf 44.857 \pm 0.054$ & $\bf -97.255 \pm 0.084$ & & $\bf 128.389 \pm 0.23$ & $-219.82 \pm 0.641$ \\
 \hline
 \hline
 GMM & \parbox[t]{0.15mm}{\multirow{7}{*}{\rotatebox[origin=c]{90}{FMNIST}}} & -- & $-302.96 \pm 0.47$ & \parbox[t]{0.15mm}{\multirow{7}{*}{\rotatebox[origin=c]{90}{CalTech}}} & -- & $-88.63 \pm 0.43$  \\
 RAE & & $102.723 \pm 0.777$ & $-138.762 \pm 1.427$  & & $40.408 \pm 0.755$ & $-50.47 \pm 2.073$ \\
 GVAE-CV & & $111.719 \pm 0.134$ &  $-131.816 \pm 0.295$ &  & $49.579 \pm 4.588$ & $-41.339 \pm 0.84$ \\
 GVAE & & $121.618 \pm 0.453$ & $\mathbf{-127.644 \pm 0.041}$  & & $38.349 \pm 0.74$ & $-44.067 \pm 0.24$ \\
 GAN-AE & & $132.943 \pm 4.372$ & $-130.872 \pm 1.795$ & & $40.396 \pm 0.743$ & $\bf -40.684 \pm 0.131$ \\
 %\hline
 \textcolor{black}{GNCN-t1/Rao} & & $95.803 \pm 0.025$ & $-136.632 \pm 0.465$ & & $34.595 \pm 0.062$ & $-44.621 \pm 0.128$ \\
 \textcolor{black}{GNCN-t1-$\Sigma$/Friston} & & $\bf 81.734 \pm 0.421$ & $-133.506 \pm 0.267$ & & $\bf 31.169 \pm 0.023$ & $-42.526 \pm 0.059$ \\
 \textcolor{black}{GNCN-t2-L$\Sigma$} & & $97.802 \pm 0.714$ & $-132.041 \pm 0.312$ & & $33.287 \pm 0.011$ & $-41.523 \pm 0.058$ \\
 GNCN-PDH & & $\bf 90.058 \pm 0.375$ & $\bf -130.031 \pm 0.162$ & & $\bf 26.764 \pm 0.02$ & $\bf -40.821 \pm 0.195$ \\
 \hline
\end{tabular}% 
\end{center}
\vspace{-0.5cm}
\end{table}

\begin{figure}[!t]
\begin{center}
\begin{subfigure}{0.5\textwidth}
  \centering
  \includegraphics[width=6.25cm]{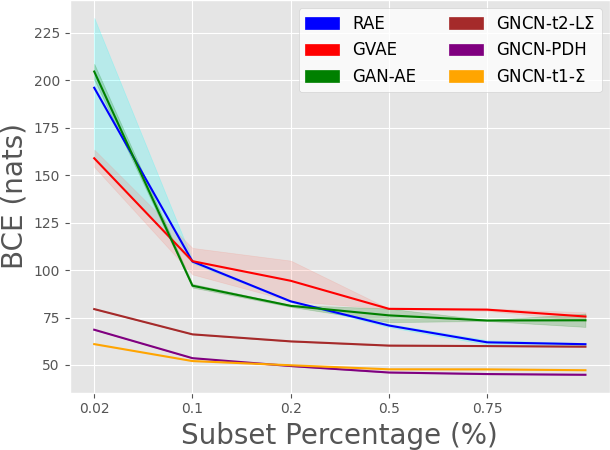}  
  \caption{MNIST.}
  \label{fig:mnist_converge}
\end{subfigure}%
\begin{subfigure}{0.5\textwidth}
  \centering
  \includegraphics[width=6.25cm]{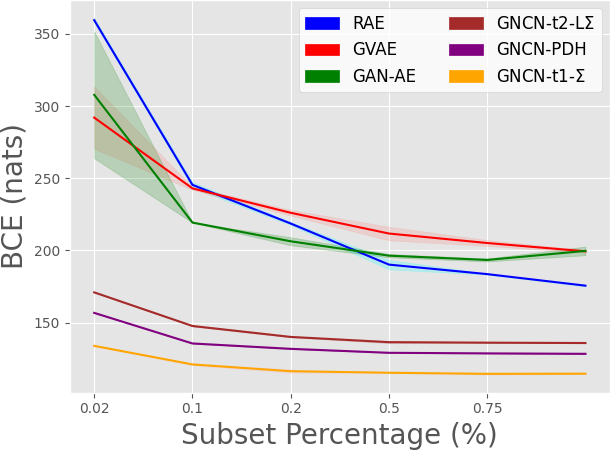}  
  \caption{KMNIST.}
  \label{fig:kmnist_converge}
\end{subfigure}%
\\
\begin{subfigure}{0.5\textwidth}
  \centering
  \includegraphics[width=6.25cm]{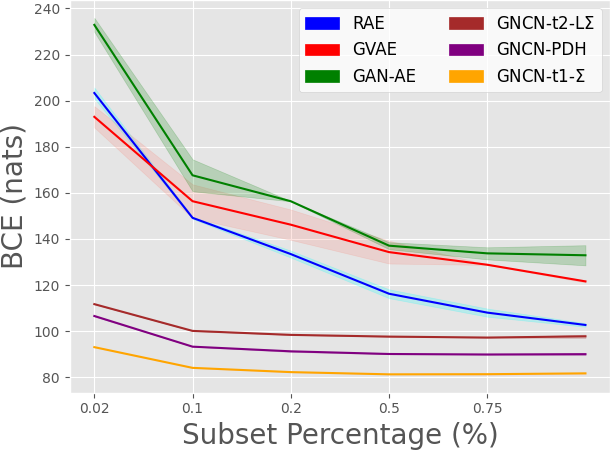}  
  \caption{FMNIST.}
  \label{fig:fmnist_converge}
\end{subfigure}%
\begin{subfigure}{0.5\textwidth}
  \centering
  \includegraphics[width=6.25cm]{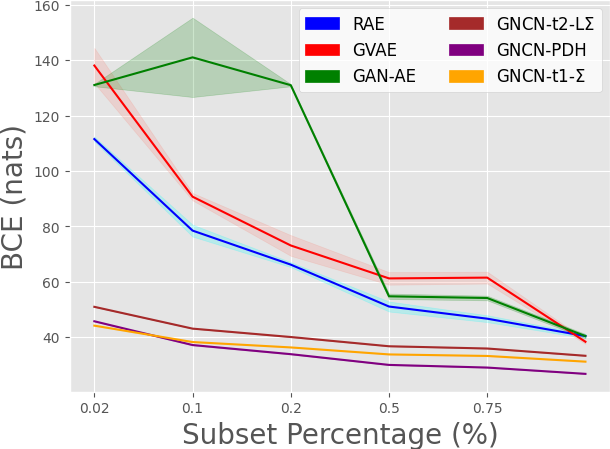}
  \caption{CalTech.}
  \label{fig:ctech_converge}
\end{subfigure}
\end{center}
\vspace{-0.45cm}
\caption{
\textcolor{black}{
\textbf{Data efficiency measurements:} 
Test BCE loss measurements (lower is better) for pattern reconstruction on data subsets of increasing size (averaged over $10$ trials). Curves depict the mean (solid line) and the standard deviation (lighter-colored envelope) over $10$ trials. (Note that GNCN-t1-$\Sigma$ is also referred to as GNCN-t1-$\Sigma$/Friston.)
}
}
\label{fig:convergence}
\vspace{-0.5cm}
\end{figure}

% Results - first, auto-associative generative modeling
\textbf{Generative Modeling Results:} The model framework that we have described so far would be immediately useful for creating an auto-associative memory of sensory input, i.e., upon receiving a particular sensory input, the model would be able to recall seeing it by accurately reconstructing it. Furthermore, since the model is learning an estimator of the input distribution $p(\mathbf{x})$, we may sample from it (see Supplementary Methods) to synthesize or ``hallucinate'' data patterns, as we will demonstrate later in this section.

%We will show that the NGC framework learns a viable generative and auto-associative model of data. 
To evaluate our framework (see Supplementary Methods), nine approaches were compared across four image datasets (Table \ref{results:gen_results}), each of which contained a training subset (from which we created an additional validation subset) and a test subset (see Supplementary Methods). 
One is a Gaussian mixture model (GMM) and eight are neural models -- four of these are backprop-based (see Supplementary Methods and Supplementary Note 6)  
\textcolor{black}{and four are NGC models, i.e., a GNCN-t1/Rao (which is equivalent to the model of \cite{rao1999predictive} and \cite{whittington2017approximation}), a GNCN-t1-$\Sigma$/Friston (which is equivalent to the model of \cite{friston2008hierarchical}), a GNCN-t2-L$\Sigma$, and a GNCN-PDH.}
%, two are related predictive coding (PC) models, i.e., the model proposed in \cite{rao1999predictive} (PC-Rao) and the model proposed in \cite{friston2008hierarchical} (PC-Friston), and two are NGC-based. 
All neural models were constrained to have their top-most layer to contain $20$ processing elements (the \textcolor{black}{GNCN-t2-L$\Sigma$} and GNCN-PDH were restricted to $20$ neural columns) and all had approximately the same total number of synapses. With respect to NGC models, $T = 50$ (see Supplementary Note 4).
The regularized auto-encoder (RAE), the auto-associative network least equipped to serve as a data synthesizer, reaches lower reconstruction error, in terms of binary cross entropy (BCE), compared to the other backprop-based generative models, i.e., the Gaussian variational autoencoder (GVAE), the constant-variance GVAE (CV-GVAE), and the adversarial autoencoder (GAN-AE). However, while the CV-GVAE, GVAE, and GAN-AE models yield worse reconstruction than the RAE, they obtain much better log likelihoods, especially compared to the GMM baseline, indicating that they strong data samplers. It makes sense that these models obtain better likelihood at the expense of BCE given that their optimization objective imposes a strong pressure to craft a proper (Gaussian) distribution over latent variables in addition to reconstructing data samples (in the case of the GAN-AE, the pressure comes from forcing the discriminator to distinguish between fake and real latent variables).
Interestingly enough, we see \textcolor{black}{
that the GNCN-t2-L$\Sigma$ and GNCN-PDH obtains competitive log likelihood with the CV-GVAE, GVAE, and GAN-AE with the GNCN-PDH yielding the best log likelihood out of all GNCN models for all four datasets.  Notably, the GNCN models result in the best reconstruction across all four datasets, with the GNCN-t1-$\Sigma$/Friston and GNCN-PDH offering the lowest BCE.
}
On MNIST, we note that the \textcolor{black}{GNCN-PDH} outperforms some other related prior models \cite{bengio2013generalized} that used our same experimental setup, e.g., a restricted Boltzmann machine with $\log p(\mathbf{x}) = -112$ nats %RBM+blur
, a denoising autoencoder with $\log p(\mathbf{x}) = -142$ nats % DAE
(trained via backprop) and $\log p(\mathbf{x}) = -116$ nats (using the walk-back algorithm). As indicated by our results, an \textcolor{black}{NGC model's} sampling and reconstruction ability is very promising. %, though we provide commentary on how it could be further improved in Methods.
\textcolor{black}{Worthy of note, however, is that we find that backprop-based autoencoders appear to do better at matching the data's class frequency than the GNCNs studied in this article (see Supplementary Note 3).}

\begin{figure}[!t] %[!tbp]
\centering
\minipage{0.225\textwidth}%
  \includegraphics[width=\linewidth]{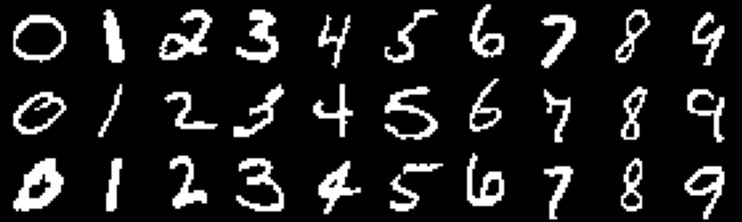}\\
  \vspace{-0.26cm}
  \includegraphics[width=\linewidth]{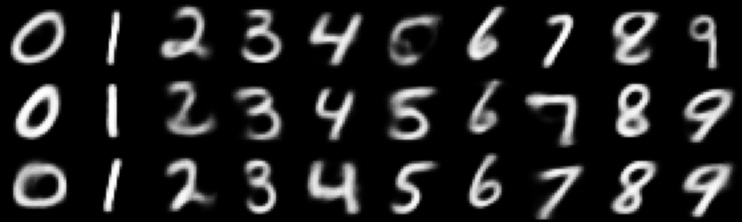}\\
  \vspace{-0.26cm}
  \includegraphics[width=\linewidth]{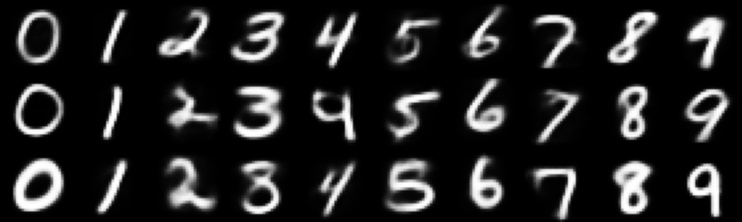}\\
  %\caption{A really Awesome Image}\label{fig:awesome_image1}
\endminipage\hspace{0.125cm}
\minipage{0.225\textwidth}%
  \includegraphics[width=\linewidth]{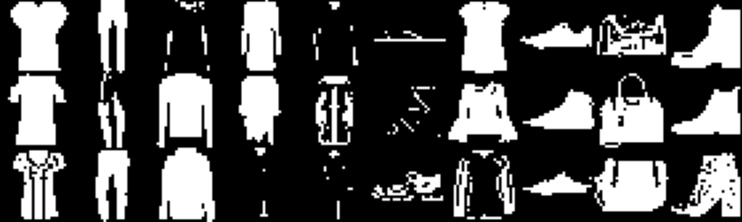}\\
  \vspace{-0.26cm}
  \includegraphics[width=\linewidth]{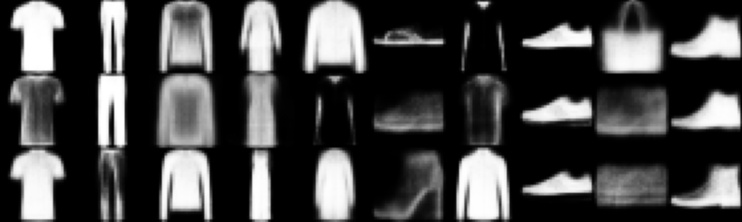}\\
  \vspace{-0.26cm}
  \includegraphics[width=\linewidth]{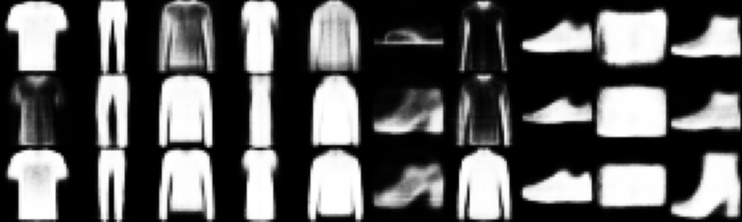}\\
  %\caption{A really Awesome Image}\label{fig:awesome_image2}
\endminipage\hspace{0.125cm}
\minipage{0.225\textwidth}%
  \includegraphics[width=\linewidth]{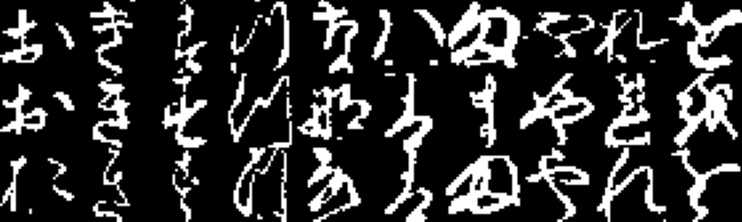}\\
  \vspace{-0.26cm}
  \includegraphics[width=\linewidth]{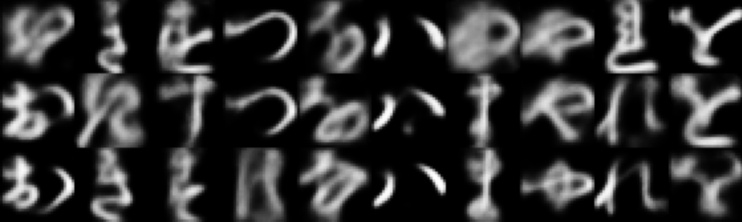}\\
  \vspace{-0.26cm}
  \includegraphics[width=\linewidth]{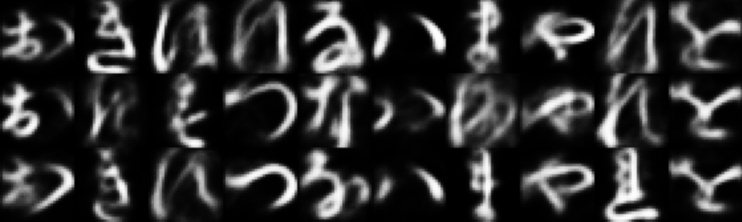}\\
  %\caption{A really Awesome Image}\label{fig:awesome_image3}
\endminipage\hspace{0.125cm}
\minipage{0.225\textwidth}%
  \includegraphics[width=\linewidth]{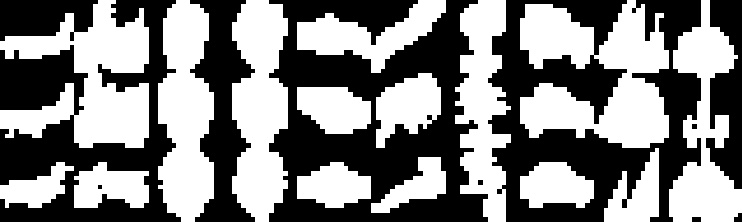}\\
  \vspace{-0.26cm}
  \includegraphics[width=\linewidth]{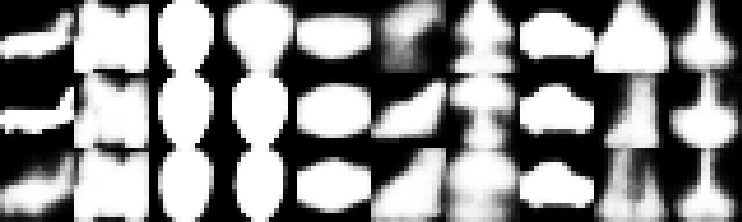}\\
  \vspace{-0.26cm}
  \includegraphics[width=\linewidth]{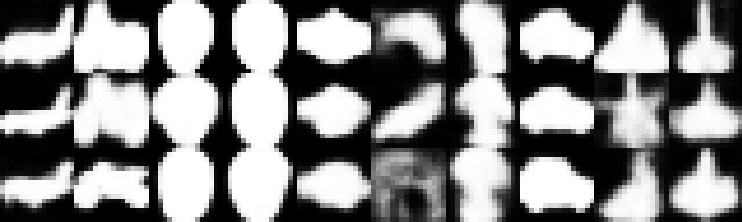}\\
  %\caption{A really Awesome Image}\label{fig:awesome_image3}
\endminipage
\vspace{-0.35cm}
\caption{
\textbf{Model sample visualization:} 
Shown are samples from models with overall best BCE-$\log p(\mathbf{x})$ balance (each column corresponds to one class). Top row shows dataset samples, middle row shows adversarial autoencoder (GAN-AE) samples, and bottom row shows generative neural coding network (GNCN-t2-L$\Sigma$) samples. Columns are arranged, left to right, as: a) MNIST, b) KMNIST, c) FMNIST, and d) CalTech.}
\label{fig:model_class_samples}
\vspace{-0.4cm}
\end{figure}

As shown in Figure \ref{fig:convergence}, we measured the data efficiency of the \textcolor{black}{GNCN-t1-$\Sigma$/Friston, GNCN-t2-L$\Sigma$, and GNCN-PDH as well as} several representative backprop-models (RAE, GVAE, and GAN-AE). Specifically, we train each model (to convergence) using a subset of the training data of varying size (we created six versions of each dataset's training subset using either $2\%$, $10\%$, $20\%$, $50\%$, $75\%$, or $100\%$) of the original sample), and plot final test BCE at convergence (versus percentage of original data used on the x-axis), for each of the models. 
Desirably, the result demonstrates that the predictive processing \textcolor{black}{NGC models} generalize better given varying amounts of data, even with less data compared to the autoencoder models.
%, with the GNCN and GNCN-PDH being among the most performant overall. 
This benefit is useful given that it offsets the higher per-sample processing cost of the predictive processing models.
%As shown in Figure \ref{fig:convergence}, where the final model test BCE was measured throughout training, the GNCN consistently generalizes sooner than all of the backprop-based models (similar loss curve patterns were observed on the training set). Even after only passing through the dataset once, the GNCN reaches much lower BCE than the others. For example, on MNIST, the GNCN reached a $\mbox{BCE} = 57$ nats by the end of the first pass while the GVAE reached $\mbox{BCE} = 194$ nats and the RAE reached $\mbox{BCE} = 168$ nats.
Furthermore, in Figure \ref{fig:model_class_samples}, we present random samples from each class obtained from either: 1) the original dataset, 2) ancestrally sampling the GAN-AE, or 3) ancestrally sampling \textcolor{black}{the GNCN-t2-L$\Sigma$} (see Supplementary Table 1 for additional visualization of nearest-neighbor samples that match an original data point for each class). Note that we trained well-regularized MLP classifier trained on the original dataset and then used it to automatically annotate the samples produced by each model.
%\footnote{Specifically, for the GVAE and GNCN models, we display the Bernoulli mean parameter values (produced at the output layer of each model) which is then projected back to pixel space as is typically done with generative models in the machine learning community.}
Observe that both the \textcolor{black}{GNCN-t2-L$\Sigma$} and GAN-AE yield reasonably good-looking sample images for all four datasets and the differences in perceptual quality between the models' sets of samples is marginal. This is encouraging, given that our goal was to demonstrate that \textcolor{black}{an NGC model} could be competitive with backprop-based generative models.

\begin{table}[t]
\begin{center}
\caption{
\textbf{Downstream performance across datasets:}
(a) The classification error (Err) of a log-linear model fit to each model's latent codes and the masked mean squared error (M-MSE) of the model's pattern completion ability on the test set are reported (lower is better -- two best scores are in \textbf{bold} with respect to each metric/column). (Metrics averaged over $10$ trials - we report their mean and standard deviation.) 
(b) Masked patterns in first row and pattern completions from the GNCN-PDH in the second row. }
\label{results:diagonistics}
 \begin{tabular}{l | l c c | l c c } 
 \hline
  % & \multicolumn{3}{c}{\textbf{MNIST}} & \multicolumn{3}{c}{\textbf{KMNIST}} \\
 (a) \textbf{Model} & & \multicolumn{1}{c}{\textbf{Err} (\%)} & \multicolumn{1}{c}{\textbf{M-MSE}} &  & \multicolumn{1}{c}{\textbf{Err} (\%)} & \multicolumn{1}{c}{\textbf{M-MSE}}  \\ %[0.5ex] 
 \hline\hline
 %Baseline & \parbox[t]{0.15mm}{\multirow{5}{*}{\rotatebox[origin=c]{90}{MNIST}}} & $90$ & $28.85$ &
 DSRN & \parbox[t]{0.15mm}{\multirow{5}{*}{\rotatebox[origin=c]{90}{MNIST}}} & $\mathbf{1.93 \pm 0.04}$ & -- & \parbox[t]{0.15mm}{\multirow{5}{*}{\rotatebox[origin=c]{90}{KMNIST}}} & $\mathbf{10.02 \pm 0.08}$ & -- \\
 %PCN & & $X$ & -- & &  $X$ &  -- \\
 RAE & & $11.99 \pm 0.47$ & $22.10 \pm 1.44$ & & $39.32 \pm 0.95$ & $30.69 \pm 0.45$ \\
 GVAE-CV & & $17.25 \pm 0.50$ & $20.37 \pm 0.15$ & & $40.22 \pm 0.05$ & $23.69 \pm 1.61$ \\
 GVAE & & $9.13 \pm 0.18$ & $20.06 \pm 0.57$ & & $48.09 \pm 2.32$ & $31.04 \pm 0.26$ \\
 GAN-AE & & $13.03 \pm 1.68$ & $15.87 \pm 0.24$ & & $38.66 \pm 0.87$ & $22.97 \pm 0.94$ \\
 %\hline
 \textcolor{black}{GNCN-t1/Rao} & & $8.57 \pm 0.03$ & $4.54 \pm 0.01$ & & $30.75 \pm 0.06$ & $9.16 \pm 0.022$ \\
 \textcolor{black}{GNCN-t1-$\Sigma$/Friston} & & $5.40 \pm 0.01$ & $\bf 3.11 \pm 0.02$ & & $22.89 \pm 0.59$ & $\bf 7.64 \pm 0.08$ \\
 \textcolor{black}{GNCN-t2-L$\Sigma$} & & $2.38 \pm 0.04$ & $4.15 \pm 0.02$ & & $\bf 15.45 \pm 0.43$ & $9.23 \pm 0.02$ \\
 GNCN-PDH & & $\bf 2.28 \pm 0.04$ & $\bf 3.04 \pm 0.02$ & & $17.01 \pm 0.32$ & $\bf 8.43 \pm 0.04$ \\
 \hline
 \multicolumn{5}{c}{ } \\
 % & \multicolumn{3}{c}{\textbf{FMNIST}} & \multicolumn{3}{c}{\textbf{CalTech}} \\
 \hline
 % Baseline & \parbox[t]{0.15mm}{\multirow{5}{*}{\rotatebox[origin=c]{90}{FMNIST}}} & $90$ & $67.79$ & \parbox[t]{0.15mm}{\multirow{5}{*}{\rotatebox[origin=c]{90}{CalTech}}} & $99$ & $33.30$ \\
 DSRN & \parbox[t]{0.15mm}{\multirow{5}{*}{\rotatebox[origin=c]{90}{FMNIST}}} & $\mathbf{9.75 \pm 0.09}$ & -- & \parbox[t]{0.15mm}{\multirow{5}{*}{\rotatebox[origin=c]{90}{CalTech}}} & $\mathbf{31.56 \pm 0.01}$ & -- \\
 %PCN & & $X$ &  -- & &  $X$ &  -- \\
 RAE & & $25.14 \pm 0.24$ & $56.83 \pm 0.73$ & & $38.10 \pm 0.49$ & $23.56 \pm 0.91$ \\
 GVAE-CV & & $24.19 \pm 0.24$ & $53.46 \pm 2.39$ & & $38.64 \pm 1.69$ & $27.22 \pm 1.43$ \\
 GVAE & &  $30.27 \pm 0.57$ & $50.81 \pm 4.63$ & & $40.36 \pm 0.43$ & $24.08 \pm 2.31$ \\
 GAN-AE & & $24.67 \pm 0.35$ & $42.31 \pm 3.82$ & & $37.34 \pm 1.01$ & $19.47 \pm 0.88$ \\
 %\hline
 \textcolor{black}{GNCN-t1/Rao} & & $20.28 \pm 0.01$ & $6.81 \pm 0.01$ & & $34.23 \pm 0.01$ & $2.11 \pm 0.03$ \\
 \textcolor{black}{GNCN-t1-$\Sigma$/Friston} & & $17.78 \pm 0.12$ & $\bf 6.01 \pm 0.04$ & & $30.60 \pm 0.28$ & $\bf 1.97 \pm 0.01$ \\
 \textcolor{black}{GNCN-t2-L$\Sigma$} & & $\bf 16.85 \pm 0.04$ & $7.46 \pm 0.05$ & & $\bf 29.11 \pm 0.37$ & $2.23 \pm 0.04$ \\
 GNCN-PDH & & $17.11 \pm 0.26$ & $\bf 6.01 \pm 0.02$ & & $29.54 \pm 0.18$ & $\bf 1.82 \pm 0.01$ \\
 \hline
\end{tabular}% 
\begin{tabular}{cccc}
%\hline
%\multicolumn{4}{c}{\textit{GNCN Completions}} \\
\hline
\multicolumn{2}{c}{(b) \hspace{0.115cm} \textbf{MNIST}} & \multicolumn{2}{c}{\textbf{KMNIST}} \\
 \includegraphics[width=0.055\linewidth]{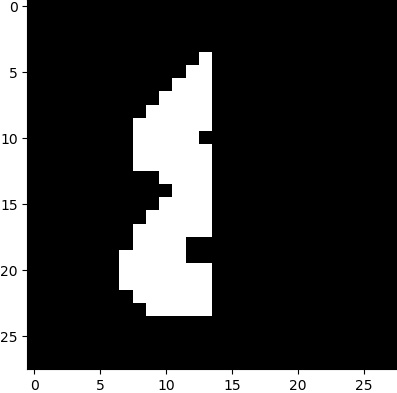}\hspace{-0.35cm} & \includegraphics[width=0.055\linewidth]{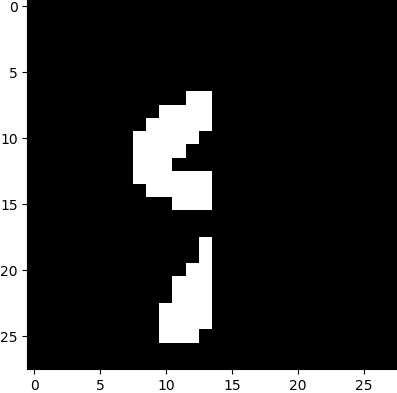}\hspace{-0.35cm} & \includegraphics[width=0.055\linewidth]{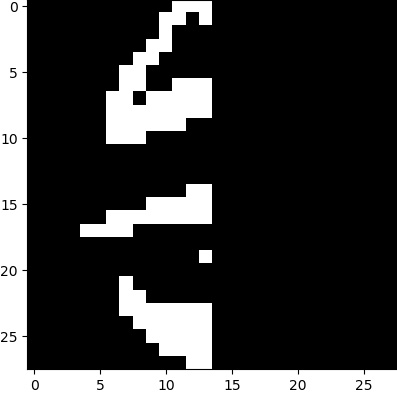}\hspace{-0.35cm} & \includegraphics[width=0.055\linewidth]{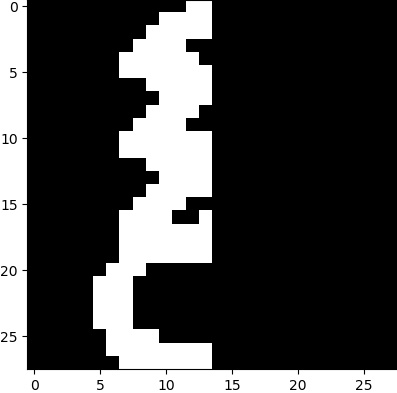} \\
 \includegraphics[width=0.055\linewidth]{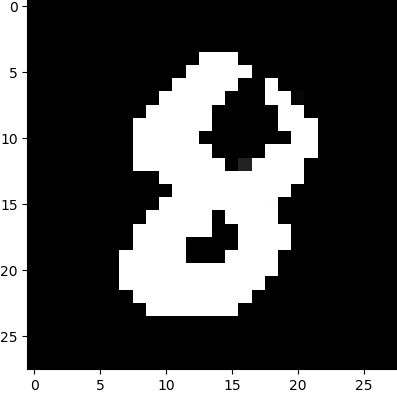}\hspace{-0.35cm} & \includegraphics[width=0.055\linewidth]{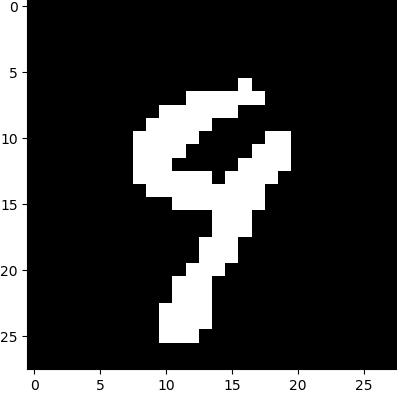}\hspace{-0.35cm} & \includegraphics[width=0.055\linewidth]{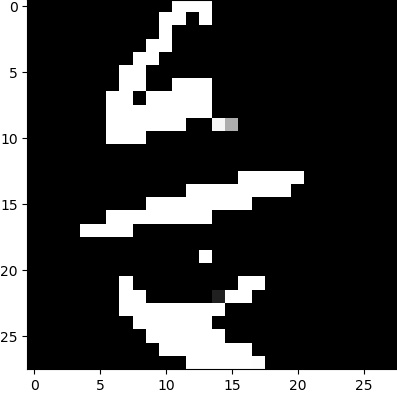}\hspace{-0.35cm} & \includegraphics[width=0.055\linewidth]{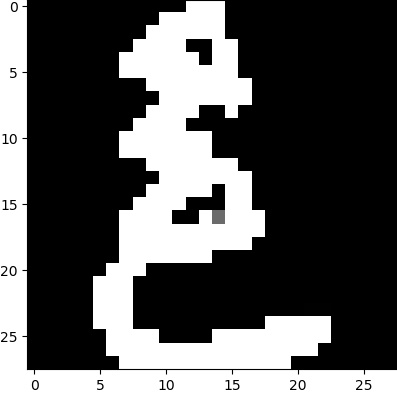} \\
 \hline
 \multicolumn{4}{c}{} \\
 \hline 
 \multicolumn{2}{c}{\hspace{0.25cm}\textbf{FMNIST}} & \multicolumn{2}{c}{\textbf{CalTech}} \\
 \includegraphics[width=0.055\linewidth]{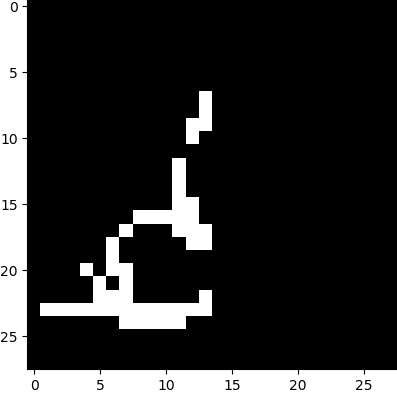}\hspace{-0.35cm} & \includegraphics[width=0.055\linewidth]{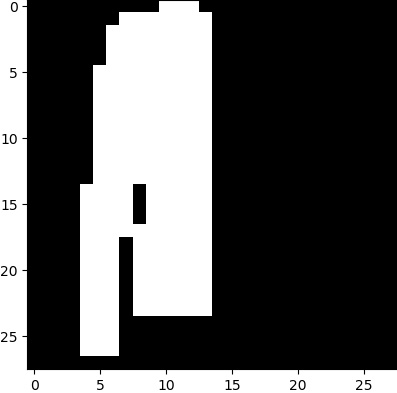}\hspace{-0.35cm} & \includegraphics[width=0.055\linewidth]{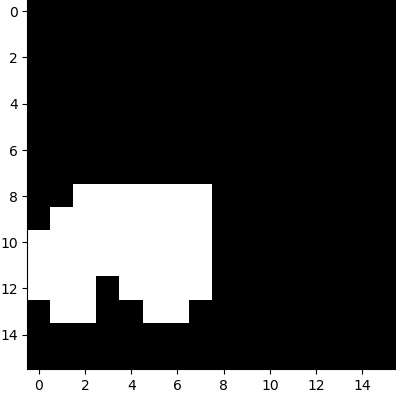}\hspace{-0.35cm} & \includegraphics[width=0.055\linewidth]{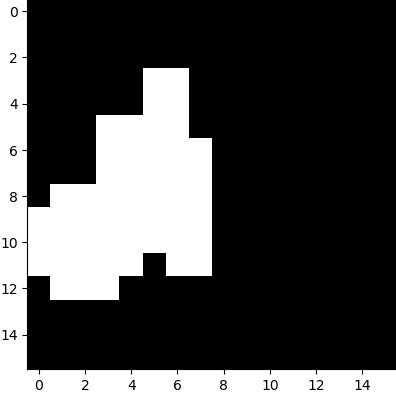} \\
 \includegraphics[width=0.055\linewidth]{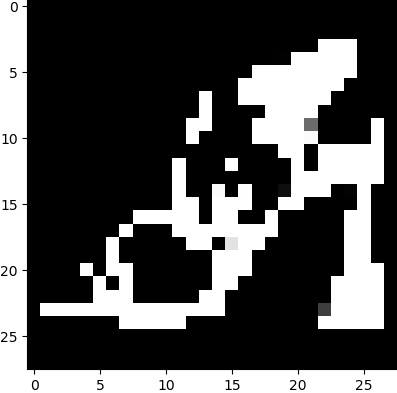}\hspace{-0.35cm} & \includegraphics[width=0.055\linewidth]{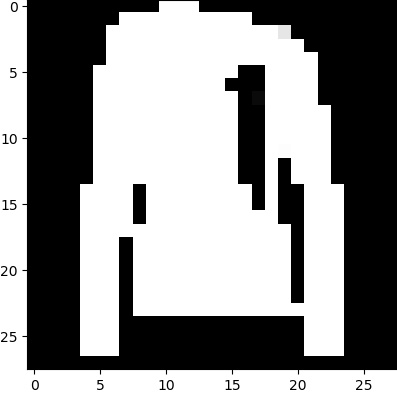}\hspace{-0.35cm} & \includegraphics[width=0.055\linewidth]{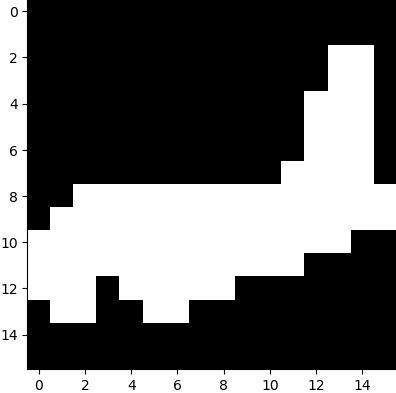}\hspace{-0.35cm} & \includegraphics[width=0.055\linewidth]{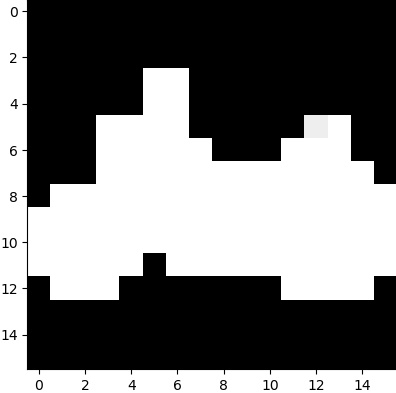} \\
 \hline 
\end{tabular}
\end{center}
\vspace{-0.5cm}
\end{table}

\subsection{Neural Generative Coding Yields Strong Downstream Pattern Classifiers}
\label{sec:results_classify}

All of the generative models we have experimented with in this paper are unsupervised in nature, meaning that by attempting to learn a density estimator of the data's underlying distribution, the representations acquired by each might prove useful for downstream applications, such as image categorization. To evaluate each how useful each model's latent representations might be when attempting to discriminate between samples, we evaluate the performance of a simple log-linear classifier, i.e., maximum entropy, that is fit to each model's topmost latent variable using the labels accompanying each dataset. 
For all models, we measure the classification error (\textbf{Err}), as a percentage, on the test set of each benchmark in Table \ref{results:diagonistics}, where the closer a model is to $0\%$, the better.
In addition, we provide the results for simple, purely discriminative baseline for context (the DSRN), which is simply a backprop-trained sparse recitfier network \cite{glorot2011deep} that is constrained to have the same number of synapses as the generative models. As we see in Table \ref{results:diagonistics}, in terms of test error, \textcolor{black}{the NGC models (GNCN-t1/Rao, GNCN-t1-$\Sigma$/Friston, GNCN-t2-L$\Sigma$, and GNCN-PDH) are competitive with the purely discriminatively-trained DSRN and outperform all of the other generative models (even outperforming the DSRN in one out of the four cases).}
%(though it does not outperform the DSRN in two out of four cases).

\begin{figure}[!t]%[ht]%[!tbp]
\begin{center}
\begin{subfigure}[t]{0.495\textwidth}
\centering
\begin{tabular}{c@{\hskip -0.8cm}c@{\hskip -0.4cm}}
\includegraphics[width=5.0cm]{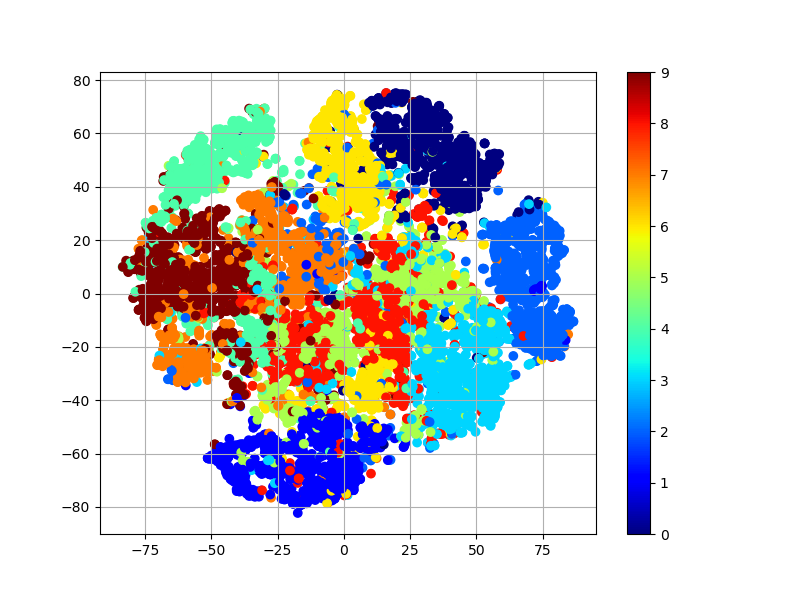} &
\includegraphics[width=5.0cm]{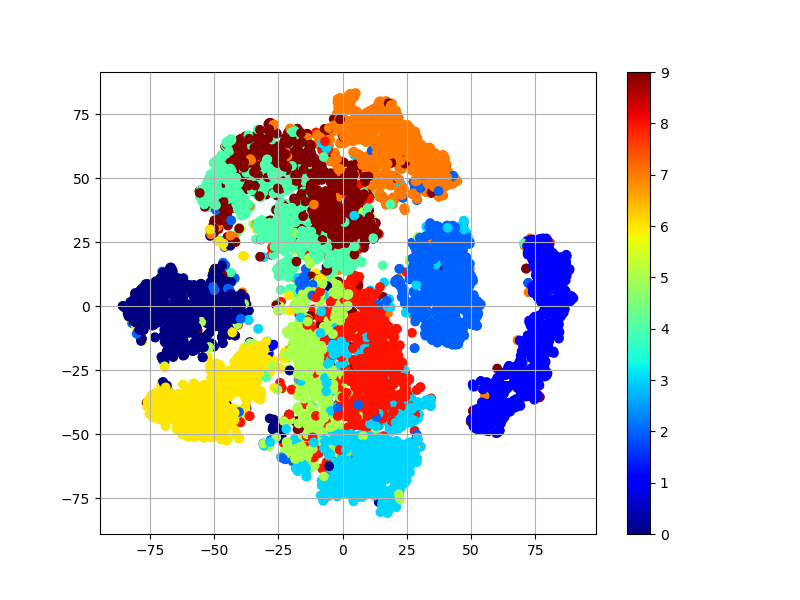}
\end{tabular}
\caption{}
\end{subfigure}
\begin{subfigure}[t]{0.495\textwidth}
\centering
\begin{tabular}{ c||c|c|c|c|c|c}  &\multicolumn{3}{c|}{\textbf{GNCN-t2-L$\Sigma$}}&\multicolumn{3}{c}{\textbf{GNCN-t1-$\Sigma$/Friston}}\\
 & $\rho_a^1$ & $\rho_a^2$ &  $\rho_a^3$ &  $\rho_a^1$ &  $\rho_a^2$ &  $\rho_a^3$ \\
 \hline
 \textbf{MNIST} & $0.21$ & $0.18$ & $0.16$ & $0.74$ & $0.60$ & $0.63$ \\
 \textbf{KMNIST} & $0.24$ & $0.21$ & $0.18$ & $0.35$ & $0.50$ & $0.67$ \\
 \textbf{FMNIST} & $0.21$ & $0.20$ & $0.18$ & $0.41$ & $0.44$ & $0.68$ \\
 \textbf{CalTech} & $0.21$ & $0.19$ & $0.17$ & $0.50$ & $0.68$ & $0.69$ \\  
 \hline
\end{tabular}
\caption{}
\end{subfigure}
\end{center}
\caption{
\textcolor{black}{
\textbf{Examination of model latent codes:}
(a) t-SNE plot of latent codes of the GAN-AE and GNCN-t2-L$\Sigma$ on the MNIST database. 
%Notice that the GNCN, even though it is unsupervised, yields representations that are more easily separable by category than those yielded by the GVAE. 
(b) Sparsity levels in the GNCN-t2-L$\Sigma$ versus the GNCN-t1-$\Sigma$/Friston \cite{friston2008hierarchical}, $\rho^\ell_a$ indicates sparsity for layer $\ell$. (Analysis performed on training set.)}
}
\label{fig:tsne_latents}
\vspace{-0.5cm}
\end{figure}

In Figure \ref{fig:tsne_latents}, we provide qualitative evidence that the latent representations of \textcolor{black}{an NGC model (a GNCN-t2-L$\Sigma$)} appear to yield a stronger, natural separation of the test data points into seemingly class-respective clusters as compared to the GAN-AE (one of the best performing backprop-based models with respect to both log likelihood and reconstruction error). Again, we emphasize that the \textcolor{black}{GNCN-t2-L$\Sigma$} acquired these representations without labeled information, meaning that the class-based relationships have emerged as a result of its very sparse neural activities.  This offers some promising evidence that \textcolor{black}{an NGC model's} representations offer benefits beyond the original density estimation task, allowing reuse of the same system for downstream tasks like categorization. 

We hypothesize that the \textcolor{black}{GNCN-t2-L$\Sigma$'s} latent codes make it easier for a linear classifier to separate out patterns by category as a result of the fact that they are sparse. Crucially, this model's lateral structure creates columns of neural processing units that fight for the right to explain the input, meaning that only a few within a group would be active when processing particular patterns. % - conduct quick analysis measuring the sparsity of each layer?
To quantity this sparsity in the \textcolor{black}{GNCN-t2-L$\Sigma$}, we measure and report in the table of Figure \ref{fig:tsne_latents} (b) the (mean) proportion $\rho^\ell_{a}$ of neurons at layer $\ell$ that were active (we counted whether each unit for a given pattern satisfied $z^\ell_i > \epsilon$, where $\epsilon = 1e^{-6}$)) \textcolor{black}{in each model on each dataset. We compare this model to the GNCN-t1-$\Sigma$/Friston \cite{friston2008hierarchical}, one of the best performing GNCNs that used a kurtotic prior to induce sparsity, and observe that the lateral inhibition/excitation greatly reduces the amount of neural firing while still obtaining top performance.}
Note that the activities of the \textcolor{black}{GNCN-t2-L$\Sigma$} are quite sparse (the highest/worst-case value was on KMNIST in layer $1$ with a sparsity of $21$\%) and, furthermore, the number of active neurons is lower in deeper layers, with $11$-$15$\% sparsity for all datasets in the top layer $\mathbf{z}^3$.
We believe that the \textcolor{black}{GNCN-t2-L$\Sigma$'s (and GNCN-PDH's)} structured form of sparsity aids it in modeling input, i.e., yielding strong likelihood and reconstruction error, facilitating better separation of nonlinear data, and \textcolor{black}{opening the door to designing models that more directly adhere to homeostatic constraints similar to those imposed by the brain}. Though models like the GAN-AE, GVAE, and CV-GVAE reach good log likelihoods too, it is possible that since their learning process focuses on shaping their latent spaces to dense, multivariate Gaussian distributions, there is little chance for useful sparsity to emerge as a by-product, reducing the possibility that the latent space might result in beneficial side-effects.

From a neurobiological perspective, it is well-known that lateral synapses often facilitate contextual processing \cite{cui2017sparseHTM} and offer a natural form of ``activity sharpening'' \cite{adesnik2010lateral}. %lamsa2007anti
We believe that this is an important structural prior that biases the \textcolor{black}{an NGC model like the GNCN-t2-L$\Sigma$ or GNCN-PDH} towards acquiring more economical representations \cite{barlow1972single}, where progressively fewer neurons at layers progressively farther away from the input stimulus work to encode information (or are non-zero).
%more ``causal-like'' factor representations when attempting to explain the data with a top-down variable structure. 
%Furthermore, our form of structural sparsity allows us to interpret neural columns as more complex processing units (or, loosely, like latent variables) that will activate the same way when learning occurs and when it does not. % <-- good for appendix/methods
Much like the more recent incarnations of spike-and-slab coding \cite{goodfellow2013spikeslab}, our form of structurally-enforced sparsity has a much stronger regularization effect on the model and ensures that the generative distribution of the neural activities is truly sparse. This is unlike more traditional approaches where sparsity is weakly enforced through the use of factorial kurtotic priors applied during inference  \cite{olshausen1996emergence,rao1999predictive}.
Our NGC frameowrk not only yields naturally sparse codes but also offers flexibility in exploring other types of lateral connectivity patterns beyond the choice made in this study.% (see Methods for more details).

\subsection{Neural Generative Coding Can Conduct Pattern Completion}
\label{sec:results_completion}
Another interesting ability attributed to auto-associative memory models is their ability to complete partially-corrupted or incomplete patterns. In the real world, this type of scenario would often occur in the form of object occlusion, where the view of an object might be partially obstructed from the agent's view. Being able to ``imagine'' the rest of the object might prove useful when planning to grasp it or manipulate it in some fashion.
To test each model's ability to complete patterns, we conducted an experiment where the right half of each image in each dataset was masked and each model was tasked with predicting the deleted portions. In Table \ref{results:diagonistics}, we report the masked mean squared error (M-MSE, see Supplementary Methods) of each model on each dataset's test set. %, which is computed per image and summed over all images in the test sample.

Interestingly enough, we see that the \textcolor{black}{NGC models outperform} the other baselines in terms of pattern completion (with \textcolor{black}{GNCN-t1-$\Sigma$/Friston and GNCN-PDH offering the most} competitive performance, with respect to measured M-MSE. We furthermore provide some examples of original data patterns that were masked in Table \ref{results:diagonistics} (in the first and third rows) contrasted with the GNCN-PDH's completed patterns  (in the second and fourth rows). 
We hypothesize that the reason \textcolor{black}{an NGC model} is able to conduct pattern completion better than the autoencoder models is due to its iterative inference process. 
%This is if further supported by the observation that the two baseline predictive processing models, PC-Rao and PC-Friston, also outperform the autoencoders (though the GNCN models perform the best overall).
%Much like the analysis on downstream classification, we hypothesize that the GNCN is able to conduct better pattern completion because it is not as constrained to learn an actual distribution over its top-most hidden layer and thus able to extract sparse latent codes given its lateral synapses.

\section{Discussion}
\label{sec:discussion}

% general point of view -- DNNs have worked well across many applications but are backprop-based...backprop models don't do a lot of what brain does, could building models that break free of backprop and emulate some of these yield better models?
Generative models based on artificial neural networks (ANNs) have yielded promising tools for estimating and sampling from complicated probability data distributions \cite{kingma2013auto,goodfellow2014generative}. By re-considering how these models might operate and learn, drawing inspiration from one promising neuro-mechanistic account of how the brain interacts and adapts to its environment, i.e., predictive processing \cite{clark2015surfing}, %\cite{clark2015surfing,spratling2016predictive,keller2018predictive}, 
we have shown that learning a viable generative model is possible. Specifically, we propose the neural generative coding (NGC) computational framework for learning neural probabilistic models of data, implementing \textcolor{black}{several concrete instantiations of NGC which we called generative neural coding networks (GNCNs)}. 
In our experiments, we observe that the \textcolor{black}{GNCNs are} not only competitive with several powerful, modern-day backprop-based models on the task of estimating the marginal distribution of the data but \textcolor{black}{they} can generalize beyond the task \textcolor{black}{they were originally} trained to do. Specifically, we investigated the performance of \textcolor{black}{all of the models} on downstream tasks such as pattern completion and pattern categorization and discovered that the unsupervised \textcolor{black}{NGC models outperformed} all of the examined backprop-based baselines and \textcolor{black}{were} even competitive with an ANN that was directly trained to specialize for classification.
As a result, for systems as complex as probabilistic generative models, we have demonstrated that crafting a more fundamentally brain-inspired approach to information processing and credit assignment can yield artificial neural systems that extract rich representations of input data in an unsupervised fashion.
Our results demonstrate, on four datasets, that even though extra computation is needed to process each input for a fixed (yet small) stimulus presentation time, the \textcolor{black}{NGC models converge} far sooner than comparable backprop-based ones, generalizing well earlier.

\begin{table}[!t]%[ht]
\centering
\begin{tabular}{c|m{1.25cm}|m{1.25cm}|m{11cm}}
Class & \multicolumn{1}{c}{Data} & \multicolumn{1}{c}{Output} & \multicolumn{1}{c}{\textcolor{black}{Top} Selected Features} \\
\hline
\hline
``0'' & \includegraphics[width=1.25cm]{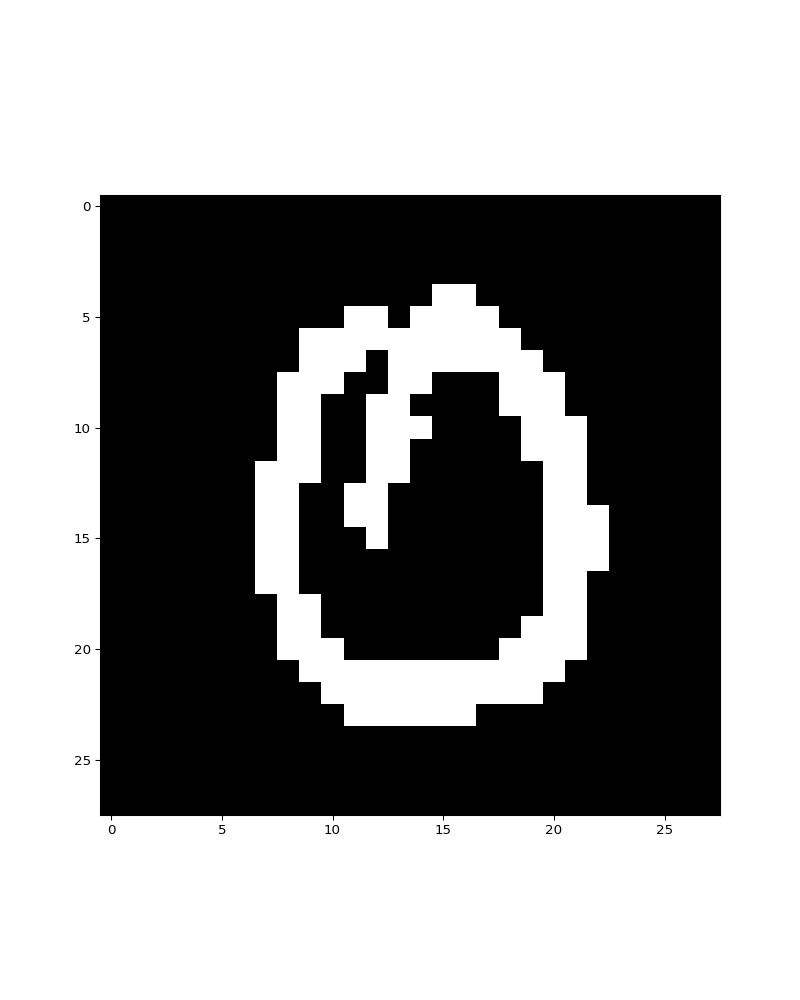} & \includegraphics[width=1.25cm]{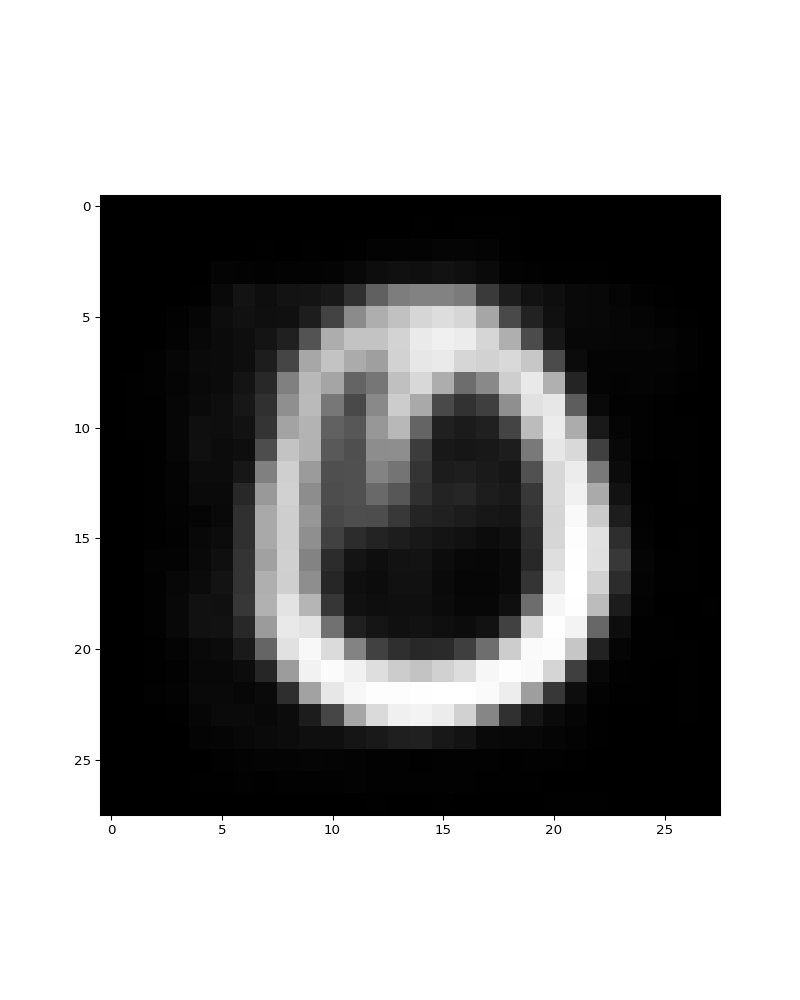} & \includegraphics[width=11cm]{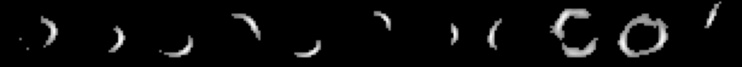}\\
``1'' & \includegraphics[width=1.25cm]{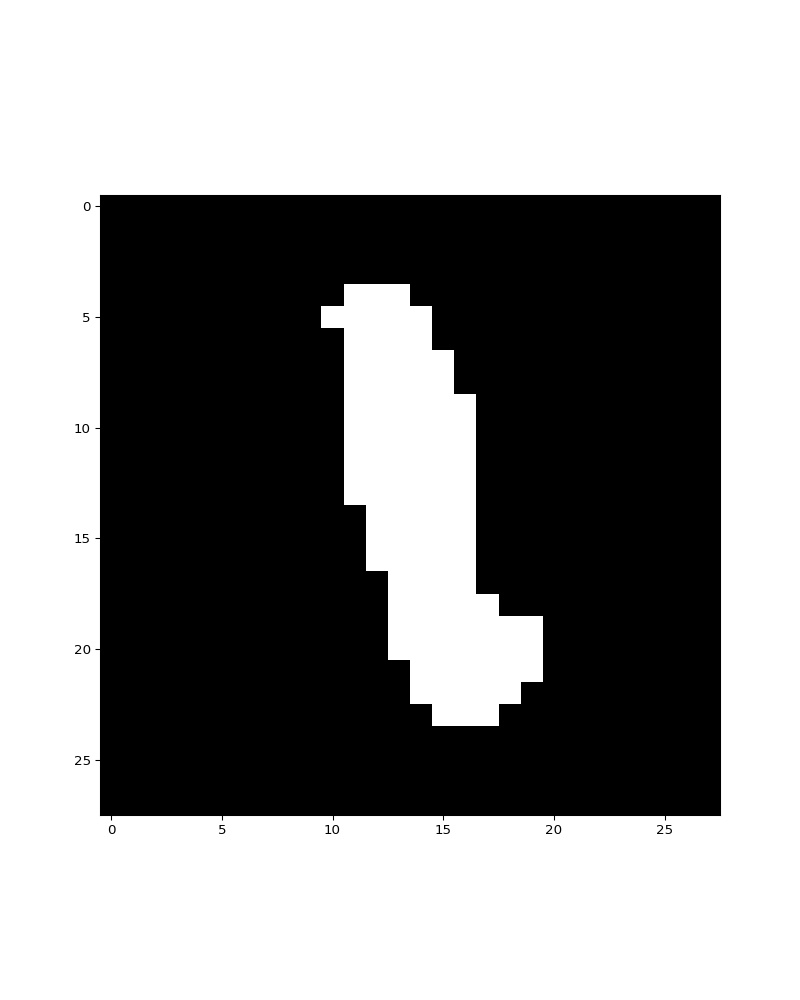} & \includegraphics[width=1.25cm]{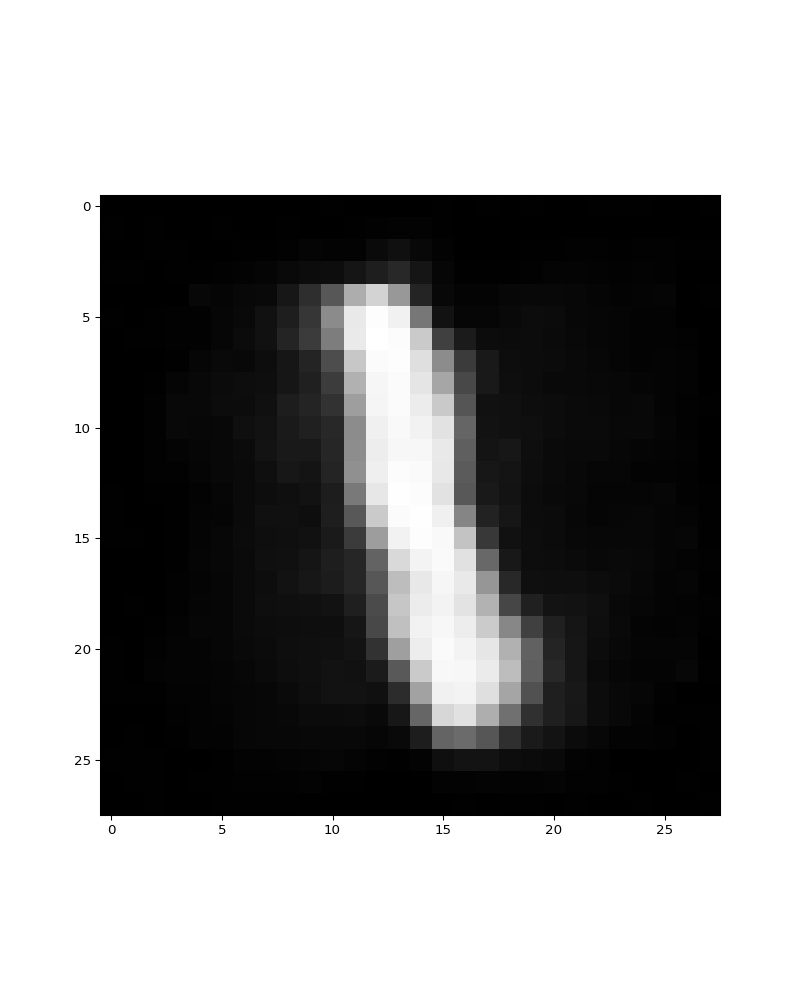} & \includegraphics[width=11cm]{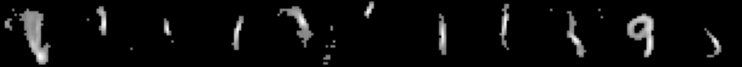}\\
``2'' & \includegraphics[width=1.25cm]{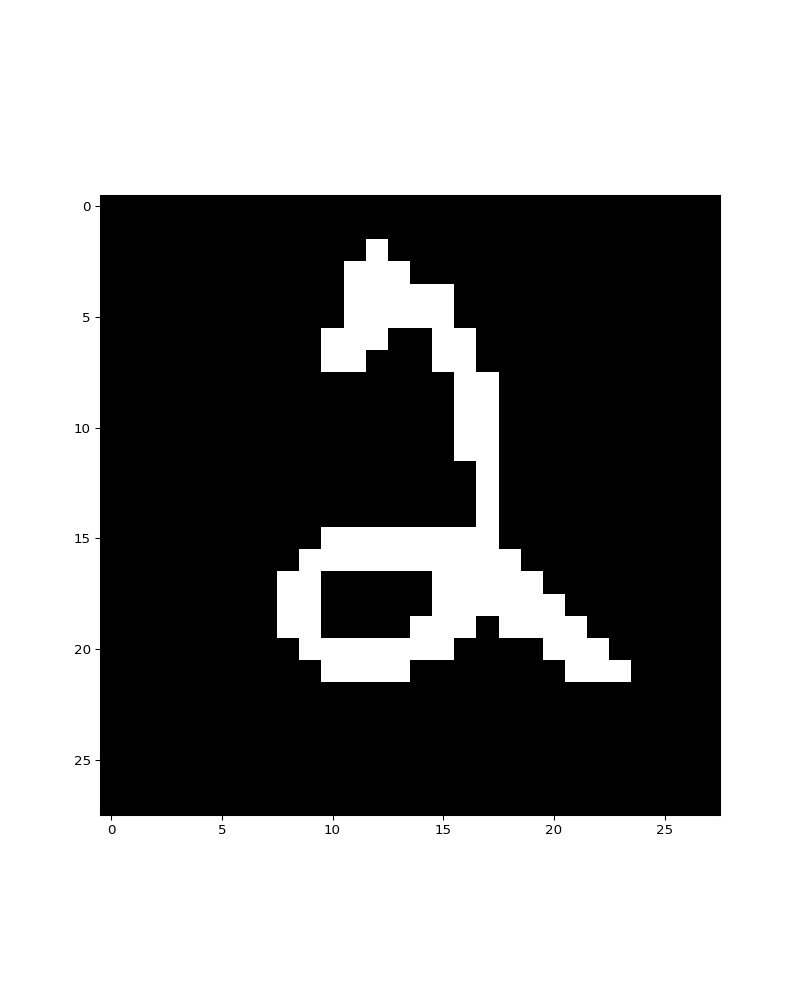} & \includegraphics[width=1.25cm]{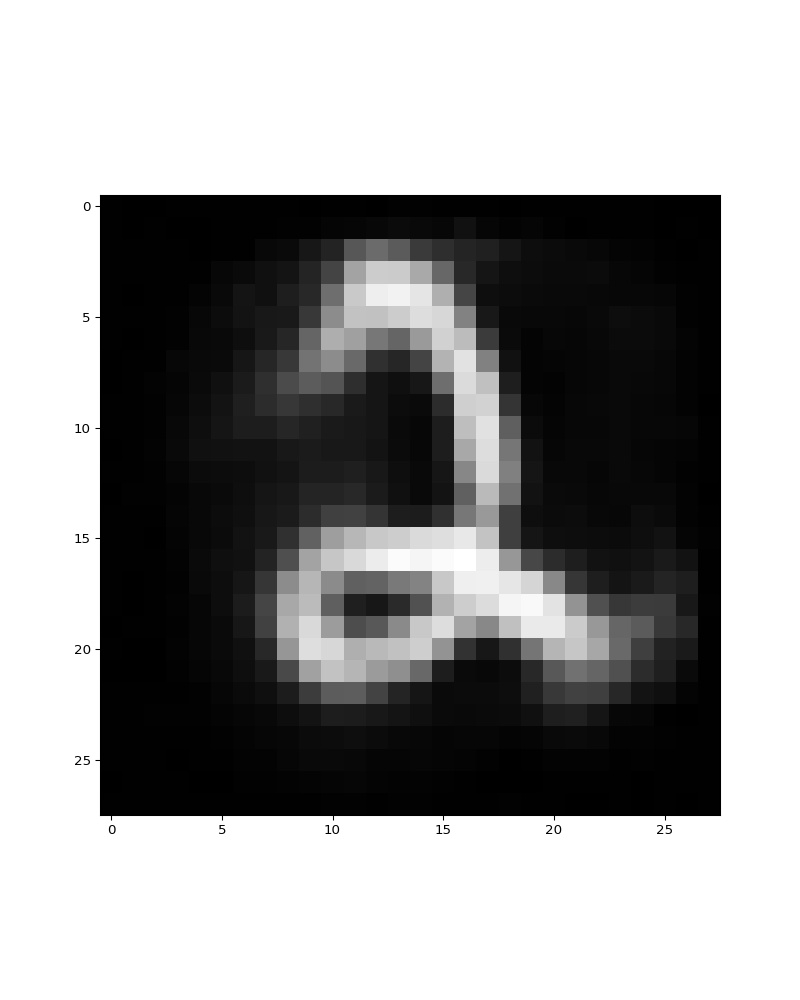} & \includegraphics[width=11cm]{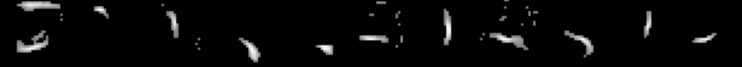}\\
``3'' & \includegraphics[width=1.25cm]{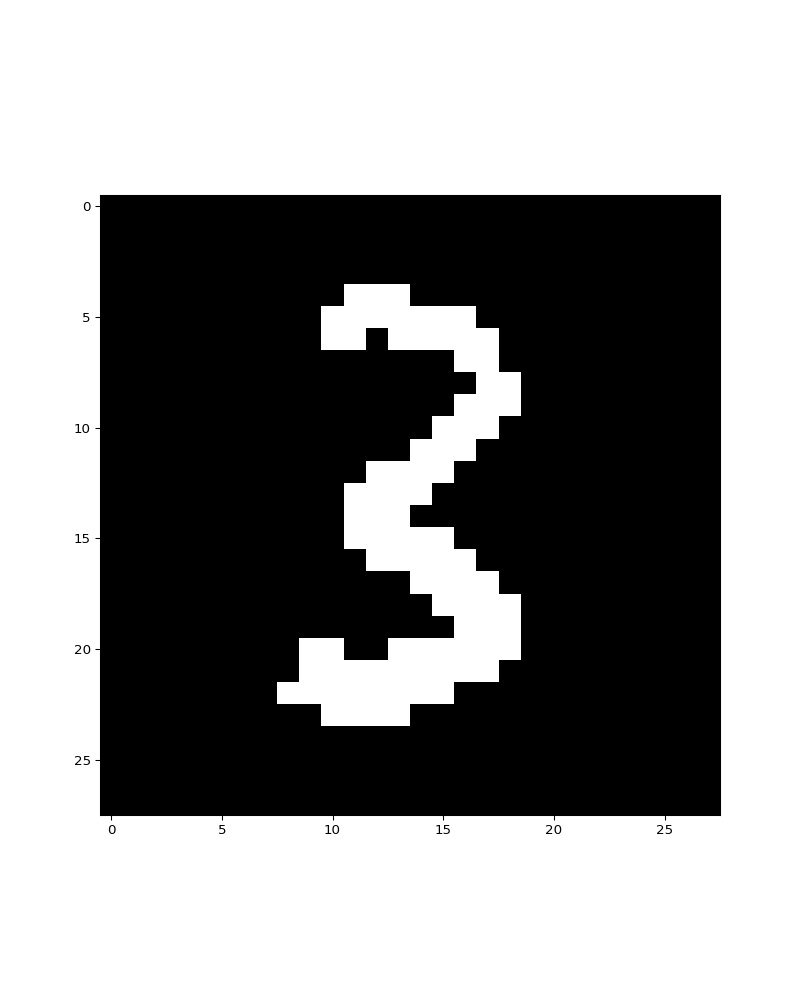} & \includegraphics[width=1.25cm]{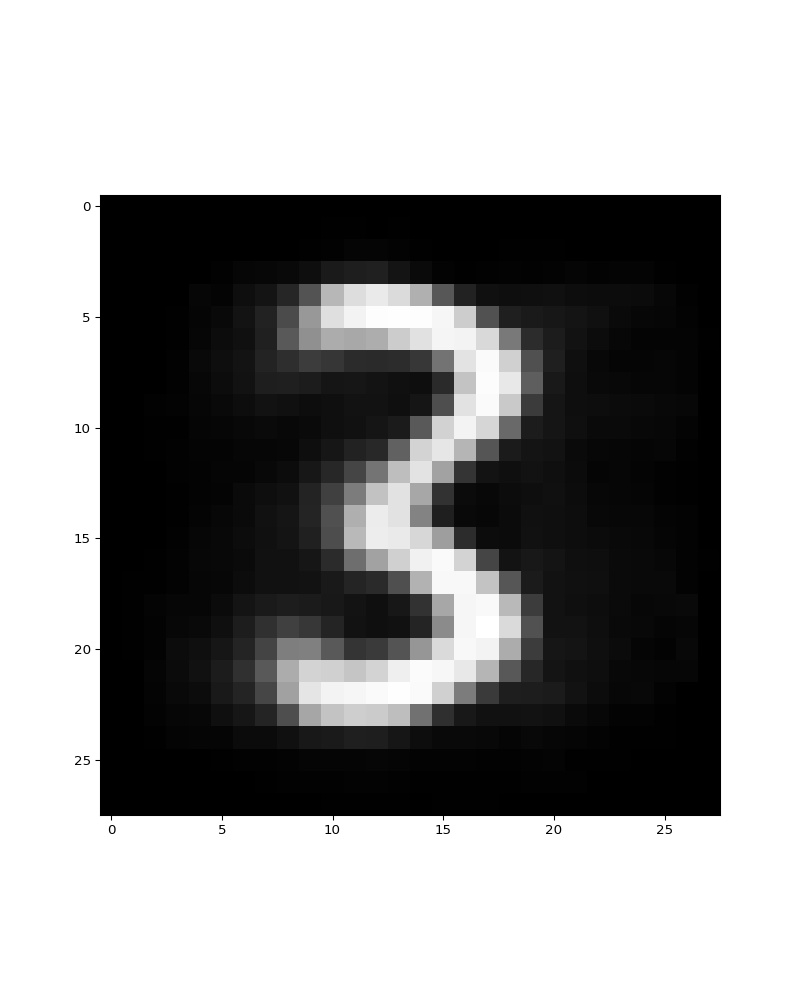} & \includegraphics[width=11cm]{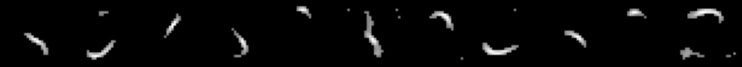}\\
``4'' & \includegraphics[width=1.25cm]{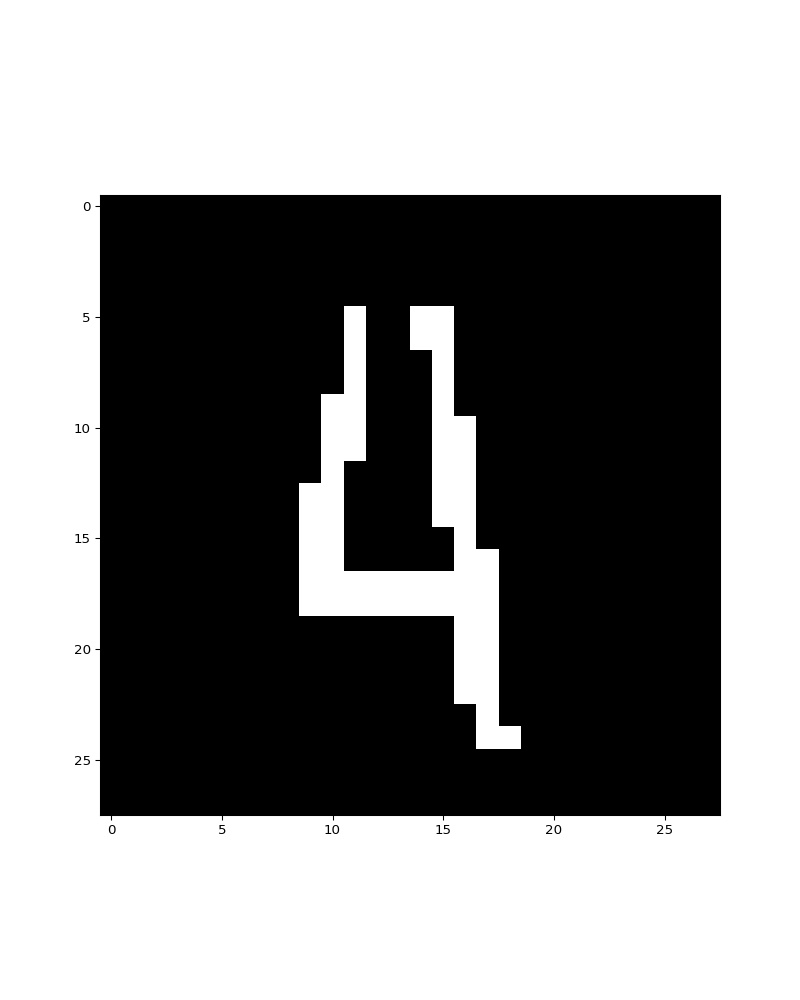} & \includegraphics[width=1.25cm]{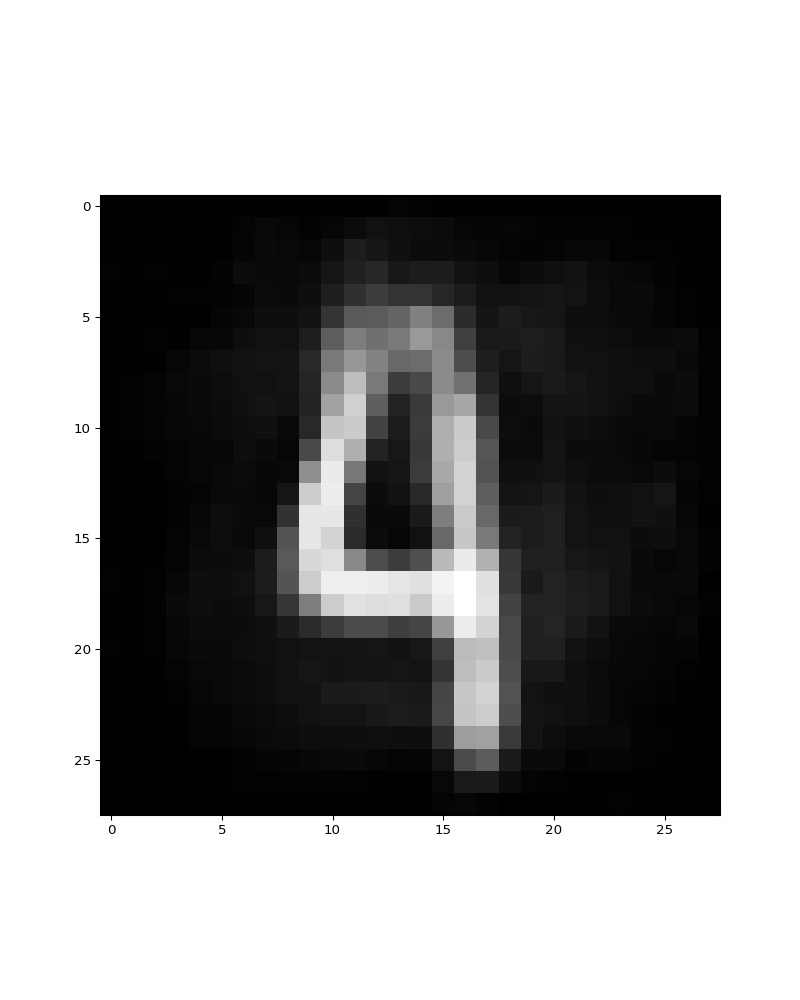} & \includegraphics[width=11cm]{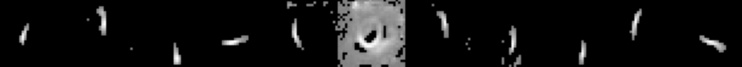}\\
``5'' & \includegraphics[width=1.25cm]{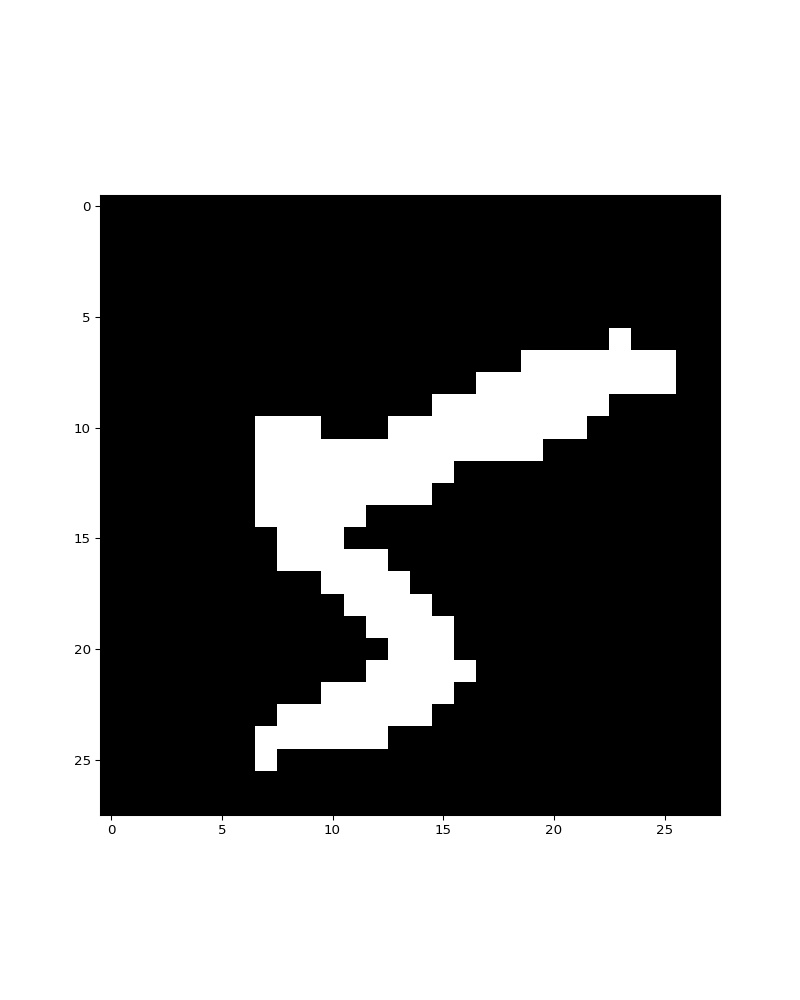} & \includegraphics[width=1.25cm]{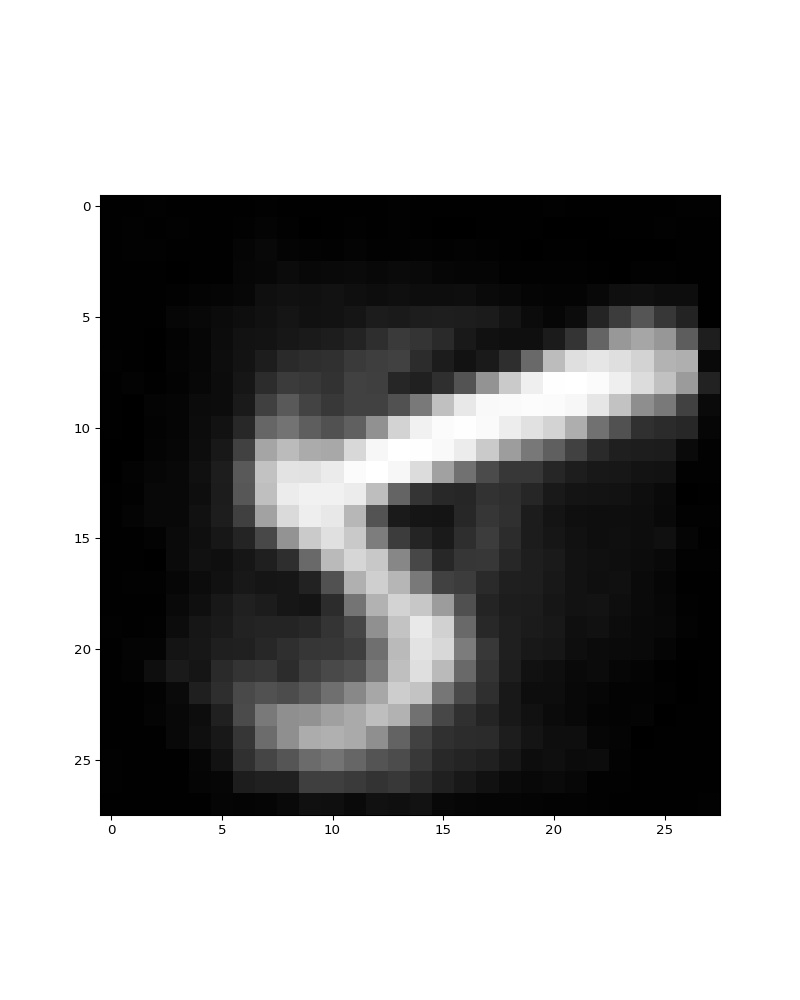} & \includegraphics[width=11cm]{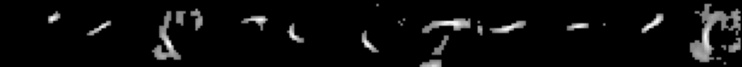}\\
``6'' & \includegraphics[width=1.25cm]{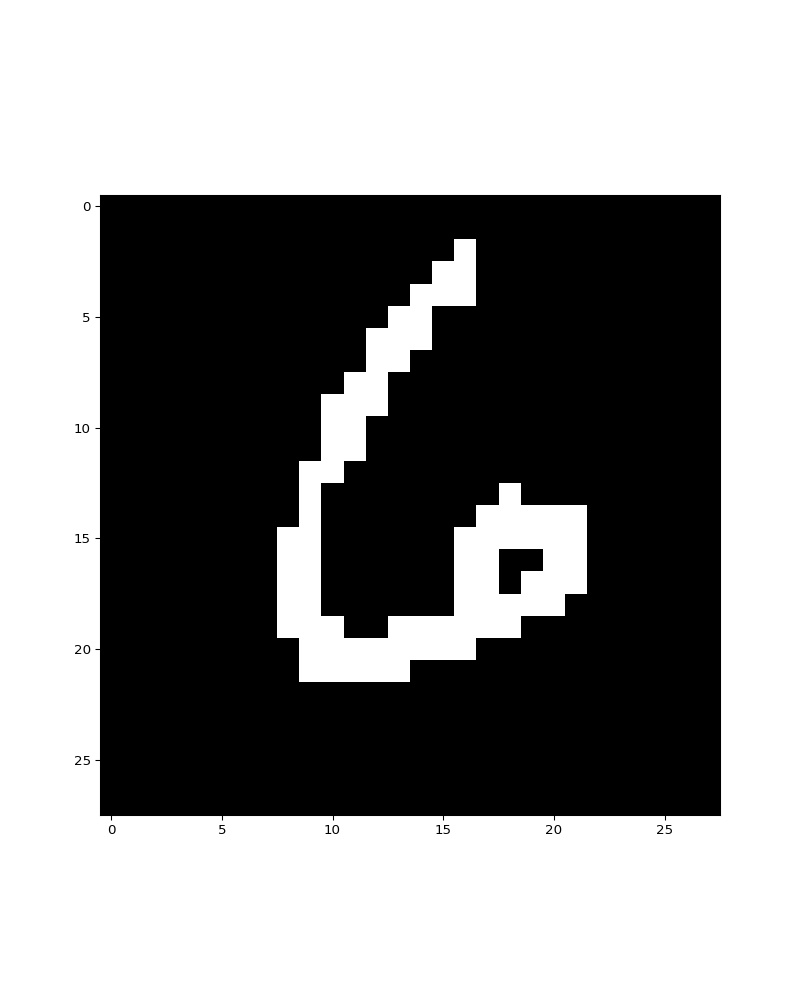} & \includegraphics[width=1.25cm]{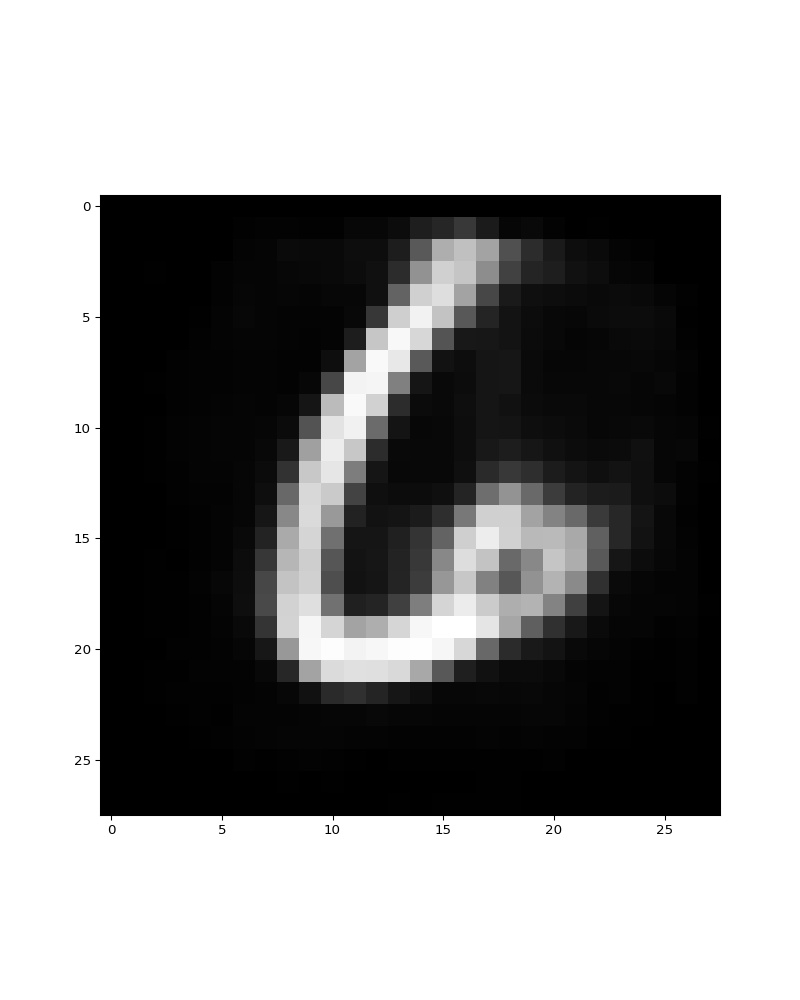} & \includegraphics[width=11cm]{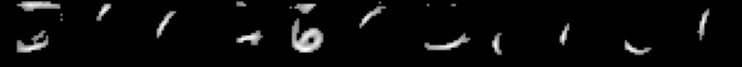}\\
``7'' & \includegraphics[width=1.25cm]{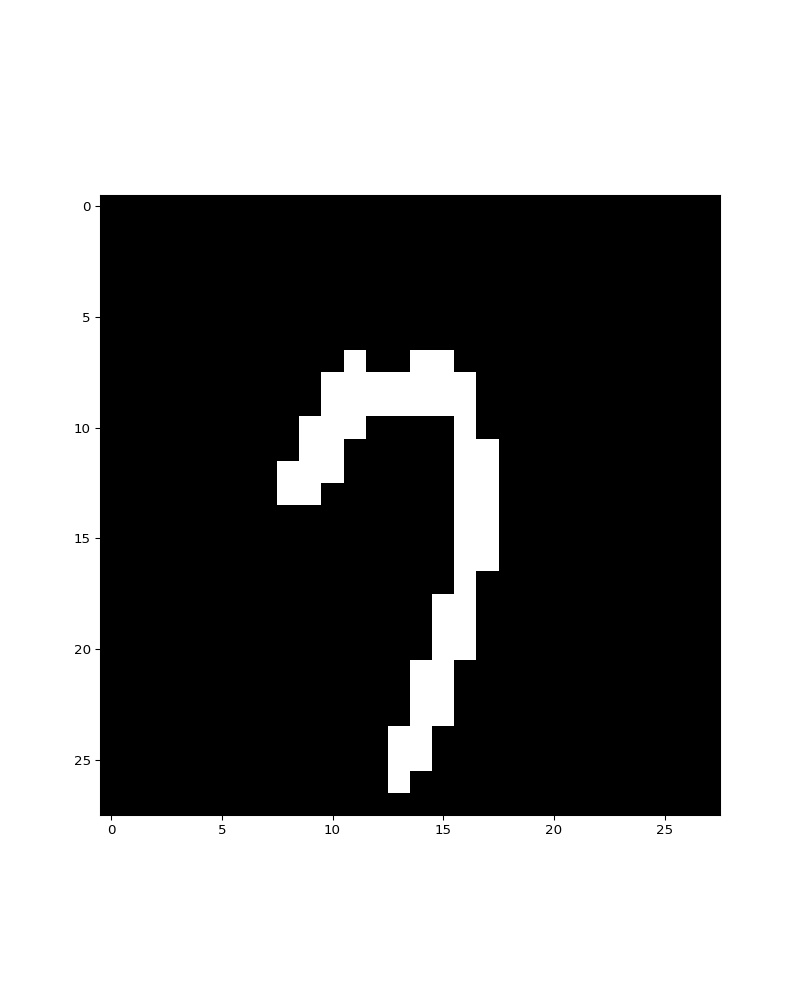} & \includegraphics[width=1.25cm]{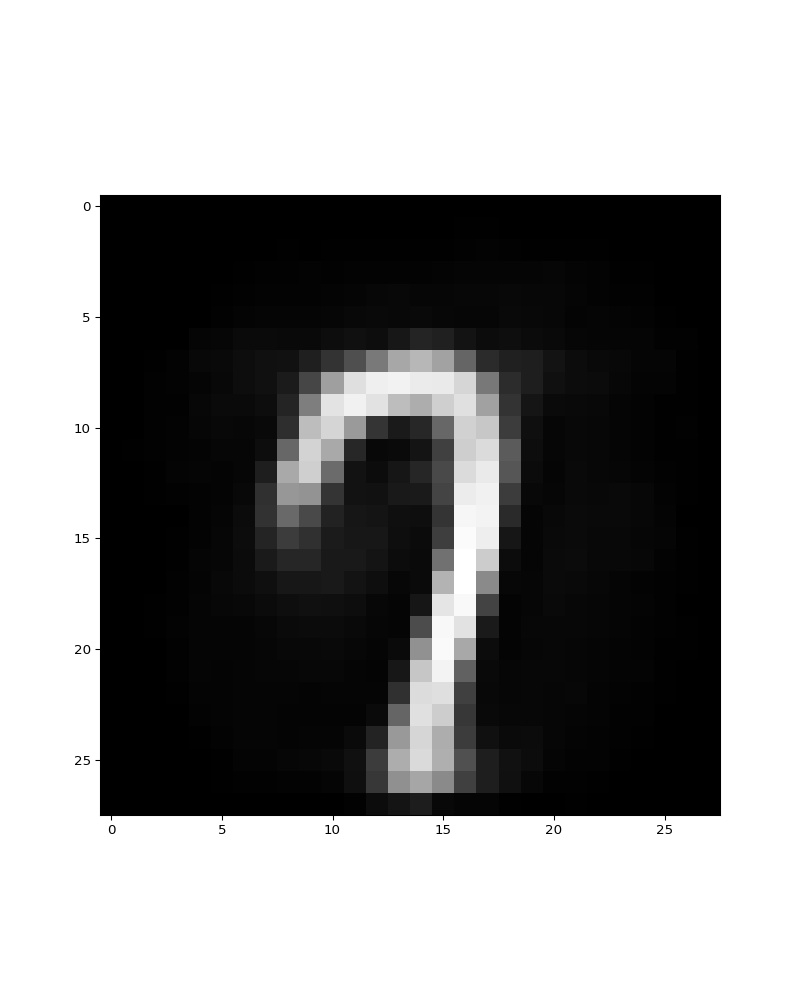} & \includegraphics[width=11cm]{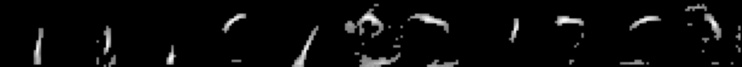}\\
``8'' & \includegraphics[width=1.25cm]{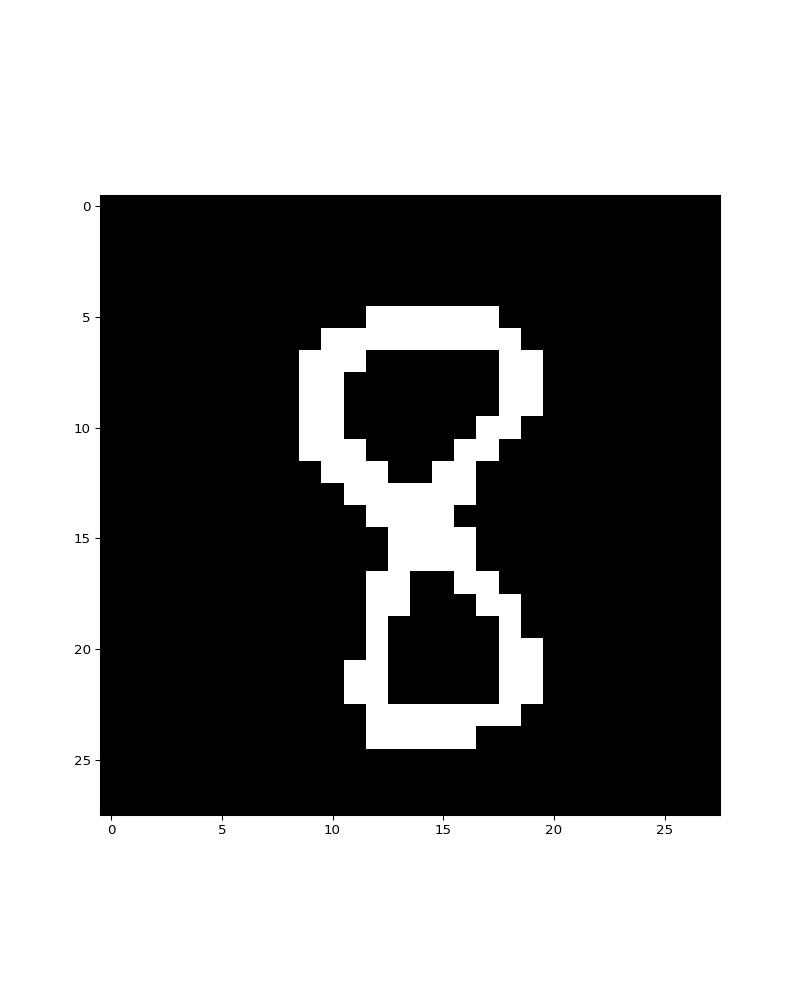} & \includegraphics[width=1.25cm]{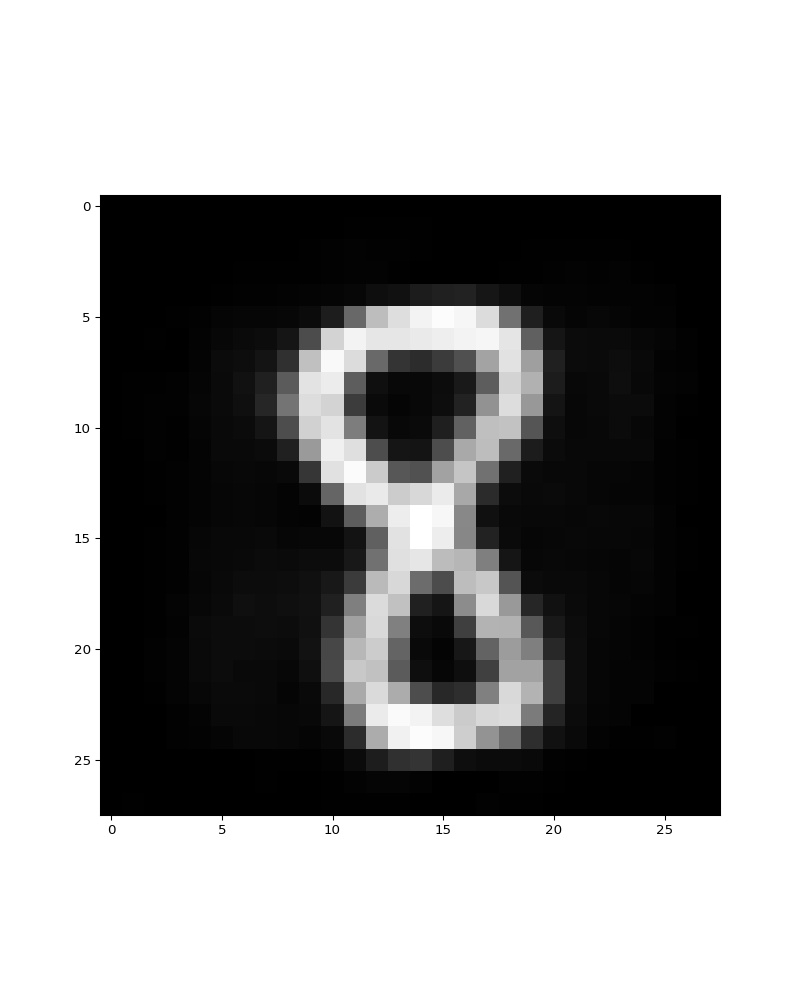} & \includegraphics[width=11cm]{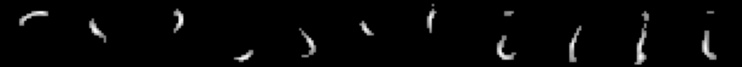}\\
``9'' & \includegraphics[width=1.25cm]{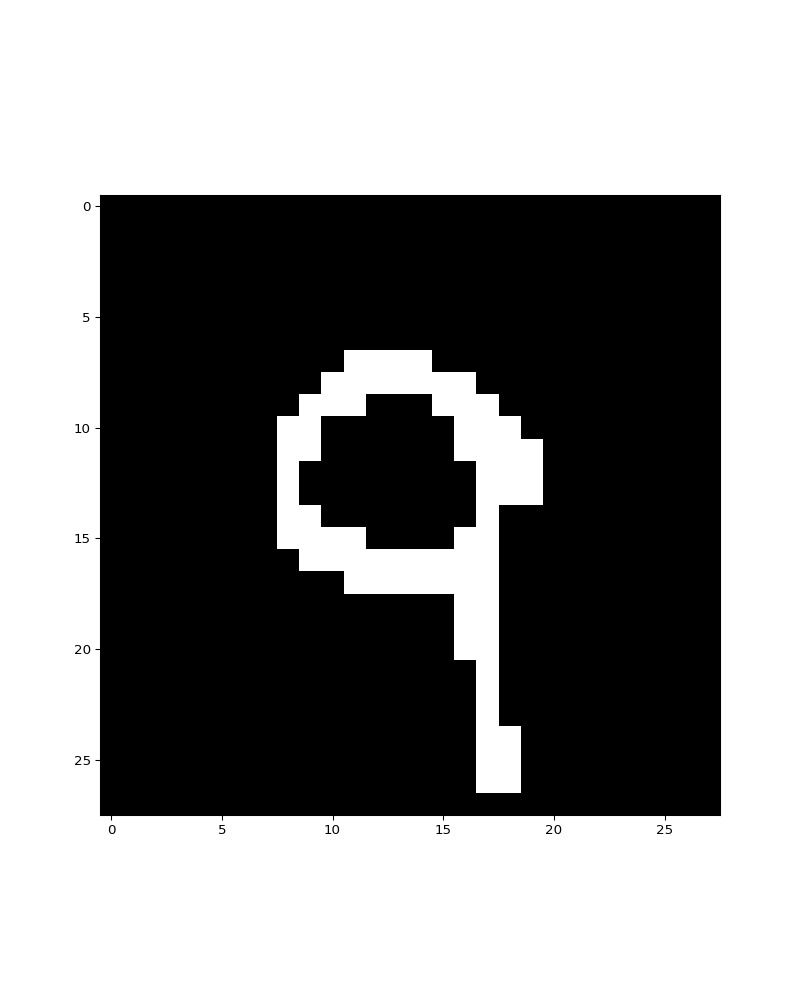} & \includegraphics[width=1.25cm]{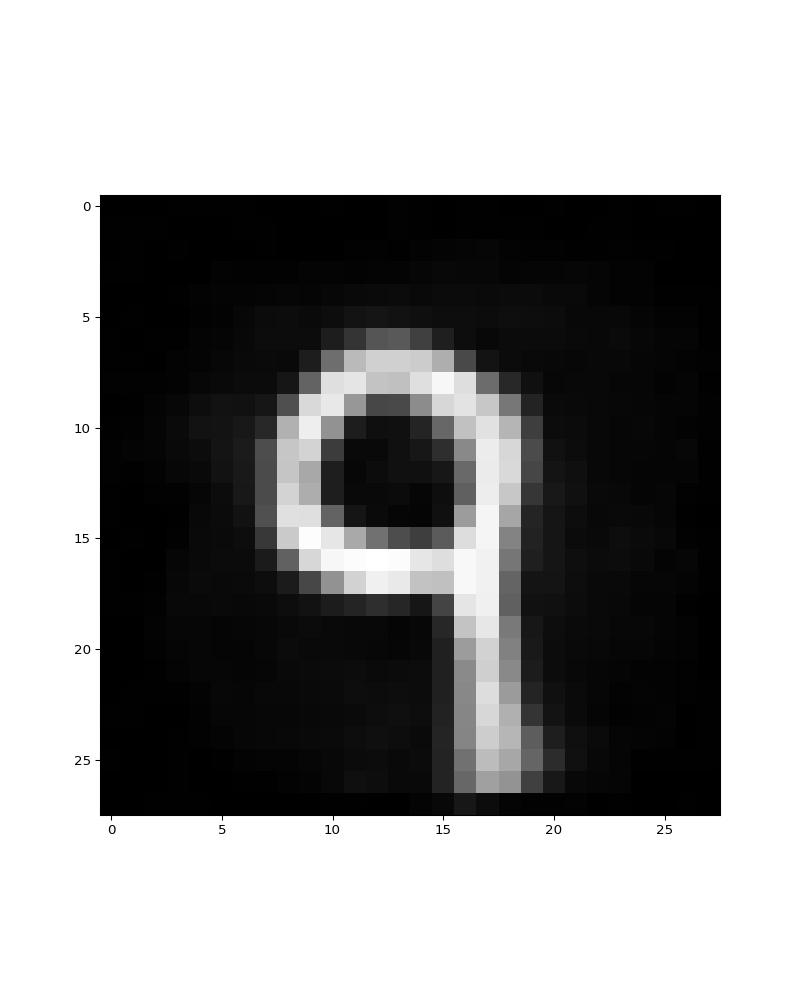} & \includegraphics[width=11cm]{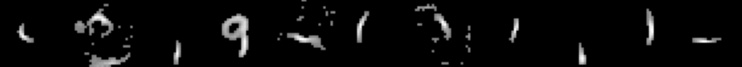}\\
\end{tabular}
\caption{
\textbf{Feature Composition Analysis of the generative neural coding network:} 
The ``Class'' shows the class label of a sample from MNIST, the ''Data'' column depicts the original pattern, and the ``Output'' column presents the weighted super-position of the $30$ layer $1$ synaptic weight vectors (we present the top $11$ in the ``\textcolor{black}{Top} Selected Features'' column) associated with the $30$ state neurons with highest activity when reconstructing the original datapoint. Note that the feature maps above were extracted from \textcolor{black}{a GNCN-t2-L$\Sigma$} trained on MNIST and were first normalized and values lower than a threshold of $0.5$ were set to black in order to improve the clearness of the specifically extracted feature.}
\label{results:feature_analysis}
\vspace{-0.5cm}
\end{table}

\begin{figure}[!t]%[ht]
\centering
\includegraphics[width=0.75\textwidth]{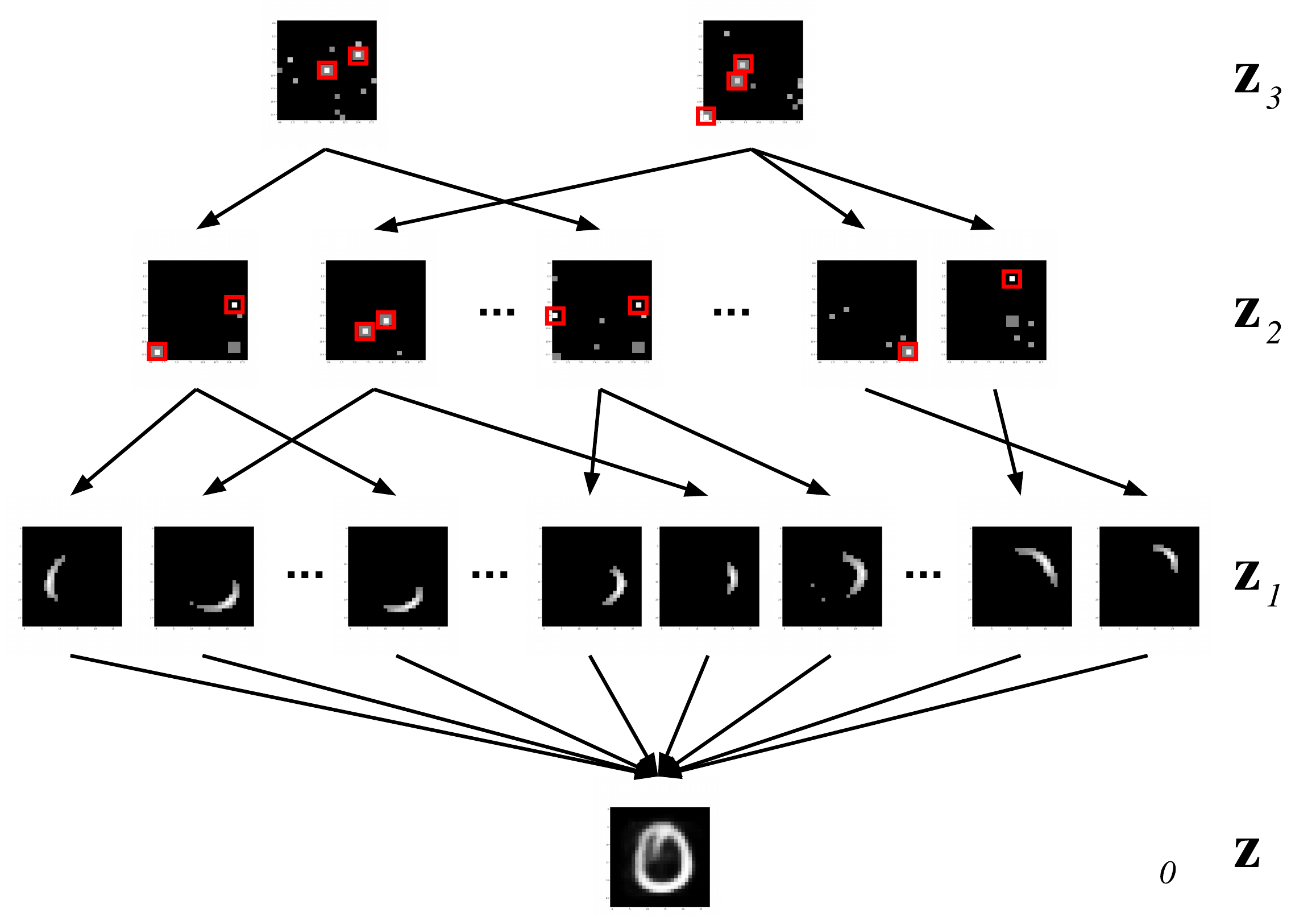}
\caption{
\textbf{The feature composition process:}
Illustration of an \textcolor{black}{NGC network} composing the digit zero that was randomly sampled from the MNIST database (all the feature maps shown in this image have been extracted from one of the \textcolor{black}{GNCN-t2-L$\Sigma$ models} trained on MNIST). Note that layers $2$ and $3$ ($\mathbf{z}_2$ and $\mathbf{z}_3$) control a hierarchical neuron selection pathway which leads to excitation (turning ``on'') or suppression (turning ``off'') of certain low-level neurons in layer $1$ ($\mathbf{z}_1$) that contain visual features that are super-imposed to generate a final output image. The image maps shown corresponding to $\mathbf{z}_2$ and $\mathbf{z}_3$ are visualized synaptic weight vectors that appear to serve as blueprints for choosing units (white indicate neurons to excite and black indicates neurons to suppress, after normalization was applied to the original maps and values lower than a threshold of $0.5$ were set to black). Red squares have been added to indicate the neurons with some of the highest activity values that select the most important visual features that will be used to compose a final output predicted image. }
\label{fig:gncn_feature_compose}
\vspace{-0.55cm}
\end{figure}

%%%%%%%%%%%%%%%%%%%%%%%%%%%%%%%%%%%%%%%%%%%%%%%%%%%%%%%%%%%%
To offer additional insight into what \textcolor{black}{an NGC model's} intermediate representations might look like, we examined, using the MNIST dataset, the generative synaptic weights for each layer of a trained  model, of which the visualization results for the bottom layer (where $\mathbf{z}^1$ predicts $\mathbf{z}^0$) are shown in Table \ref{results:feature_analysis} (see Supplementary Note 8 for details of the analysis).
Upon examination, we observe that \textcolor{black}{a GNCN-t2-L$\Sigma$} appears to learn a latent command structure in its upper layers (layers $\mathfrak{N}^2$ and $\mathfrak{N}^3$ -- these appear to provide maps for turning off or on state neurons in the level below) that work in tandem to drive a dynamic composition of low-level visual features in layer $\mathfrak{N}^1$. In Figure \ref{fig:gncn_feature_compose}, we illustrate how these higher-level maps interact with the layer $\mathfrak{N}^1$ features, ultimately composing a final output image by super-imposing an intensity-weighted set of low-level features (see the ``Output'' column of Table \ref{results:feature_analysis}).

%%%%%%%%%%%%%%%%%%%%%%%%%%%%%%%%%%%%%%%%%%%%%%%%%%%%%%%%%%%%

% TODO: talk about implications for further integration with neuroscience (lateral connections/sparsity, precision-weighting, etc.) --> also mention that we are working in the direction of (computational) neuroscience to statistical learning (which has many useful implications) but going the other way around would be an important, but challenging different step in of itself
% one main benefit of NGC is it offers a very deep-learning friendly way of scaling one neuro-mechanistic/neuro-bio account of cortical function to machine learning level tasks...talk about how it is one key way of bringing free energy principle to machine learning more directly...might mention that it is possible that GNCN is not as task-specific/specialized as other architectures and could prove useful for general purpose learning?
How the brain conducts credit assignment is a fundamental and open question in both (computational) neuroscience and cognitive science. There are many theories that posit how this might happen and our proposed NGC framework represents a scalable, computational instantiation of only one of them, the theory of predictive processing, suggesting that cortical regions in the brain communicate prediction errors and/or predictions across different regions through a hierarchical message passing scheme \cite{wacongne2011evidence}. Observe that the direction that this study takes is one that starts from cognitive neuroscientific concepts and ends in the development of a statistical learning algorithm. Specifically, we have shown that one way of emulating some neuro-biological principles yield agents that learn more general-purpose representations, as our results on downstream classification and pattern completion provide evidence for. As for implications for computational neuroscience itself, while \textcolor{black}{an NGC model can} embody concepts such as lateral competition-driven sparsity, hierarchical message passing, and local Hebbian-like synaptic adjustment, our framework lacks many important details that might allow it to serve as a means to make falsifiable claims that can be proven or disproven by neurobiological experiments.
In addition, \textcolor{black}{an NGC model} suffers from other criticisms of many predictive processing models -- for example, it requires a one-to-one pairing of error neurons with state neurons \cite{bogacz2017tutorial} (though this can potentially be resolved by decoupling the error neurons and introducing an extra synaptic weight matrix that connects a pool of error neurons of a one size to a pool of state neurons of a different size).
However, if the NGC framework is modified significantly to more faithfully model neurobiological details, e.g., synapses are constrained to take on only non-negative values and neurons communicate via spike trains (for example, the NGC could work with  leaky integrate-and-fire neurons, similar to that of \cite{ororbia2019spiking}), it could serve as the means to facilitate refinement of predictive brain theories such as predictive processing \cite{clark2015surfing} and principles such as free energy \cite{friston2006free}. Doing so could facilitate  stronger synergy between neural computational modeling and the design of agents that solve complex tasks examined in statistical learning. 

Given the difficulty in imagining how backprop, in the form it is implemented when training deep ANNs today, might occur in the brain \cite{grossberg1987competitive,crick1989recent}, there is value in not only developing approximations of it that might be more brain-like (moving machine learning a bit closer to computational neuroscience) but also in exploring alternatives that start from neuro-cognitive principles, theories, and mechanisms, creating new algorithms that embody particular ideas in cognitive neuroscience at the outset. Taking the second of the last two directions, which is in the spirit of this work and several others \cite{guerguiev2017towards,sacramento2018dendritic}, might allow us to more easily shed the constraints imposed by backprop in the effort to construct general-purpose learning agents capable of emulating more complex human cognitive function. Doing so might also allow the machine learning community to make further progress on problems even harder than generative modeling, such the problem of active inference \cite{friston2011action} and continual temporal prediction \cite{ororbia2020continual}.% where an agent must interact with dynamic environments.

% \subsection{Programming}
% The experiments, the baseline models, and the proposed computational framework were written in Python, using the Numpy and TensorFlow 2 libraries. Experiments were sped up using a GPU card.

% \subsection{Data Availability}
% The datasets, MNIST \cite{lecun1998mnist}, Fashion MNIST  \cite{xiao2017fashion}, NotMNIST \cite{not2018mnist}, and Caltech 101 Silhouettes \cite{caltech2018silh}, used in this study are publicly available.

% \subsection{Code Availability}
% All code used for training and analysis of the NGC models is available at the link: \texttt{https://github.com/ago109/ngc-learn}.

% \subsection{Competing Interests}
% The authors declare no competing interests.

% \subsection{Contributions}
% A.O. provided the initial idea for the project, developed the Python code used in this project, performed all subsequent simulations, carried out the mathematical analysis and development of the NGC models and framework, and wrote the paper.
% D.K. provided revisions/refinement on the mathematical derivations and equations and provided copious comments and guidance in developing several of the explanations/intuitions.
% All authors discussed the results, refined the ideas in the paper, and co-edited the manuscript.

\FloatBarrier

\bibliographystyle{acm}
\bibliography{ref}

\newpage
\section*{Appendix}

\subsection*{Datasets}
\label{sec:datasets}
All of the datasets used in this paper, except for CalTech 101, which already contained binary images, contained gray-scale pixel feature values in the range of $[0,255]$. The images in these databases were first pre-processed by normalizing the pixels to the range of $[0,1]$ by dividing them by $255$ and finally converted to binary values by thresholding at $0.5$ as in \cite{bengio2013generalized}.

The MNIST dataset %\footnote{Available at the URL:  http://yann.lecun.com/exdb/mnist/.}
contains $28\times28$ images containing handwritten digits across $10$ categories.
Fashion MNIST (FMNIST) \cite{xiao2017fashion}, which was proposed as a challenging drop-in replacement for MNIST, contains $28x28$ grey-scale images depicting clothing items (out of $10$ item classes).
Kuzushiji-MNIST (KMNIST) is another $28\times28$ image dataset containing  hand-drawn Japanese Kanji characters \cite{clanuwat2018deep}.
The Caltech 101 Silhouettes database
%\footnote{https://people.cs.umass.edu/$\sim$marlin/data.shtml} 
contains $16 \times 16$ binary pixel images across $100$ different categories.
Each training subset had $60000$ samples and the testing subset had $10000$ (the standard test split of each dataset was used in this paper), with the exception of Caltech 101, which had $8596$ training samples and $2302$ test samples. A validation subset of $2000$ samples was drawn from each training set to be used for tuning model meta-parameters (Caltech had $2257$ samples).

% \subsection*{The Prediction Equation for the GNCN-PDH}
% In the GNCN-PDH, the mean vector $\mean{z}^\ell$ for any layer $\mathfrak{N}^\ell$ is obtained in a feed-forward manner from the latent state of the neighboring two layers (biases/offset terms have been omitted for clarity):
% \begin{align}
%     \mean{z}^\ell \gets g^\ell(\mathbf{W}^{\ell+1} \cdot  \phi^{\ell+1}(\latent{z}^{\ell+1}) + \mathbf{M}^{\ell+2} \cdot  \phi^{\ell+2}(\latent{z}^{\ell+2})) ,\label{eqn:ff}
% \end{align}
% where we observed that an additional, learnable auxiliary generative matrix $\mathbf{M}^{\ell+2}$ conveys and injects state value information from the layer $\mathbf{N}^{\ell+2}$ into the prediction of layer $\mathbf{N}^{\ell}$ through a simple linear combination before application of the prediction nonlinearity $g^\ell$. Another interesting property of the GNCN-PDH is that it does not contain a complementary error matrix for $\mathbf{M}^{\ell+2}$, meaning that, by design, its forward generative pathway is different from its error message transmission pathway (yielding an asymmetric network model when one considers the error correction synapses). 

\subsection*{Baseline Model Descriptions}
\label{sec:baselines}
The baseline backprop-based models implemented for this article included a regularized auto-associative (autoencoding) network (RAE), a Gaussian variational autoencoder with fixed (spherical) variance (GVAE-CV) \cite{ghosh2020variational}, a Gaussian variational autoencoder (GVAE) \cite{kingma2013auto}, and an adversarial autoencoder (GAN-AE) \cite{makhzani2016adversarial}. All models were constrained to have four layers like the \textcolor{black}{NGC models}. For all backprop-based models, the sizes of the layers in between the latent variable and the input layer $\widetilde{\mathbf{z}}^0$ were chosen such that the total synaptic weight count of the model was approximately equal to the \textcolor{black}{GNCNs}, the linear rectifier was used for the internal activation function, i.e., $\phi^\ell(v) = max(0, v)$, and weight values (for any $\mathbf{W}^\ell$ and $\mathbf{E}^\ell$) were initialized from a centered Gaussian distribution with a standard deviation $\sigma$ that was tuned on held-out validation data. To further improve generalization ability, the decoder weights of all autoencoders were regularized and some autoencoder models had specific meta-parameters that were tuned using validation data. Notably, the GAN-AE was the only model that required a specialized gradient descent rule, i.e., Adam \cite{kingma2014adam}, in order to obtain good log likelihood and to stabilize training (stability is a known issue related to GANs \cite{goodfellow2014generative}).
Finally, we implemented an optimized Gaussian mixture model (GMM) %\cite{mclachlan1988mixture}
that was fit to the training data via expectation-maximization with the number of mixture components chosen based on preliminary experiments that yielded best performance. The only GMM implementation detail worthy of note was that we clipped the images sampled from the mixture model to lie in the range $[0,1]$ (improving likelihood slightly).

\subsection*{Experimental Setup and Task Design}
\paragraph{Training Setup:} 
The parameters of all models, whether they were updated via backprop or by the NGC learning process, were all optimized using stochastic gradient descent using mini-batches (or subsets, randomly sampled without replacement) of $200$ samples for $50$ passes through the data (epochs).
For the backprop-based models, we re-scaled the gradients \cite{pascanu2013difficulty} by re-projecting them to a Gaussian ball with radius of $5.0$, which we found ensured stable performance across trials.
%For the GNCN models, after a weight update was made, the weight matrices were normalized such that the Euclidean norms of their columns were $1.0$.
%\footnote{We did not use column normalization for the autoencoder models as we found that doing so worsened their performance.}
For each model, upon completion of training, we fit a Gaussian mixture model (GMM) to the topmost neural activity layer (to serve as the model prior). This density estimator was trained with expectation-maximization and contained $K = 75$ components, where each component $k \in K$ defines a multivariate Gaussian with mean $\mu_k$ and covariance $\Sigma_k$ parameters as well as a mixing coefficient $\pi_k$.

\paragraph{The Density Modeling Task:}
Given a dataset $\mathbf{X} \in \{0,1\}^{D \times S}$, where $S$ is the number of vector pattern samples and $D$ is the dimensionality of any given pattern, the goal is to learn a density model of $p(\mathbf{X})$, or the probability of sensory input data, where subset of $B$ vectors is denoted as $\mathbf{x} \in \{0,1\}^{D \times B}$. We parameterize the probability distribution $p(\mathbf{X})$ via $p_\Theta(\mathbf{X})$ by introducing learnable parameters $\Theta$. Since computing the marginal log likelihood $\log p_\Theta(\mathbf{X})$ directly is intractable for all the models in this paper, we estimate it by calculating a Monte Carlo estimate using $5000$ samples according to the recipe: 1) sample the GMM prior, 2) ancestrally sample directed neural generative model (whether a baseline or a GNCN) given the samples of the prior.

The reconstruction metric, binary cross-entropy (BCE), also known as the negative Bernoulli log likelihood, was computed as follows: $BCE(\mathbf{X}, \mathbf{\hat{X}}) = -\frac{1}{S} \sum^S_{s=1} \sum^D_{d=1} \Big( \mathbf{X} \log(\mathbf{\hat{X}}) + (1 - \mathbf{X}) \log(1 - \mathbf{\hat{X}}) \Big)[d,s] $ (measured in nats). $\mathbf{\hat{X}} \in \{0,1\}^{D \times S}$ is the predicted sensory input (matrix) produced from a model under evaluation. %, e.g., the GVAE or GNCN.

\paragraph{The Pattern Completion Task:}
For this task, we test each model's ability to complete patterns where the images of each dataset were partially masked and each model was tasked with completing the masked images. Specifically, half of each image $\mathbf{x}$ (containing $\sqrt{D}$ columns of $\sqrt{D}$ pixels) was masked according to a binary mask $\mathbf{m}$ where $\sqrt{D}/2$ columns were set to $1$ and the rest $0$, i.e., $\mathbf{x}_m = \mathbf{x} \odot \mathbf{m}$. We report the masked mean squared error (M-MSE) of each model on each dataset's test set, which is computed per image as follows: $\mbox{M-MSE}(\mathbf{X}, \mathbf{\hat{X}}, \mathbf{M}) = ((\mathbf{\hat{X}} - \mathbf{X}) \odot (1 - \mathbf{M}))^T \cdot ((\mathbf{\hat{X}} - \mathbf{X}) \odot (1 - \mathbf{M})) $ (measured in nats), where $\mathbf{M} \in \{0,1\}^{D \times S}$ is a masking matrix $\mathbf{m}$ where each column contains one mask vector per image vector $\mathbf{x}$. % where $[j,:]$ indexes the $j$th element of a column vector.
% set to flattened notation

\paragraph{The Classification Task:}
In this task, each model's prediction error is measured on the test sample's label set. This is possible given that each dataset $\mathbf{X}$ also comes with a set of target annotations that be encoded into a $C$-dimesnional space yielding the label matrix $\mathbf{Y} \in \mathcal{R}^{C \times D}$ (one label vector per sample), where $C$ is the number of unique categories labeled a priori.
We measure the classification error (\textbf{Err}) (as a percentage \%) on the test set, as follows:
\begin{align*}
    \mbox{Err}(\mathbf{Y},\mathbf{\hat{Y}}) = \bigg(1 - \frac{1}{S} \sum^S_{s=1} 
    \begin{array}{cc}
      \bigg\{ & 
        \begin{array}{cc}
          1 & \hat{y}_s = y_s \\
          0 & \hat{y}_s \neq y_s
        \end{array}
    \end{array}
    \bigg)
\end{align*}
where $\hat{y}_s = \arg \max_d (\mathbf{\hat{Y}}[:,s] )$, the class index chosen by the model, and $y_s = \arg \max_d (\mathbf{Y}[:,s] )$, the index of the actual class label. Note that $\mathbf{\hat{Y}} \in \mathcal{R}^{C \times D}$ is the collected set of predictions from the linear classifier fit to a model's latent space.

\subsection*{Neural Generative Coding Model Procedures}
\label{sec:model}
In this section, we detail the sampling and image completion procedures for the \textcolor{black}{GNCN models}. 

\noindent
\textbf{Sampling from the Model:} 
Since the optimization procedure, whether via gradient descent or another method is not guaranteed to find globally optimal parameter settings (since the objective function is not convex), the distribution of the latent state $\latent{z}^L$ of the final layer will not be Gaussian. To estimate it, after training on the data, we obtain the corresponding $\latent{z}_i^L$ value for each data point $\mathbf{x}_i$. We fit a Gaussian mixture model to this collection of values $\latent{z}_1^L,\dots, \latent{z}^L_N$ (where $N$ is the number of training points). Then, in order to reduce variance during sampling, we take the following approach. First, we sample $\latent{z}^L$ from the Gaussian mixture model, recursively set  $\mean{z}^{\ell-1} \gets g^{\ell-1}(\mathbf{W}^{\ell-1} \phi^{\ell}(\latent{z}^{\ell}))$ and output $\latent{z}^0$. This is similar to how variational auto-encoders are sampled in practice, where the input is a Gaussian and the output is the mean vector.

\noindent
\textbf{Image Completion:} 
In the event that incomplete sensory input $\mathbf{x}$ is provided to the GNCN, i.e., portions of $\mathbf{x}$ are masked out by the variable $\mathbf{m} \in \{0,1\}^{J_0 \times 1}$, we may infer the remaining portions of $\mathbf{x}$ by utilizing the output error neurons of the GNCN and treating the bottom sensory layer $\mathbf{z}^0$ as a partial latent state. Specifically, we update the missing portions, i.e., $1 - \mathbf{m}$, of $\mathbf{z}^0$ as follows:
\begin{align}
    \latent{z}^0 = \mathbf{x} \odot \mathbf{m} + \Big( \latent{z}^0 + \beta( -\frac{\partial \psi }{\partial \mean{z}^0} ) \Big) \odot (1 - \mathbf{m}) = \mathbf{x} \odot \mathbf{m} + \Big( \latent{z}^0 - \beta \mathbf{e}^0 \Big) \odot (1 - \mathbf{m}) \mbox{.}
\end{align}

\begin{figure}[!t]
\centering
\includegraphics[width=14.25cm]{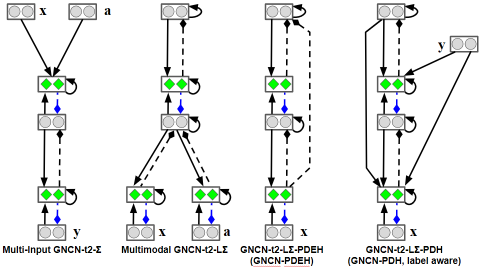} 
%\caption{CalTech.}
%\label{fig:ctech_distrib}
%\vspace{-0.45cm}
\caption{
\textcolor{black}{
\textbf{Variant NGC Architectures:}
Ordered left to right, possible architectures in NGC include: 1) a multi-input GNCN-t2-$\Sigma$, 2) a multimodal GNCN-t2-L$\Sigma$, 
3) a GNCN-t2-L with an asymmetric error transmission pathway (a GNCN-PDEH), and 4) a label-aware GNCN-PDH (or a multi-input GNCN-t2-L$\Sigma$) with an asymmetric generative pathway.
} 
}
\label{fig:ngc_alt_arch}
\vspace{-0.4cm}
\end{figure}

\textcolor{black}{
\section*{Supplementary Note 1: On The Neural Generative Coding Framework}
}

\textcolor{black}{
\textbf{NGC Model Structure and Naming Convention: }Naming GNCN models under the NGC framework entails appending a suffix to the end of the model sub-name. The hyphenated suffix ``-t1'' refers to a GNCN model that uses ``Type 1'' error synapses, or, specifically, error synapses that are a function of the GNCN's forward generative weights, i.e., they are not physically separate/distinct synapses. The suffix ``-t2'' refers to a GNCN model with ``Type 2'' error synapses, or separate learnable synaptic parameters that focus on solely transmistting error messages through the model.
An additional suffix is added to the model depending on whether on it contains lateral synapses in its state variables, i.e., ``-L'', and whether or not it contains lateral precision weights in its error neurons, i.e, ``-$\Sigma$''. As a complete example, a GNCN that contains Type 2 error synapses and lateral synapses in both its state and error neurons would be designated as GNCN-t2-L$\Sigma$.
}
\\
% \begin{table}[h]
%     \centering
\begin{center}
\begin{tabular}{l|c|c|c|c|c}
     \textbf{NGC Model Name} & Error Type & Precision Wghts & Lateral Wghts & $\alpha_m$ Value & Uses $\partial \phi$\\
     \hline
     GNCN-t1/Rao & Type 1 & No & No & $0$ & Yes\\
     GNCN-t1-$\Sigma$/Friston & Type 1 & Yes & No & $0$ & Yes\\
     GNCN-t2-L$\Sigma$ & Type 2 & Yes & Yes & $0$ & No\\
     GNCN-t2-L$\Sigma$-PDH (GNCN-PDH) & Type 2 & Yes & Yes & $1$ & No
\end{tabular}
\end{center}
% \caption{\textcolor{black}{\textbf{NGC Model Details:} Architectural details of each GNCN model experimented with in this study.}
% }
% \label{tab:my_label}
%\end{table}

\textcolor{black}{
While this study explores four variant GNCN models that can be derived from the general NGC computational framework we put forth, there are many other possible architectures that can handle other styles of problems. This flexibility is owed to the fact that our NGC framework supports asymmetry with respect to the structure of the forward generative pathway and that of the error transmission pathway. Additionally, if an NGC model utilizes a non-hierarchical structure in its generative/prediction neural structure, as in the case of one of this paper's main models, it receives one final suffix ``-PDH'' (for partially decomposeable hierarchy). Note that in the case of the GNCN-t2-L$\Sigma$-PHD, we, in this paper, for further convenience, abbreviate it to GNCN-PDH. Alternatively, if an NGC model contains a non-hierarchical error structure (given that is forward and error transmission pathways are not strictly required to be symmetric), it receives the final suffix ``-PDEH'' (for partially decomposeable error hierarchy). Note that special cases of our framework have been provided special name modifications, e.g., GNCN-t1 is also referred to as GNCN-t1/Rao \cite{rao1999predictive} and GNCN-t1-$\Sigma$ is also referred to as GNCN-t1-$\Sigma$/Friston \cite{friston2008hierarchical}.
}

\textcolor{black}{
Figure \ref{fig:ngc_alt_arch} depicts four other alternative neural circuit structures that would tackle, respectively, 
1) clamped, multiple input to generated/predicted outputs (as in the case of tasks such as direct classification), 
2) multi-modal generative modeling (for example, crafting a circuit that jointly learns to synthesize an image and discrete one-hot encoding of a word/character at time step $t$ within a sequence),
3) a generative model where upper layers receive error messages from layers other than its immediately connected one (a GNCN-PDEH, where PDEH means partially decomposable error hierarchy), i.e., layer $\ell = 2$ receives error messages from layer $\ell = 1$ and $\ell = 0$, and
4) a label-aware generative model that forms a partially decomposable hierarchy in its forward generative structure (GNCN-PDH driven by labels as input).
}

\section*{Supplementary Note 2: Related Work on Biologically-Inspired Models \& Learning Algorithms}

As mentioned in the introduction, some of the more notable criticisms of backprop include:
\begin{enumerate}[nosep]
    \item Synapses that make up the forward information pathway need to directly be used in reverse to communicate teaching signals (the weight transport problem), 
    \item Neurons need to be able to communicate their own activation function's first derivative,
    \item Neurons must wait for the neurons ahead of them to percolate their error signals way back before adjusting their own synapses (the update-locking problem),
    \item There is a distinct form of information propagation through a long, global error feedback pathway that only affect weights but does not (at least directly) affect the network's internal representations,
    \item The error signals have a one-to-one correspondence with  neurons.
\end{enumerate}
The properties above are inherent to backprop and do not conform to known biological feedback mechanisms underlying neural communication in the brain \cite{crick1989recent}. The brain, in contrast, is heavily recurrently connected \cite{andersen1963hippocampus,douglas1995recurrent}, allowing for complementary pathways to form that would allow for percolation of error/mismatch information \cite{long2016hippocampal,wacongne2012neuronal}.
Biological neurons communicate binary spike signals, making it unlikely that they also sport specialized circuitry to communicate the derivative of a loss function with respect to their activities \cite{hinton2007recognize}
(though some recent studies have suggested that real neurons might communicate with rate codes \cite{london2010sensitivity}).
Furthermore, it is more commonly accepted that neurons in the brain learn ``locally'' \cite{hebb1949organization,el2018locally}, modulated globally by signals provided through neuromodulators such as dopamine, %deneve2017brain
i.e., they operate with only immediately available information (such as their own activity and that of nearby neurons that they are connected to), making it unlikely that a global feedback pathway drives synaptic weight adjustments.
In addition, several of the problems above result in practical problems -- the weight transport problem has been shown to create memory access pattern issues in hardware implementations \cite{crafton2019local} and the global feedback pathway itself is one key source behind the well-known exploding and vanishing gradient problems \cite{pascanu2013difficulty} % glorot2010understanding 
in deep ANNs, yielding unstable or ineffective learning unless specific heuristics are employed.

In recent research, developing learning procedures that enable backprop-level learning while embodying elements of actual neuronal function has seen increasing interest in the machine learning community. However, while insights provided by each development have proven valuable, increasing the evidence that shows how a backprop-free form of adaptation can be consistent with some aspects  real networks of neurons, many of these ideas only address one or a few of the issues described earlier.
Random feedback alignment algorithms \cite{baldi2016learning,lillicrap2016random} address the weight transport problem, and to varying degrees, the update locking problem \cite{nokland2016direct,mostafa2018deep,frenkel2019learning}, but are fundamentally emulating backprop's differentiable global feedback pathway to create teaching signals (and pay a reduction in generalization ability the farther away they deviate from backprop \cite{bartunov2018assessing,frenkel2019learning}). In addition, experimentally, the success of these approaches depends on how well the feedback weights are chosen a prior (instead of learning them). Other procedures, like local representation alignment \cite{ororbia2019biologically} and target propagation \cite{hinton1988learning,lee2015difference}, which also resolve the weight transport problem and eshew the need for differentiable activations, fail to address the update locking problem, since the various incarnations of these require a full forward pass to initiate inference. 
Procedures have been proposed to address the update locking problem, such as the method of synthetic gradients \cite{jaderberg2017decoupled}, but still require backprop to compute local gradients. It remains to be seen how these algorithms could be adapted to learn generative models.
Other systems, such as those related to contrastive Hebbian learning (CHL)  \cite{movellan1991contrastive,o1996biologically,scellier2018generalization}, %welling2002new
are much more biologically-plausible but often require symmetry between the forward and backward synaptic pathways, i.e., failing to address weight transport. More importantly, CHL requires long settling phases in order to compute activities and teaching signals, resulting in long computational simulation times.
%(at the very minimum to compute a useful negative phase set of values with which to compute difference signals for learning) 
However, while some of these algorithms have shown some success in ANN training \cite{lee2015difference,lillicrap2016random,ororbia2019biologically}, they focus on classification, which is purely supervised and arguably a simpler problem than generative modeling. 

Boltzmann machines \cite{ackley1985learning}, which are generalizations of Hopfield networks \cite{little1974existence,hopfield1982neural} to incorporate latent variables, are a type of generative network that also contains lateral connections between neurons much as the models in our NGC framework do. While training for the original model was slow, a simplification was later made to omit lateral synapses, yielding a bipartite graphical model referred to as a harmonium, trained by contrastive divergence \cite{hinton2002training}, a local contrastive Hebbian learning rule.
%However, these systems were originally trained with the metaheuristic simulated annealing and required visiting individual neurons one at a time in order to properly compute the synaptic adjustment signal. The Boltzmann model was later simplified to omit lateral connections, yielding a bipartite graphical model referred to as a harmonium that could be stacked to create a multi-layer model \cite{salakhutdinov2009deep}, allowing for the development of a training recipe called contrastive divergence \cite{hinton2002training}, a local contrastive Hebbian learning rule.
However, while powerful, the harmonium could only synthesize reasonable-looking samples with many iterations of block Gibbs sampling and the training algorithm suffered from mixing problems (leading low sample diversity among other issues) \cite{dumoulin2013challenges}.
%However, while the harmonium could be used to synthesize data using block Gibbs sampling, it required many iterations to produce samples that looked like actual images. Furthermore, the contrastive divergence learning process suffers from mixing problems that lead to low sample diversity and poor approximations of the log likelihood gradient \cite{dumoulin2013challenges}. 
Another kind of generative model \cite{hinton2005kind} can be trained with the wake-sleep algorithm %\cite{hinton1995wake} 
(or the up-down algorithm in the case of deep belief networks  \cite{hinton2005kind}), where an inference (upward) network and a generative (downward) network are jointly trained to invert each other. Unfortunately, wake-sleep suffers from instability, struggling to produce good samples of data most of the time, due to the difficulty both networks have in inverting each other due to layer-wise distributional shift.
%largely due to the fact that the distributions of the layers of the two networks drift during learning, making it difficult for each to learn how to invert the other.
Motivated by the deficiencies in models learned by contrastive divergence/wake-sleep, algorithms have been created for auto-encoder-based models \cite{bengio2013generalized} but most efforts today rely on backprop.
%such as the walk-back procedure have been proposed for autoencoder-based generative models \cite{bengio2013generalized} but most efforts today rely on backprop to compute synaptic adjustments. 

\section*{Supplementary Note 3: On Evaluating Mode Capture}

\begin{table}[!t]
\begin{center}
 \begin{tabular}{l | l c c | l c c} 
 \hline
  % & \multicolumn{3}{c}{\textbf{MNIST}} & \multicolumn{3}{c}{\textbf{KMNIST}} \\
 \textbf{Model} & & \multicolumn{1}{c}{\textbf{G-KL}} & \multicolumn{1}{c}{\textbf{NDB}} & & \multicolumn{1}{c}{\textbf{G-KL}} & \multicolumn{1}{c}{\textbf{NDB}} \\ %[0.5ex] 
 \hline\hline
  Baseline & \parbox[t]{0.15mm}{\multirow{7}{*}{\rotatebox[origin=c]{90}{MNIST}}} & $1229.3871$ & $1.00 \pm 0.0$ &  \parbox[t]{0.15mm}{\multirow{7}{*}{\rotatebox[origin=c]{90}{KMNIST}}} & $1988.3081$ & $1.00 \pm 0.0$ \\
 RAE & & $180.621 \pm 5.966$ & $0.72 \pm 0.014$ & & $668.821 \pm 36.12$ & $0.72 \pm 0.014$ \\
 GVAE-CV & & $254.976 \pm 3.036$ & $0.63 \pm 0.042$ & & $704.495 \pm 29.394$ & $0.645 \pm 0.049$ \\
 GVAE & & $295.973 \pm 23.266$ & $0.71 \pm 0.028$ & & $804.943 \pm 18.041$ & $0.685 \pm 0.021$ \\
 GAN-AE & & $540.408 \pm 25.943$ & $0.665 \pm 0.035$ & & $1325.641 \pm 39.115$ & $0.74 \pm 0.014$ \\
 %\hline
 \textcolor{black}{GNCN-t1/Rao} & & $682.454 \pm 1.732$ & $0.77 \pm 0.042$ & & $1316.362 \pm 2.463$ & $0.755 \pm 0.049$ \\
 \textcolor{black}{GNCN-t1-$\Sigma$/Friston} & & $531.206 \pm 3.735$ & $0.775 \pm 0.021$ & & $1205.361 \pm 6.5$ & $0.79 \pm 0.014$ \\
 \textcolor{black}{GNCN-t2-L$\Sigma$} & & $596.67 \pm 0.106$ & $0.78 \pm 0.014$ & & $1218.252 \pm 2.761$ & $0.795 \pm 0.12$ \\
 GNCN-PDH & & $446.053 \pm 3.369$ & $0.69 \pm 0.002$ & & $1150.434 \pm 2.68$ & $0.745 \pm 0.007$ \\
 \hline
 \hline
 Baseline & \parbox[t]{0.15mm}{\multirow{7}{*}{\rotatebox[origin=c]{90}{FMNIST}}} & $1811.2068$ & $1.0 \pm 0.0$ & \parbox[t]{0.15mm}{\multirow{7}{*}{\rotatebox[origin=c]{90}{CalTech}}} & $704.3340$ & $1.0 \pm 0.0$  \\
 RAE & & $551.805 \pm 14.6$ & $0.74 \pm 0.028$ & & $57.235 \pm 4.925$ & $0.34 \pm 0.042$ \\
 GVAE-CV & & $661.615 \pm 4.182$ & $0.725 \pm 0.007$ & & $216.271 \pm 8.387$ & $0.32 \pm 0.014$ \\
 GVAE & & $797.931 \pm 1.731$ & $0.685 \pm 0.021$ & & $205.223 \pm 1.316$ & $0.27 \pm 0.002$ \\
 GAN-AE & & $1279.573 \pm 43.131$ & $0.805 \pm 0.035$ & & $266.874 \pm 8.692$ & $0.365 \pm 0.092$ \\
 %\hline
 \textcolor{black}{GNCN-t1/Rao} & & $1173.705 \pm 8.71$ & $0.79 \pm 0.014$ & & $314.684 \pm 1.871$ & $0.325 \pm 0.021$ \\
 \textcolor{black}{GNCN-t1-$\Sigma$/Friston} & & $1040.528 \pm 1.019$ & $0.805 \pm 0.021$ & & $300.303 \pm 0.659$ & $0.4 \pm 0.028$ \\
 \textcolor{black}{GNCN-t2-L$\Sigma$} & & $1180.764 \pm 7.35$ & $0.86 \pm 0.002$ & & $290.617 \pm 0.299$ & $0.395 \pm 0.049$ \\
 GNCN-PDH & & $1091.876 \pm 8.8$ & $0.76 \pm 0.014$ & & $204.032 \pm 1.067$ & $0.31 \pm 0.099$ \\
 \hline
\end{tabular}% 
\end{center}
%\vspace{-0.5cm}
\caption{
\textbf{Distributional measurements:}
Model ability to match the implicit modes underlying each dataset, i.e., MNIST, KMNIST, FMNIST, and CalTech. Metrics reported include the Gaussian Kullback-Leibler divergence (G-KL) and number of statistically different bins (NDB). (Metrics averaged over $10$ trials - we report their mean and standard deviation.)
}
\label{results:distrib_results}
\end{table}

\begin{figure}[!t]
\begin{center}
\begin{subfigure}{0.5\textwidth}
  \centering
  \includegraphics[width=6.25cm]{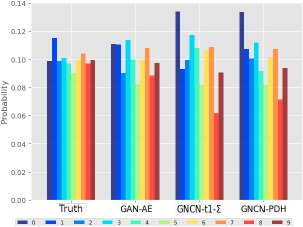}  
  \caption{MNIST.}
  \label{fig:mnist_distrib}
\end{subfigure}%
\begin{subfigure}{0.5\textwidth}
  \centering
  \includegraphics[width=6.25cm]{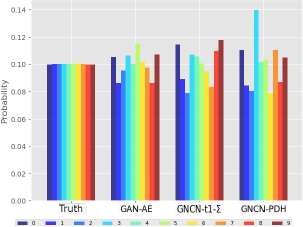} 
  \caption{KMNIST.}
  \label{fig:kmnist_distrib}
\end{subfigure}%
\\
\begin{subfigure}{0.5\textwidth}
  \centering
  \includegraphics[width=6.25cm]{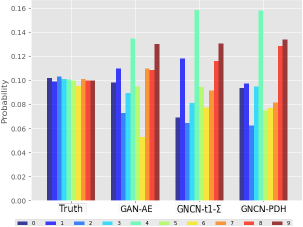} 
  \caption{FMNIST.}
  \label{fig:fmnist_distrib}
\end{subfigure}%
\begin{subfigure}{0.5\textwidth}
  \centering
  \includegraphics[width=6.25cm]{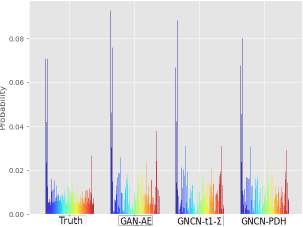} 
  \caption{CalTech.}
  \label{fig:ctech_distrib}
\end{subfigure}
\end{center}
%\vspace{-0.45cm}
\caption{
\textcolor{black}{
\textbf{Class distribution visualization:}
Approximate label distributions (averaged over $10$ trials) produced by each model -- samples from each model were automatically annotated by a regularized MLP discriminator separately trained on original ground truth data. (a) MNIST class distribution, (b) KMNIST class distribution, (c) FMNIST class distribution, and (d) CalTech class distribution. (Note that GNCN-t1-$\Sigma$ is also referred to as GNCN-t1-$\Sigma$/Friston.)
} 
}
\label{fig:distribution}
\vspace{-0.4cm}
\end{figure}

Given that a common problem in training generative models is mode collapse, we measure some distributional properties of several of our baseline models \textcolor{black}{and NGC models}.
Note that it is difficult to directly and automatically determine the labels of the samples produced by the unsupervised models investigated in this paper (aside from manual qualitative inspection) and it is an open research area/problem to develop better measurements for evaluating mode-capturing ability of generative models in general.
Among the myrid of current experimental approaches proposed for measuring the degree of modal capture, we opted to implement and measure the number of statistically different bins (NBD) from \cite{richardson2018gans}, which has been argued to be a potentially useful metric for evaluating the degree of mode collapse that a given generative model might have experienced (values closer to $0$ mean that the model has likely captured most of the modes of the data's underlying distribution). To complement this metric, we also measure the Kullback-Leibler divergence (KL-D) between a pool of samples generated by our model (equal to the size of the original dataset) and the original dataset samples -- we estimate the empirical mean and covariance matrices of each and calculate the closed-form multivariate Gaussian KL-D, to be specific.

To further utilize the labels, we also present results using another approach proposed in \cite{santurkar2018classification}. Specifically, we train a well-regularized multilayer perceptron (MLP) classifier (as mentioned above) on the full original dataset and use it to automatically annotate the samples produced by a given generative model. Once the samples have been annotated, we compute the frequencies of each label class (and furthermore, normalize these values to lie in the range $[0,1]$) and plot these class probabilities in Figure \ref{fig:distribution} (we also plot the ground truth distribution for reference).
Also note that, since even a powerful MLP classifier such as the one we trained incurs error (it naturally cannot reach $100$\% test error), the frequency measurements should be taken with annotator error in mind.

Based on the results presented in Figure \ref{fig:distribution} and Table \ref{results:distrib_results}, we see that while it does not appear that the predictive \textcolor{black}{processing NGC} models suffer from any severe form of mode collapse, they do not appear to capture the frequency distribution of the classes quite as well as the GAN-AE. %(labeled as ``adv\_ae'' in the figure)
With respect to the G-KL and NBD metrics, it does appear that the GNCN-PDH and GNCN-t1-$\Sigma$ do well (\textcolor{black}{often performing among the top} scores across all datasets), though the GAN-AE and GVAE appear to perform the best currently with respect to these mode measurements. Future work will entail uncovering the potential reason why \textcolor{black}{the NGC models} do not, at least upon first examination, match the class frequency of the data as well as the autoencoders.

\begin{table}[!t]
\centering
\begin{tabular}{m{2.2cm}|m{7.1cm}|m{7.1cm}}
\textbf{Model} & \multicolumn{1}{c}{MNIST Samples} & \multicolumn{1}{c}{KMNIST Samples} \\
\hline
\hline
Truth & \includegraphics[width=7cm]{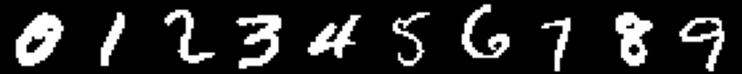}& \includegraphics[width=7cm]{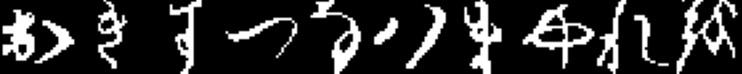}\\
RAE & \includegraphics[width=7cm]{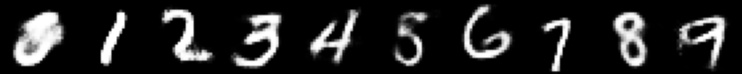}& \includegraphics[width=7cm]{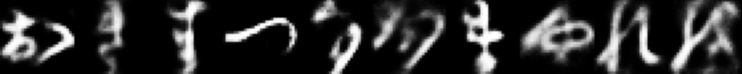}\\
GAN-AE & \includegraphics[width=7cm]{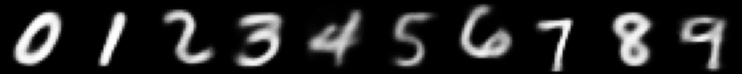}& \includegraphics[width=7cm]{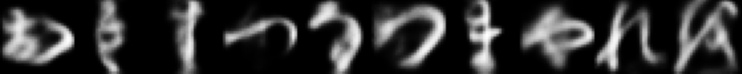}\\
\textcolor{black}{GNCN-t1-$\Sigma$} & \includegraphics[width=7cm]{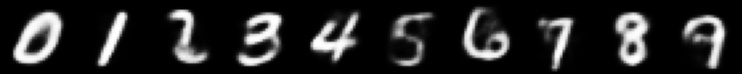}& \includegraphics[width=7cm]{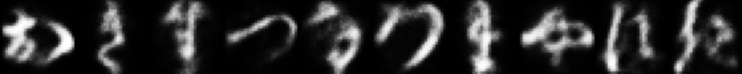}\\
\textcolor{black}{GNCN-t2-L$\Sigma$} & \includegraphics[width=7cm]{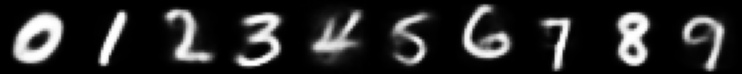}& \includegraphics[width=7cm]{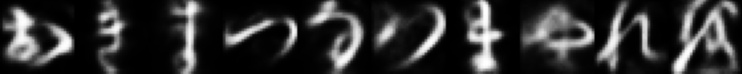}\\
\hline
 & \multicolumn{1}{c}{FMNIST Samples} & \multicolumn{1}{c}{CalTech Samples} \\
\hline
\hline
Truth & \includegraphics[width=7cm]{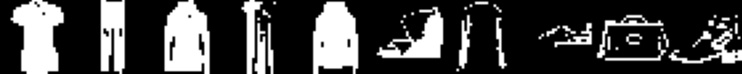}& \includegraphics[width=7cm]{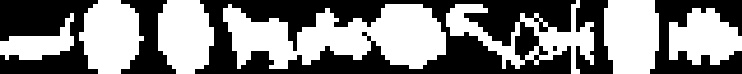}\\
RAE & \includegraphics[width=7cm]{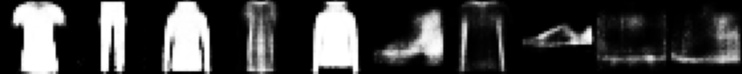}& \includegraphics[width=7cm]{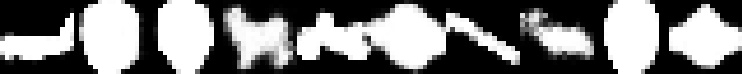}\\
GAN-AE & \includegraphics[width=7cm]{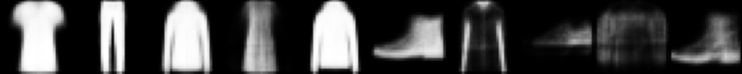}& \includegraphics[width=7cm]{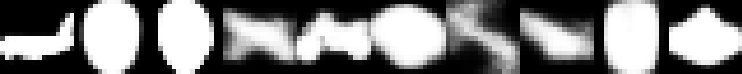}\\
\textcolor{black}{GNCN-t1-$\Sigma$} & \includegraphics[width=7cm]{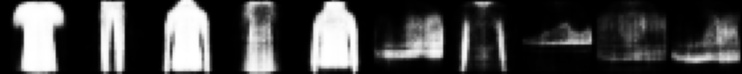}& \includegraphics[width=7cm]{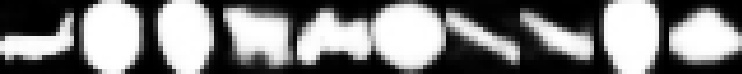}\\
\textcolor{black}{GNCN-t2-L$\Sigma$} & \includegraphics[width=7cm]{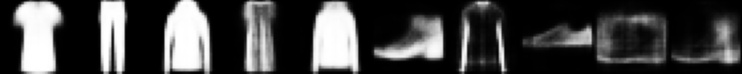}& \includegraphics[width=7cm]{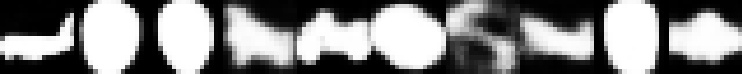}\\
\end{tabular}
%\vspace{-0.25cm}
\caption{
\textbf{Nearest neighbor sample matches:}
For each dataset, i.e., MNIST, KMNIST, FMNIST, and CalTech, we depict the nearest neighbor samples for each class from: ground truth (Truth), the regularized autoencoder (RAE), the adversarial autoencoder (GAN-AE), \textcolor{black}{the 
GNCN-t1-$\Sigma$ (also referred to as GNCN-t1-$\Sigma$/Friston)}, %PC-Friston model (Friston),
and \textcolor{black}{the GNCN-t2-L$\Sigma$}. Specifically, we present the top $10$ nearest neighbor matches to a randomly sampled digit for each class from the original dataset.}
\label{fig:neighbors}
\vspace{-0.5cm}
\end{table}

\section*{Supplementary Note 4: On Model Complexity of Neural Generative Coding}

\subsection*{On Setting Parameter Complexity}
Model complexity was selected based on consideration of the hardware resources available and preliminary experimentation with the validation set of each dataset, i.e., each benchmark had a validation subset that was randomly sampled without replacement (per class) from the training set, as described in the paper. Backprop-based variational autoencoders are typically designed with low-dimensional latent spaces in mind so we investigated different latent code sizes in the range of $[5,30]$ and found that $20$ worked best.% (this is consistent with the general size selected for VAEs, typically around $10$ (XXXXX) ). 
Then, we constrained the \textcolor{black}{GNCNs that had lateral synapses} to only have $20$ neural columns (since each column roughly functions as a latent variable) \textcolor{black}{in their topmost layers} and explored values for the lower levels in the range of $[100,400]$ and found that $360$ was a good choice \textcolor{black}{(GNCN-t1 \cite{rao1999predictive} and GNCN-t1-$\Sigma$ \cite{friston2008hierarchical} were set to use $360$ neurons in the top layer)}.
For the autoencoder models, % For both the autoencoder models
%and other predictive processing models, i.e., PC-Friston \cite{friston2008hierarchical} and PC-Rao \cite{rao1999predictive}, 
the hidden layer sizes were set to be equal and a coarse grid search was conducted through the range between $100$ and the maximum number possible for equal-sized layers such that the model could only have a maximum number of parameters equal to the total of \textcolor{black}{the NGC models} (meaning they could have fewer synapses if that yielded better performance on the validation set), to ensure fair comparison.
 
The optimal dimensionality of $\mathbf{z}^\ell$ was also tuned through preliminary experimentation using held-out validation data. We chose the linear rectifier activation function for the NGC models because we desired strictly positive activity values (which work well with the formulation of lateral inhibition we present in this work). \textcolor{black}{For the GNCN-t1-$\Sigma$ \cite{friston2008hierarchical},} we found that the linear rectifier worked best while the hyperbolic tangent worked best for GNCN-t1 \cite{rao1999predictive} \textcolor{black}{(for both of these models, the coefficient weighting the kurtotic prior was tuned on validation data)}.

\subsection*{On Model Run-time Complexity}

Note that \textcolor{black}{an NGC model,}
%, as well as other predictive processing models like PC-Rao \cite{rao1999predictive} and PC-Friston \cite{friston2008hierarchical}, 
require multiple steps of processing to obtain their latent activities for a given input. Naturally, per sample, this means that all of the predictive processing models we have explored would be slower than the feedforward autoencoder models (which conduct inference with a single feedforward pass).
In short, while an autoencoder (with $L+1$ layers) would roughly only require $2 * L$ matrix multiplications (the most expensive operation in the neural systems we investigate), any NGC predictive processing model would require at least $2 * L * T$ multiplications. 
Note that, as observed in our data efficiency plots in the main paper, we note that this more expensive per-sample cost is desirably offset by convergence with fewer samples in comparison to backprop models. Furthermore, specialized hardware could take advantage of the NGC's inherent parallelism to speed up the process.

One key to reducing the inherent cost of an NGC model's iterative processing include designing alternative state update equations (whereas the ones explored in this paper embody a form of Euler integration, one could design higher-order integration steps, such as those based on the midpoint method or Runga-Kutta). Another solution could be to craft an amortized inference process, where another neural model (much akin to the encoder in a variational autoencoder) learns to infer the value of the state variables at the end of an expensive iterative inference process so as to ultimately reduce the amount of processing steps per sample required. We leave investigation of these remedies for future work.

\section*{Supplementary Note 5: On the Omission of Activation Derivatives}

Removing the activation derivative as we did in our \textcolor{black}{GNCN-t2-L$\Sigma$ and GNCN-PDH models} could be argued to lead to possible value fluctuations that lead to unstable dynamics. Nonetheless, experimentally, we did not observe any strong weight fluctuations in our simulations and we believe that such fluctuations (much like the growing weight value problem inherent to many Hebbian update rules) are unlikely given that our model weight (columns) are constrained to have unit norms. Furthermore, the step size $\beta$ is usually kept within the range of $[0.02, 0.1]$ and the leak variable $-\gamma \mathbf{z}^\ell$ helps to smooth out the values and prevent the occurrence of large values/magnitudes (serving as sort of an L2 penalty over the latent state activities). 

In \cite{ororbia2018conducting,ororbia2019biologically}, it was argued that so long as the activation function was monotonically increasing (with a similar condition imposed for stochastic activation functions), then the learning process would be stable and the benefit that the point-wise derivative offered would be absorbed by the error synaptic weights introduced to carry error signals.
However, note that an approximation for the derivative in the form of a prefactor (derivative) term could be designed/introduced to further safeguard against potential fluctuations (and this will be the subject of future work).

\section*{Supplementary Note 6: Generating the Lateral Competition Matrices}
\label{sec:model_details}

In the main paper, we introduced a lateral competition matrix $\mathbf{V}^\ell$ that directly affects the latent state $\latent{z}^\ell$. It is created to contain the self-excitation weights and lateral inhibition weights by using the following matrix equation: $\mathbf{V}^\ell = \alpha_h (\mathbf{M}^\ell) \odot (1 - I) - \alpha_e (I) $, where $I$ is the identity matrix and the masking matrix $\mathbf{M}^\ell \in \{0,1\}^{J_\ell \times J_\ell}$ is set by the experimenter (placing ones in the slots where it is desired for neuron pairs to laterally inhibit one another). In this study, we set $\alpha_e = 0.13 $ (the self-excitation strength) and $\alpha_h = 0.125$ (the lateral inhibition strength). Our mask matrix $M^\ell$, which emphasized a type of group or neural-column form of competition, was generated by the following process:
\begin{enumerate}
\item create $J_\ell / K$ matrices of shape $J_\ell \times K$ of zeros, i.e., $\{\mathbf{S}_1, \mathbf{S}_2, \cdots, \mathbf{S}_k, \cdots, \mathbf{S}_{C}\}$ (where $C = J_\ell / K$)
\item in each matrix $\mathbf{S}_k$  insert ones at all combinations of coordinates $c = \{ 1, \cdots, k, \cdots, K \}$ (column index) and $r = \{ 1 + K * (k-1), \cdots, k + K * (k-1), \cdots, K + K * (k-1) \}$ (row index)
\item concatenate the $J_\ell / K$ matrices along the horizontal axis, i.e., $\mathbf{M}^\ell = <\mathbf{S}_1, \mathbf{S}_2, \cdots, \mathbf{S}_C>$.
\end{enumerate}

Note that for our proposed integration of the latetal synapses in the state neuron layers, we start out with a probabilistic model and then modify it by introducing sparsity driven by lateral synaptic weights (a neuroscience-inspired idea), which directly modify the values of the state neurons (serving as sort of filter). While we cannot justify them in the probabilistic model, our experiments in the Results section of the main paper show that they improve over those that do not employ them, such as GNCN-t1-$\Sigma$ \cite{friston2008hierarchical} and GNCN-t1 \cite{rao1999predictive} (which both impose a Laplace distribution over their state neurons to encourage sparsity). Part of our future work will be to derive a probabilistic interpretation of our particular extensions to \textcolor{black}{the NGC model}.

With respect to backprop-based neural systems, one could also introduce stateful neurons with similar connectivity to our lateral and precision synapses above by introducing recurrence (as is done in recurrent neural networks). However, to update the weight parameters, one would have to resort to backprop through time (BPTT) and unroll the model over $T$ steps in time. This would require creating a very deep computational graph and storing the activities and gradients at each time step before a final update to each synaptic weight matrix could be calculated, creating a very expensive memory footprint. \textcolor{black}{An NGC model}, in contrast, does not require unrolling and the large memory footprint associated with BPTT-trained recurrent networks.

\section*{Supplementary Note 7: Autoencoder Baseline Model Descriptions}
\label{sec:baseline_details}

To make learning the decoder function ($\mbox{NN}$) described in the main paper tractable, it is common practice in the deep learning literature to introduce a supporting function known as the \emph{encoder} \cite{kingma2013auto}. The encoder ($\mbox{NN}_e$), parameterized by a feedforward network, takes in the input stimulus $\mathbf{x}$ and maps it to $\mathbf{z}$ or to a distribution over $\mathbf{z}$. Depending on the choice of encoder, one can recover one of the four main baselines we experimented with in this paper.

For all backprop-based baseline models in this paper, the decoder of each was regularized with an additional L2 penalty. Specifically, this meant that their data log likelihood objectives always took the form: $\psi_{reg} = \psi + \Omega(\Theta_{\mbox{NN}})$, where $\Theta_{NN} = \{\mathbf{W}^L, \cdots, \mathbf{W}^\ell, \cdots, \mathbf{W}^1 \}$ contains all of the weight matrix parameters of the decoder $\mbox{NN}$. $\Omega_{\mbox{NN}}$ is the regularization function applied to the decoder, i.e., $\Omega_{\mbox{NN}} = -\lambda \sum_{\mathbf{W}^\ell \in \Theta} ||\mathbf{W}^\ell||^2_2 $ where $||\mathbf{W}^\ell||_2$ denotes computing the Frobenius norm of $\mathbf{W}^\ell$. During training/optimization with gradient ascent, we do not constrain the column norms of any of the weight matrices for any of the baseline models (as we do for the GNCNs) as we found that doing so worsened their generalization ability. 

Furthermore, the number of total layers in the decoder for any model was set to be four -- one output and one input layer with two hidden layers in between. The encoder was constrained to be the same -- one input and one output layer with layers in between (in the case of the GVAE, CV-GVAE, and GAN-AE, the encoder's output is technically split into two blocks, as described later). The sizes of the hidden layers were set such that the total number of learnable model weights were approximately equal across all baselines and GNCNs (maximum was $1,400,000$ synapses), which means that all models were forced to have the same parameter complexity to avoid any unfair advantages that might come from over-parameterization.

\noindent 
\textbf{Regularized Auto-encoder (RAE):} The encoder $\mbox{NN}_e$ is designed to be a feedforward network of $L$ layers of neurons. Each layer is a nonlinear transformation of the one before it, where $\mathbf{\hat{z}}^\ell = \phi^\ell(\mathbf{E}^\ell \cdot \mathbf{\hat{z}}^{\ell-1})$. Like in the decoder, $\phi^\ell$ is an activation function and $\mathbf{E}^\ell$ is a set of tunable weights. In this paper, we chose $\phi^\ell$ to be the linear rectifier, i.e., $\phi^\ell(v) = \mbox{max}(0, v)$. The bottom layer activation was chosen to be the logistic link, i.e., $\phi^0(\mathbf{z}) = 1/(1 + exp(-\mathbf{z})$.

Note that in the RAE, the input to the decoder is now $\mathbf{z} = \mathbf{\hat{z}}^L$, i.e., the noise sample vector is set equal to top-most layer of neural activities of the encoder.
The data log likelihood that the RAE optimizes is:
\begin{align}
    \psi = \sum_j \Big(\mathbf{x}[j]\log \mathbf{z}^0[j] + (1-\mathbf{x}[j])\log (1-\mathbf{z}^0[j])\Big) 
\end{align}
where updates to each weight matrix $\mathbf{E}^\ell$ (of the encoder) and $\mathbf{W}^\ell$ (of the decoder) are updated by computing the relevant gradients $\frac{\partial \psi}{\partial \mathbf{E}}$ and $\frac{\partial \psi}{\partial \mathbf{W}}$, respectively. The weight gradients are then used to update model parameters via gradient ascent.

\noindent 
\textbf{Gaussian Variational Auto-encoder (GVAE):} Instead of using an encoder to only produce a single value for $\mathbf{z}$, we could instead modify this network to produce the parameters of a distribution over $\mathbf{z}$ instead. If we assume that this distribution is a multivariate Gaussian with a mean $\mu_z$ and a diagonal covariance $\sigma^2_z = \mathbf{\Sigma}_z \odot \mathbf{I}$, we can then modify the RAE's encoder function to instead be: $(\mathbf{\mu}_z, \sigma^2_z) = \mbox{NN}_e(\mathbf{x})$. Specifically, the top-most layer of $\mbox{NN}_e(\mathbf{x})$ is actually split into two separate output layers as follows: $\mu_z = \mathbf{E}^L_\mu \cdot \mathbf{z}^{L-1}$ and $\sigma^2_z = \exp( \mathbf{E}^L_\sigma \cdot \mathbf{z}^{L-1} )$ (this is also known as the variational autoencoder, or VAE \cite{kingma2013auto}).
$\mathbf{E}^L_\mu$ is the tunable weight matrix for the mean and $\mathbf{E}^L_\sigma$ is the tunable weight matrix for the variance.
The data log likelihood for the GVAE is as follows:
\begin{align}
    \psi = \sum_j \Big(\mathbf{x}[j]\log \mathbf{z}^0[j] + (1-\mathbf{x}[j])\log (1-\mathbf{z}^0[j])\Big) - \mbox{D}_{KL}\Big( q(\mathbf{z}|\mathbf{x}) || p(\mathbf{z}) \Big) \label{eqn:vae_obj}
\end{align}
where $q(\mathbf{z}|\mathbf{x}) = \mathcal{N}(\mu_z, \sigma^2_z)$ (the Gaussian parameters produced by the encoder $\mbox{NN}_e(\mathbf{x})$) and $p(\mathbf{z}) = \mathcal{N}(\mu_p,\sigma^2_p)$ where $\mu_p = 0$ and $\sigma^2_p = 1$ (an assumed unit Gaussian prior over $\mathbf{z}$). The second term in the above objective is the Kullback-Leibler divergence $D_{KL}$ between the distribution defined by the encoder and the assumed prior distribution. This term is meant to encourage the output distribution of the encoder to match a chosen prior distribution, acting as a powerful probabilistic regularizer over the model's latent space. Note that this divergence term serves as a top-down pressure on the top-most layer of the encoder while the gradients that flow from the encoder (via the chain rule of calculus) act as a sort of bottom-up pressure. Note that since $NN_e(\mathbf{x})$ is a distribution, the input to the decoder is, unlike the RAE, a sample of the encoder-controlled Gaussian, i.e., $\mathbf{z} = \mu_z + \sqrt{\sigma^2_z} \odot \epsilon$ where $\epsilon \sim \mathcal{N}(0,1)$.

Gradients of the likelihood in Equation \ref{eqn:vae_obj} are then taken with respect to all of the encoder and decoder parameters, including the new mean and variance encoder weights $\mathbf{E}^L_\mu$ and $\mathbf{E}^L_\sigma$, which are subsequently updated using gradient ascent. All the other activation functions of the GVAE are set to be the linear rectifier, except for the output function $\phi^0$ of the decoder, which, like the RAE, is set to be the logistic sigmoid.

\noindent 
\textbf{Constant-Variance Gaussian Variational Auto-encoder (CV-GVAE):} This model \cite{ghosh2020variational} is identical to the GVAE except that the variance parameters $\sigma^2_z$ of the encoder are omitted and a fixed (non-learnable) value is chosen instead for the variance (meaning that the diagonal covariance is collapsed further to a single scalar). The exact value for this variance meta-parameter, for each benchmark, was chosen from the range $[0.025, 1.0]$ by tuning performance to a held-out set of image samples. 

\noindent
\textbf{Generative Adversarial Network Autoencoder (GAN-AE):} This model, also referred to as an adversarial autoencoder \cite{makhzani2016adversarial}, largely adheres to the architecture of the GVAE except that the second term, i.e., the Kullback-Leibler divergence term, in the data log likelihood is replaced with the adversarial objective normally used to train implicit density estimators like the generative adversarial network (GAN) \cite{goodfellow2014generative}. As a result, we integrate a third feedforward network, i.e., $p_r = \mbox{NN}_d(\mathbf{z})$, into the generative model (this module is also referred to as the discriminator). The discriminator is tasked with distinguishing whether an input vector comes from the desired prior distribution $p(\mathbf{z})$ (set to be a unit Gaussian as in the GVAE) or comes from the encoder network $\mbox{NN}_e(\mathbf{z})$ distribution. This task is posed as a binary classification problem, where a sample from the encoder $\mathbf{z}_f \sim \mathcal{N}(\mu_z, \sigma^2_z)$ is assigned the label of $c = 0$ (fake sample) and a sample drawn from the prior $\mathbf{z}_r \sim \mathcal{N}(0, 1)$ is assigned a label of $c = 1$ (real sample). These fake and real samples are fed through the discriminator which returns a scalar value for each, representing the probability $p_r = p(c=1|\mathbf{z})$. 
This leads to the modified data log likelihood objective below:
\begin{align}
    \psi = \sum_j \Big(\mathbf{x}[j]\log \mathbf{z}^0[j] + (1-\mathbf{x}[j])\log (1-\mathbf{z}^0[j])\Big) + \Big( \log(\mbox{NN}_d(\mathbf{z}_r)) + ( 1 - \log(\mbox{NN}_d(\mathbf{z}_f)) ) \Big) \mbox{.} \label{eqn:adv_obj}
\end{align}
However, to update the weights of the GAN-AE, we do not compute partial derivatives of Equation \ref{eqn:adv_obj} directly. Instead, following in line with the typical multi-step optimization of \cite{goodfellow2014generative,makhzani2016adversarial}, upon presentation of a sample or mini-batch of samples, we compute gradients with respect to $\mbox{NN}_e(\mathbf{x})$, $\mbox{NN}(\mathbf{z})$, and $\mbox{NN}_d(\mathbf{z})$ separately. Specifically, if we group all of the encoder weights under $\Theta_{NN_e}$, all of the decoder weights under $\Theta_{NN}$, and all of the discriminator weights $\Theta_{NN_d}$, then the gradients of the objective are computed in three separate but successive steps shown below:
\begin{align}
    \Delta_{auto} &= \frac{\partial \sum_j \Big(\mathbf{x}[j]\log \mathbf{z}^0[j] + (1-\mathbf{x}[j])\log (1-\mathbf{z}^0[j])\Big)}{\partial (\Theta_{NN_e} \cup \Theta_{NN})} & \mbox{(Autoencoder gradients)} \label{eqn:enc_grad} \\
    \Delta_{gen} &= \frac{\partial \Big( 1 - \log(\mbox{NN}_d(\mathbf{z}_f)) \Big)}{\partial \Theta_{NN}}  & \mbox{(Generator gradients)} \label{eqn:gen_grad} \\
    \Delta_{disc} &= \frac{\partial \Big( \log(\mbox{NN}_d(\mathbf{z}_r)) + ( 1 - \log(\mbox{NN}_d(\mathbf{z}_f)) ) \Big)}{\partial \Theta_{NN_d}}  & \mbox{(Discriminator gradients)} \label{eqn:disc_grad}
\end{align}
where the above gradient calculations are each followed by a separate gradient ascent update to their relevant target parameters, i.e., $\Delta_{auto}$ is used to update $\Theta_{NN_e} \cup \Theta_{NN}$, $\Delta_{gen}$ is used to update $\Theta_{NN_e}$, and $\Delta_{disc}$ is used to update $\Theta_{NN_e} \cup \Theta_{NN_d}$.

The number of synaptic weights associated with the discriminator were included in the model's total parameter count and had two hidden layers of linear rectifier units. Again, like the GVAE, the hidden layer functions $\phi^\ell$ of the encoder and decoder were chosen to be the linear rectifier and the output $\phi^0$ of the decoder was set to be the logistic sigmoid.

\section*{Supplementary Note 8: Feature Analysis of Neural Generative Coding}

The analysis we conducted on the \textcolor{black}{GNCN-t2-L$\Sigma$}'s intermediate representations involved, using the MNIST dataset, examining the generative synaptic weights for each layer of a trained \textcolor{black}{GNCN-t2-L$\Sigma$} model.
%We observe that it appears that the GNCN learns a latent command structure to drive a dynamic composition of low-level visual features.
Specifically, when we viewed the weight vectors that conveyed predictions of state neurons in layer $1$ to layer $0$ (the output), we found that the features resembled rough strokes and digit components (of different orientations/translations). When we viewed the weight vectors relating neurons in layer $2$ to layer $1$ and layer $3$ to layer $2$ 
%(annotated examples for the digit zero are depicted Figure \ref{fig:gncn_feature_compose})
, we found that they rather resembled neural selection ``blueprints'' or maps that seem to be used to select or trigger lower-level state neurons. 

In Figure 7 of the main paper, we illustrate how these higher-level maps seem to interact with the low-level stroke visual features/dictionary. Based on our simple analysis, it appears that the \textcolor{black}{GNCN-t2-L$\Sigma$} learns a type of multi-level command structure, where neurons in one level learn to turn off and turn on neurons in the neighboring level below them, further scaling those they choose to activate by an intensity coefficient.
When we reach layer $1$, the state neurons chosen from the command structure of layers $2$ and $3$, as well as their final produced intensity coefficients (ranging from $[0,\infty)$ due to the fact that any layer of the \textcolor{black}{GNCN-t2-L$\Sigma$} in this paper uses the linear rectifier activation function) resemble and work to produce a composition of low-level features that ultimately produce a complete object or digit. This means that the \textcolor{black}{GNCN-t2-L$\Sigma$} (as well as the other predictive processing models like those of \cite{friston2008hierarchical}, i.e., the GNCN-t1-$\Sigma$, and \cite{rao1999predictive}, i.e., the GNCN-t1, since they process sensory inputs in the same way as the \textcolor{black}{GNCN-t2-L$\Sigma$}) learns to compose and produce a weighted summation of low-level features akin to the results of sparse coding \cite{olshausen1996emergence} driven by a complex, higher-level neural latent structure. In the ``Output'' column of Table 3 of the main paper, we empirically confirm this by summing up the top most highly-activated state neurons in layer $\mathfrak{N}^1$ (multiplying each by its activation coefficient) -- this simple super-position visually yields digits quite similar to the original one presented to the \textcolor{black}{GNCN-t2-L$\Sigma$ model}. 

While this simple feature analysis is promising, we remark that future work should involve developing methodology to map \textcolor{black}{an NGC model's} layerwise activities to those of actually brain activity (using fMRI data) or to a biological model such as HMAX, such as using the method proposed in \cite{ramakrishnan2015visual}.

% \bibliographystyle{acm} 
% \bibliography{supp_ref}

\end{document}